\documentclass{ucthesis}
\usepackage{fancyhdr}
\usepackage{hyperref}
\usepackage[sort=use,style=long3colheader]{glossaries}

\usepackage{epsf}
\usepackage{amsmath}
\usepackage{amsfonts}
\newtheorem{thm}{Theorem}[chapter]
\newtheorem{cor}[thm]{Corollary}
\newtheorem{lem}[thm]{Lemma}

\newtheorem{defin}{Definition}[chapter]
\newtheorem{exmpi}{Example}[chapter]
{}
\newenvironment{defn}{\begin{defin}\em}{\end{defin}}
\newenvironment{exmp}{\begin{exmpi}\em}{\end{exmpi}}
\newenvironment{pf}{\noindent{\bf Proof.\ }}{$\Box$ \\ }

\newcommand{\comp}[2]{{#1\circ#2}}
\newcommand{\mathbold}[1]{\mbox{\boldmath $\bf#1$}}
\renewcommand{\d}[2]{d_\nu\left(#1,#2\right)}
\newcommand{\rangel}[1]{\rangle_{#1}}
\newcommand{\E}[2]{\langle #1\rangel{#2}}
\newcommand{\h}{{\pmb h}}
\renewcommand{\H}{{\mathcal H}}
\newcommand{\XY}{{X\times Y}}
\newcommand{\de}{=}
\renewcommand{\P}{{\mathcal P}}
\newcommand{\Zm}{{Z^m}}
\newcommand{\Zn}{{Z^n}}
\newcommand{\Zmn}{{Z^{(m,n)}}}
\newcommand{\Znm}{{Z^{(n,m)}}}
\newcommand{\Ztmn}{{Z^{(2m,n)}}}
\newcommand{\Gtmn}{{\Gamma_{(2m,n)}}}

\newcommand{\z}{{\bf z}}

\newcommand{\w}{{\bf w}}
\newcommand{\x}{{\bf x}}
\newcommand{\y}{{\bf y}}

\newcommand{\Q}{{\mathcal Q}}

\newcommand{\zh}{{\hat z}}

\newcommand{\fhat}{{\hat f}}

\newcommand{\rhat}{{\hat \rho}}

\newcommand{\be}{\begin{equation*}}

\newcommand{\ep}{{\varepsilon}}

\newcommand{\A}{{\mathcal A}}
\newcommand{\C}{{\mathcal C}}
\newcommand{\N}{{\mathcal N}}
\newcommand{\M}{{\mathcal M}}
\newcommand{\F}{{\mathcal F}}
\newcommand{\G}{{\mathcal G}}

\renewcommand{\(}{\left(}
\renewcommand{\)}{\right)}
\renewcommand{\[}{\left[}
\renewcommand{\]}{\right]}
\newcommand{\Pv}{{\vec{P}}}

\newcommand{\Fbar}{{\overline\F}}

\newcommand{\Pbar}{{\bar{P}}}

\renewcommand{\hbar}{{\overline \h}}

\newcommand{\fbar}{{\bar f}}
\newcommand{\lbar}{{\bar l}}
\newcommand{\zv}{{\vec{z}}}
\newcommand{\gv}{{\vec{g}}}
\newcommand{\fv}{{\vec f}}

\newcommand{\hv}{{\vec h}}
\newcommand{\xv}{{\vec x}}
\newcommand{\yv}{{\vec y}}
\newcommand{\wv}{{\vec w}}
\newcommand{\vv}{{\vec v}}
\newcommand{\av}{{\vec a}}
\newcommand{\Xv}{{\vec X}}
\newcommand{\pr}[3]{{{#1}_1#2\dots#2{#1}_{#3}}}
\newcommand{\iton}{{1\leq i\leq n}}
\newcommand{\prd}[4]{{{#1}_{#2}#3\dots#3{#1}_{#4}}}
\newcommand{\R}{{\mathbb{R}}}

\newcommand{\Rptm}{{(\mathbb{R}^+)^{(2m,n)}}}
\newcommand{\dl}{{\mbox{\boldmath$d_{L^1}$}}}
\newcommand{\dimp}{\operatorname{dimp}}
\newcommand{\plus}{\operatorname{plus}}
\newcommand{\minus}{\operatorname{minus}}

\newcommand{\Tbar}{{\overline T}}
\newcommand{\Sbar}{{\overline S}}
\newcommand{\Xbar}{{\overline X}}
\newcommand{\Hbar}{{\overline \H}}
\newcommand{\bool}{{\{\pm 1\}}}

\newcommand{\jfootnote}[1]{\footnote{#1}}
\newcommand{\jcaption}[2]{\caption{#1\label{#2}}}

\renewcommand{\eqref}[1]{(\ref{#1})}
\makeglossaries
\ssp
\begin{document}
\nocite{Haussler}
\nocite{Retal}
\nocite{H}
\nocite{WD}
\nocite{Pollard}
\nocite{PollardEP}
\nocite{Valiant}
\nocite{CT}
\nocite{VC1}
\nocite{Dudley78}
\nocite{Keifer}
\nocite{VC2}
\nocite{Dud}
\nocite{BH}
\nocite{Getal}
\nocite{Edelman}

\thispagestyle{empty}
\begin{center}
\vspace*{30mm}
{\Large \bf{LEARNING INTERNAL REPRESENTATIONS}}\\
\vspace*{30mm}
{\large \bf Jonathan Baxter} \\
j@baxters.biz
\end{center}
\vspace*{30mm}
A thesis accepted for the degree of Doctor of Philosophy at The
Flinders University of South Australia.
\begin{center}
\vspace*{50mm}
\copyright Jonathan Baxter 1994
\end{center}

\tableofcontents
\pagebreak
\thispagestyle{plain}
\begin{center}
{\large Abstract}
\end{center}
\addcontentsline{toc}{chapter}{Abstract}
\vspace*{10mm}

Most machine learning theory and practice is concerned with learning a
single task. In this thesis it is argued that in general there is
insufficient information in a single task for a learner to generalise
well and that what is required for good generalisation is information
about {\em many similar learning tasks}. The information about similar
learning tasks forms a body of prior information that can be used to
constrain the learner and make it generalise better. Examples of
learning scenarios in which there are many similar tasks are
handwritten character recognition (if one includes the Kanji
characters) and spoken word recognition.

After proving that learning without prior information is impossible
except in the simplest of situations, the concept of the {\em
environment} of a learner is introduced as a probability measure over
the set of learning problems the learner might be expected to
learn. It is shown how a sample from the environment may be used to
learn a {\em representation}, or recoding of the input space that is
appropriate for the environment. Learning a representation can
equivalently be thought of as learning the appropriate features of the
environment.  Using Haussler's statistical decision theory framework
for machine learning, rigorous bounds are derived on the sample size
required to ensure good generalisation from a representation learning
process.  These bounds show that under certain circumstances learning
a representation appropriate for $n$ tasks reduces the number of
examples required of each task by a factor of $n$. It is argued that
environments such as character recognition and speech recognition fall
into the category of learning problems for which such a reduction is
possible.

Once a representation is learnt it can be used to learn {\em novel}
tasks from the same environment, with the result that far fewer
examples are required of the new tasks to ensure good
generalisation. Rigorous bounds are given on the number of tasks and
the number of samples from each task required to ensure that a
representation will be a good one for learning novel tasks.

All the results on representation learning are generalised to cover
any form of automated hypothesis space bias that utilises information
from similar learning problems.

It is shown how gradient-descent based procedures for training
Artificial Neural Networks can be generalised to cover representation
learning. Two experiments using the new procedure are performed. Both
experiments fully support the theoretical results.

The concept of the environment of a learning process is applied to the
problem of {\em vector quantization} with the result that a {\em
canonical} distortion measure for the quantization process emerges.
This distortion measure is proved to be optimal if the task is to
approximate the functions in the environment.

Finally, the results on vector quantization are reapplied to
representation learning to yield an improved error measure for
learning in classifier environments. An experiment is presented
demonstrating the improvement.

\pagebreak
\thispagestyle{plain}
\begin{center}
{\large Declaration}
\end{center}
\vspace*{20mm}
\addcontentsline{toc}{chapter}{Declaration}
I certify that this thesis does not incorporate without acknowledgement any
material previously submitted for a degree or diploma in any university; and
that to the best of my knowledge and belief it does not contain any material
previously published or written by another person except where due reference
is made in the text.
\vspace{20mm}
\begin{center}
Jonathan Baxter
\end{center}
\vspace{20mm}

I believe that this thesis is properly presented, conforms to the
specifications of the thesis and is of sufficient standard to be, {\em prima
facie}, worthy of examination.
\vspace{20mm}
\begin{center}
Professor William Moran
\end{center}

\pagebreak
\thispagestyle{plain}
\begin{center}
{\large Acknowledgements}
\end{center}
\addcontentsline{toc}{chapter}{Acknowledgements}
\vspace*{10mm}
I thank all my friends, family and colleagues at Flinders University
and Adelaide University for their support.  Particular thanks are due
to Tania and also Janice, not least of all for for her help in
submitting the thesis.

I also thank my supervisor, Professor Bill Moran for his patient
guidance of a sometimes headstrong student, and the Australian
Government and Shell Australia for their generous financial
assistance.

I am also grateful to the South Australian Center for Parallel
Computing for providing me with valuable time on their CM5 machine.

This thesis is dedicated to my mother.

\pagebreak
\chapter{Introduction}

Artificial Neural Network (ANN) models have become popular in machine
learning primarily because they offer the possibility of 
{\em nonparametric} or {\em model-free} inference.
It is thought that a neural network with a sufficiently general architecture
should be able to 
learn any training set whatsoever, thus doing away with the laborious,
difficult and
possibly unreliable process of model selection. Unfortunately, although it
is true that neural networks are {\em universal} in the sense that they can
approximate to arbitrarily high accuracy any continuous function, to do so
requires an inordinately large number of parameters, except in the very 
simplest of cases\jfootnote{Even for neural networks with only one node the 
parameter count can be astronomical. For example, a one-node neural network
performing classification on $100\times 100$ images will have $10,000$
parameters!} (of which the {\em xor} problem is probably the most famous).
Ensuring low variance when simultaneously estimating a large number of 
parameters requires prohibitively large training sets, which typically are not 
available in practice. Thus the only option available to the neural network
researcher in most learning problems 
is to use their prior knowledge of the learning problem to try and
reduce the effective number of parameters. 
This is basically the same as introducing a model for the
problem, thus negating the original aim of model-free inference. This
problem is now well recognised in the neural network literature and goes by
the name of {\em the bias/variance dilemma}. A more extensive discussion is
presented in Geman et.\ al.\ (1992)\label{G1}.  

In machine learning in general\jfootnote{Classical and modern methods of
statistical inference should be included as branches
of machine learning also, although a
statistician would probably prefer to view machine learning as a branch of
statistical inference.},
not only in neural network research, it is not the learning problem
itself that poses the most difficulty,
but the task of identification of appropriate prior information, and
how the information should be used to bias the model space. 
Once a manageable sized model has been found, the 
task of parameter estimation is relatively trivial. 
Almost all successes in the application
of neural networks to real-world problems can be attributed to an appropriate
choice of network architecture and data representation so that the effective 
number of parameters is small (in this context see Geman et.\ al.\ again).

The main conclusion to be drawn from the general failure of neural network 
techniques to perform truly model-free inference 
in practice is not that neural networks are a bad model for
learning (although that may be the case for other reasons), but 
that there is simply {\em insufficient
information} in most training sets to specify a solution, no matter what
learning technique is used. On face value this appears to be a quite
depressing conclusion, for it implies that in practice machine learning is a
pretty hopeless task. However, the situation is not quite that bleak, for 
even though there is, in general,  insufficient information in any
individual training set to specify a solution to the problem, it appears
that in many cases the training set together with the {\em prior knowledge} 
of the researcher do contain enough information to solve the problem.  
If the information
source from which researchers derive their prior knowledge could be located,
then perhaps this source could be connected directly to an appropriate
machine learning algorithm, which would then be able to crunch out its own
model biases. This thesis demonstrates one way of doing this.

There are many methods used to bias a large class of models down to a more
manageable size. Approaches used in neural networks include restrictions on
network size, limitations on the number of different weight values,
limitations on the number of non-zero weights, and perhaps the most
successful method of all---reduction in input dimension through appropriate
recoding of the data. This last technique is commonly called ``finding a good
representation for the data''. The work in this thesis is primarily
concerned with the problem of finding a good representation, although
generalisations of the theory are given at the end of chapter \ref{repchap}
that provide a basis for extension to other kinds of model bias.

There are essentially two sources of prior information about learning
problems. The first and most commonly known is knowledge of the physical or
mathematical processes underlying the problem. In many cases
such knowledge is sufficient to generate a physical model of the problem
with few enough parameters that can be estimated using realistically sized
training sets. However, for many problems there is very little information
available about the physical processes underlying the generation of the data.
For example, consider image recognition or character recognition. The data is
essentially being generated by our brains---a very complicated physical
device about which we have very little information. Thus in these
problems another form of prior information is used: {\em knowledge of
similar kinds of learning problems}.

To see how knowledge of similar
learning problems provides information constraining the choice of a good 
representation, consider the problem of learning to recognise the
handwritten digit `1'.
The most extreme representation possible 
would be one that completely solves the classification problem,
i.e. a representation that outputs `yes' if its input is an image of a
`1', regardless of the 
position, orientation, noise or writer dependence of the original digit, and
`no' if any other image is presented to it. Learning to recognise the
digit `1' using such a representation would require only one example of
the digit, for after that the learner would know the image was a `1' if the
representation said `yes' to the image.

Although the representation in this example certainly reduces the complexity
of the learning problem, it does not really seem to capture what is meant by the term
{\em representation}. What is wrong is that although the representation is an
excellent one for learning to recognise the digit `1', 
{\em it could not be used for any
other learning task}. A representation that is appropriate for learning to
recognise `1' should also be appropriate for other character recognition
problems---it should be good for learning other digits, 
or the letters of the alphabet, or Kanji
characters, or Arabic letters, and so on. Thus the information necessary
to determine a good representation is not contained in a single learning
problem (recognising `1'), but is contained in many examples of similar
learning problems. The same argument can be applied to other familiar
learning problems, such as face recognition and speech recognition. A
representation appropriate for learning a single face should be appropriate
for learning all faces, and similarly a single word representation should be
good for all words (maybe even regardless of the language). 
Thus the information for determining a good representation is
not contained in a single learning problem, but in many learning problems of
a similar nature.  
This conclusion is not restricted to model bias through the use of a 
representation, but would apply to any mechanism of model bias for which
information is available regarding similar learning problems. 

How can prior information about similar learning problems be used
{\em automatically} to derive a good representation for the learning
domain? To illustrate the solution to this problem, suppose the learning
problems are all character recognition problems or similar 
(recognise `1', recognise `2', recognise `\pounds', and so
on). Denoting the space of images by $X$, the character recognition problems
can all formally be represented by Boolean functions on $X$. 
For example, the function
$h_{\text{`1'}}\colon X\to \{0,1\}$ such that $h_{\text{`1'}}(x) = 1$ if and only if $x$ is an image
of the character `1', represents the learning problem ``recognise
`1'\,''. The set of {\em similar learning problems} in this case is
formally equivalent to a set of Boolean functions $\H\colon X\to \{0,1\}$. 

\sloppy
In an ordinary machine learning situation, for example learning to
recognise the character `1', 
the learner would receive a {\em training set} $\zv = \{(x_1,
h_{\text{`1'}}(x_1)), \dots, (x_m,h_{\text{`1'}}(x_m))\}$, consisting of sample images $x_i\in X$,
and their correct classifications, $h_{\text{`1'}}(x_i)$. The images $x_i$ would be chosen
from $X$ according to some fixed (but unknown to the learner) 
probability distribution $P$ on $X$. The training set is the only
information provided to the learner.
In order to learn the function $h_{\text{`1'}}$, the learner would have 
available a {\em hypothesis space}---a set of functions 
$\H'$ from which it chooses the best approximation to $h_{\text{`1'}}$ that 
it can find. If the learner is trying to learn
$h_{\text{`1'}}$ using neural networks, $\H'$ would consist of all functions that
can be implemented by neural networks (with perhaps a restriction on the
networks' architectures). In this case the range of the functions $h'\in\H'$
would normally be the interval $[0,1]$, rather than the set $\{0,1\}$,
because continuous neural networks are in general far easier to train than
discontinuous ones (gradient descent cannot be used on discontinuous
networks).
The agreement between a network $h'\in\H'$, and the function
$h_{\text{`1'}}$ is typically measured using the {\em mean-squared} error 
measure,
\begin{equation}
\label{ordemperr}
E(h',\zv) = \frac1m\sum_{i=1}^m \left[h'(x_i)-h_{\text{`1'}}(x_i)\right]^2.
\end{equation}
In this thesis
$E(h',\zv)$ is called the {\em empirical error of $h'$ on sample $\zv$}. 
$E(h',\zv)$ is usually minimized by first selecting a random member $h'$
from $\H'$ and then performing some form of {\em gradient descent} on the parameters of
$h'$. Once a satisfactory solution $h'$ has been found, it is then used to
classify novel images $x$ by calculating $h'(x)$ and declaring the image to
be a `1' if $h'(x) > 0.5$ and not a `1' otherwise. The true utility of
$h'$ is measured not by the empirical error, $E(h',\zv)$, 
which after all can be made zero by a
procedure that simply rote-learns the entire training set $\zv$, but by the
{\em true error},
\begin{equation}
\label{ordtrueerr}
E(h',h_{\text{`1'}}) = \int_X \left[h'(x) - h_{\text{`1'}}(x)\right]^2 \, dP(x).
\end{equation}
The {\em true error} measures how well the function $h'$ will perform {\em
in practice}, i.e. how it will perform on novel examples not included in
its training set. The true error is more commonly referred to in the neural
network literature as the {\em generalisation error}.
The hope is that if the empirical error is small, and the
training set is large enough, then with high probability the true error will
also be small. However this is clearly dependent on the nature of the
learner's hyopthesis space $\H'$, for in the extreme $\H'$ may contain
functions that can rote-learn any sample $\zv$, no matter how large. In
that case $\zv$ could never be made large enough to ensure with high
probability a close agreement between the empirical and true errors. In
practice $\H'$ is never that pathological, but if the learner truly does
have no knowledge of the problems it is likely to encounter, then for it to
be sure that it will have a good $h'\in\H'$ for any problem, it has no
choice but to make $\H'$ very large indeed, with very many parameters to be
estimated, and hence will require an extraordinarily large training set to
ensure a close agreement between the empirical and true errors.
Thus, for learning to be feasible the hypothesis space $\H'$ must be
biased or reduced in some way. 

\fussy
Recall that the actual learning problem in this case (recognise `1') is
but one of many similar learning problems $h\in \H$. To bias the learner's
hypothesis space $\H'$ based on prior information about $\H$, suppose that
the occurence of similar kinds of learning problems
is governed by a probability measure $Q$ on $\H$. Call $Q$ the {\em
environment} of the learner. Any training set 
$\zv = \{(x_1,h(x_1)),\dots,(x_m,h(x_m))\}$ supplied to the learner 
is now assumed to have been generated by  first sampling from 
$\H$ according to $Q$ to generate
$h$, and then sampling $m$ times from $X$ according to $P$ to generate
$x_1,\dots, x_m$ (and hence $\zv = \{(x_1,h(x_1)),\dots,(x_m,h(x_m))\}$).
To provide prior information about similar kinds of
learning problems, $\H$ is sampled $n$ times according to $Q$ to
generate $h_1,\dots, h_n$, and then for each $i$, $1\leq i \leq n$, $X$ is
sampled $m$ times according to $P$ to generate $x_{i1},\dots,x_{im}$. The
full training set is then an $n\times m$ ($n$ rows, $m$ columns) {\em matrix},
$$
\z = 
\begin{matrix}
(x_{11},h_1(x_{11})) &\hdots & (x_{1m},h_1(x_{1m}))\\ 
\vdots & \ddots & \vdots \\
(x_{n1},h_n(x_{n1})) &\hdots & (x_{nm},h_n(x_{nm})).
\end{matrix}
$$
This training set is essentially a list of $n$ ordinary training sets, each
one consisting of examples of a character to be recognised ($h_1$ could be
`1', $h_2$ `A', $h_3$ `\%' and so on).
The learner uses the information contained in $\z$ to learn a representation 
appropriate for the environment $Q$ by first splitting its hypothesis space
in two: $\H' = \comp{\G}{\F}$\jfootnote{$\comp{\G}{\F} = \{\comp{g}{f}\colon
g\in\G, f\in\F\}$.} where $\F$ is a set of functions mapping $X$
into some space $V$, and $\G$ is a set of functions mapping $V$ into $[0,1]$.
This splitting will be denoted by 
$$
X \xrightarrow{\F} V \xrightarrow{\G} [0,1].
$$
If $\H'$ consists of neural networks, one way for it to be split is to
let $\F$ include all the hidden layers of $\H'$ and let $\G$ consist simply
of the last layer of $\H'$. In this case $V$ would be the space $[0,1]^p$
where $p$ is the number of nodes in the last hidden layer of the networks in
$\H'$ (assuming the hidden nodes are $[0,1]$-valued functions). Clearly $\H'$
could in fact be split at any layer. The representation is to be chosen 
from the space $\F$, and hence $\F$ will be called the {\em
representation space}. Each individual $f\in \F$ is a candidate
{\em representation} for the environment $Q$. The learner wishes to find a
representation $f\in \F$ such that 
a learning problem $h$ selected from $\H$ according to $Q$, will, with high
probability, be well approximated by some function $g\in\G$ composed with $f$.
In other words, $f$ should be such that if the learner learns within the
environment $Q$, using the hypothesis space
$\comp{\G}{f}$\jfootnote{$\comp{\G}{f} = \{\comp{g}{f}\colon g\in\G\}$.} rather than the
full space $\H' = \comp{\G}{\F}$, then with high probability it will be
successful. Note that learning with $\comp{\G}{f}$ will require far fewer
examples in general than learning using the full space $\comp{\G}{\F}$,
because only the parameters of $\G$ need to be estimated, rather than {\em
both} the parameters of $\G$ {\em and} the parameters of $\F$. 

To measure the effectiveness of a representation $f$ on a sample $\z$ as
above, define the {\em empirical error} $E(f,\z)$ by
$$
E(f,\z) = \frac1n\sum_{i=1}^n \inf_{g\in\G} \frac1m\sum_{j=1}^m 
\left[\comp{g}{f}(x_{ij}) - h_i(x_{ij})\right]^2.
$$
If each row of $\z$ is denoted by $\zv_i$, so that $\zv_i = \{(x_{i1},
h_i(x_{i1})), \dots, (x_{im}, h_i(x_{im}))\}$, then $E(f,\z)$ can be rewritten
as
\begin{equation}
\label{repemperr}
E(f,\z) = \frac1n\sum_{i=1}^n \inf_{g\in\G} E(\comp{g}{f}, \zv_i),
\end{equation}
where $E(\comp{g}{f}, \zv_i)$ is the ordinary empirical error of the
function $\comp{g}{f}$ on sample $\zv_i$, as defined in \eqref{ordemperr}.
The empirical error of a representation $f$ is a measure of how well the learner
can learn using $f$, assuming that the learner is able to find the best
possible $g\in \G$ for any given sample $\zv$ (this assumes such a $g$ exists;
see \ref{cel} for further discussion). For example, if the empirical
error of $f$ on $\z$ is zero, it is possible for the learner to find a
function $g_i$ for each $i$, $1\leq i\leq n$, such that the function
$\comp{g_i}{f}$ agrees exactly with the function $h_i$ on the sample $\zv_i$. 

There are two different natural ways of measuring the {\em true error} (as
distinct from the empirical error) of a
representation $f$. The first is a measure of the true performance of $f$
with respect to the $n$ functions $(h_1,\dots, h_n)$ used to derive the
training set $\z$:
\begin{equation}
\label{reptrueerr1}
E(f,h_1,\dots,h_n) = \frac1n\sum_{i=1}^n \inf_{g\in\G} E(\comp{g}{f}, h_i),
\end{equation}
where $E(\comp{g}{f}, h_i)$ is the true error of $\comp{g}{f}$ with respect
to $h_i$, as defined in \eqref{ordtrueerr}.  
In chapter \ref{repchap} it is shown that for
problem domains in which the number of parameters in $\F$ greatly exceeds
the number of parameters in $\G$, the
number of examples $m$ required of each function in the sample $\z$ to ensure
that \eqref{repemperr} and \eqref{reptrueerr1} are close is 
roughly a fraction $\frac1n$ times the number of examples required to ensure 
that \eqref{ordemperr} and \eqref{ordtrueerr} are close. This gives the same
number of examples in total as the ordinary learning situation ($\frac1n$
times as many per function, but there are $n$ functions in all) but they are
spread across many functions. This is a genuine advantage if the learner can
easily obtain data sets corresponding to many similar learning problems, as
is the case for character recognition (if sampling includes
Kanji characters, the number of different learning tasks available
is in the thousands), face recognition (there are about five billion faces in
the world at the moment), speech recognition (the number of different words
in all the different languages must number in the millions), and so on. Note
also that even though the total sample size required for good generalisation
is not reduced by sampling from many similar tasks, it is still a good idea
to sample in this way because the more tasks that are 
sampled from, the more likely it is
that the resulting representation $f$ will be a good one for the environment
concerned, and so will be useful for learning future tasks drawn from the
same environment. In fact the second natural definition of the true error of $f$ 
is a measure of how good $f$ will be for learning tasks from $\H$ in
general:
\begin{equation}
\label{reptrueerr2}
E(f, Q) = \int_\H \inf_{g\in\G} E(\comp{g}{f}, h) \, dQ(h).
\end{equation}
A representation with a small value for $E(f,Q)$ will, with high
probability, be suitable for learning any function $h\in\H$ drawn randomly
according to $Q$, assuming that the learner is able to find the best
possible $g\in \G$ in every case. 

Clearly, to ensure small deviation between
\eqref{repemperr} and \eqref{reptrueerr2}, both $n$---the number of tasks
being learnt, and $m$---the number of times each task is sampled, must be
sufficiently large. Roughly speaking, it turns out that $n$ must exceed
the number of examples that would be required for ordinary learning of a
single task, if the full hypothesis space $\comp{\G}{\F}$ is used. This is
a rather depressingly large bound, but it should be noted that it arises from
a worst-case analysis, and that in practice it is likely that far fewer
learning tasks will be required. This is certainly borne out by the
experiments presented in chapter \ref{expchap}.

Although the theoretical and experimental results presented in this thesis
show that representation learning goes some of the way towards solving the
problem of automatic model bias, it is {\em not} the claim of this thesis
that the entire problem has been solved. To ensure feasible representation
learning in practice, the representation space $\F$ will still have
to be subject to some kind of preliminary bias. However the bias will be far
less than that required in an ordinary learning scenario, and could perhaps 
be implemented simply through limited knowledge of natural neural structures
known to be effective for the problem, or through considerations of
computational complexity issues. This is a subject deserving further
investigation.

Although this thesis is principally devoted to theoretical and experimental
investigations concerning the issue of representation learning and
model-bias in general, it turns out that the framework developed also 
provides a solution to a very different problem: how to choose an
appropriate distortion measure for vector quantization. This is discussed
in chapter \ref{quantchap}.

\section{Overview of the thesis.}
Most of the thesis is devoted to representation learning. All theoretical
results relating to representation learning are derived within the {\em
statistical decision theory} framework of machine learning introduced by
Haussler (1992)\label{H1}, as this is the most refined and generally applicable
machine learning framework so far developed. 

Chapter \ref{ordchap} begins
with a discussion of the main theoretical problems associated with machine
learning, namely, how can one be confident that a given model will
accurately approximate the data, and how can one be sure of good
generalisation. Formal definitions of these concepts are given, under the
titles of ``EC'' and ``PC'' learning. A theorem of Haussler
(1992)\label{H2}
bounding the
number of examples required for PC learning is then discussed. PC and EC
learning together
are then shown to be equivalent to a variation of ``PAC'' learning as
introduced by Valiant (1984)\label{V1}. It is shown that in the absence of any
prior information about the problem domain, PAC learning is impossible,
at least if the problem is to learn continous functions on $\R$.

Chapter \ref{repchap} 
is devoted to a discussion of a more general nature
than the one in this introduction,
of how prior information may be introduced through representation learning,
and a subsequent derivation of the number of examples (and learning tasks)
required to ensure, at least in the worst case, good generalisation
performance from a representation. In the final part of the chapter a
mathematical framework is introduced 
that is applicable to any kind of machine learning procedure that uses
knowledge of similar learning problems to bias its hypothesis space (or
models). Generalisations of the sample size bounds for representation
learning are derived.

In chapter \ref{expchap} an algorithm is presented for training neural
networks to learn representations. It is based on gradient-descent
procedures for ordinary learning problems. Three separate experiments are
given, two of which use the gradient procedure and one which uses exhaustive
search. In all experiments the theoretical results of chapter \ref{repchap}
are well supported. In the first gradient-based 
experiment, a representation
appropriate for learning a simple kind of
translationally-invariant Boolean function is
learnt, and in the second a representation for learning symmetric Boolean
functions is learnt.

The next chapter, chapter \ref{quantchap}, shows how the idea of an
{\em environment} of a learning
process can be applied to the problem of vector quantization, with the
result that a natural definition of the distortion measure between two
signals emerges. This distortion measure is shown to be optimal if the task
is to approximate functions from the environment. A very brief discussion is
given showing how the distortion measure may be estimated using
representation learning techniques.

The ideas of the quantization chapter are then
reapplied to representation learning in
the final chapter, chapter \ref{flearnchap}, and as a result a superior
measure of the error of a representation is derived.
An experiment is presented 
showing the effectiveness of this improved error measure.

Most of the more technical mathematical results have been relegated to the
appendices. This is not because they are trivial or irrelevant, but because
to have included them in the main text would have upset the
flow of the discussion too much. The appendices have been written so as to
be self-contained and hence can be read through in their own right,
although for much of the motivation the reader may need
to refer to the main text. Appendix \ref{permissapp} contains definitions and
lemmas relating to measurability criterion needed elsewhere in the work.
Appendix \ref{fundapp} is devoted to a proof of the {\em fundamental
theorem} which is crucial in bounding the number of examples needed to
ensure good generalisation from a representation learner. Appendix
\ref{capapp} contains definitions and results that allow the {\em capacity}
of a complex class of functions to be bounded in terms of its simpler
constituents. Finally, appendix \ref{nnetapp} applies the results of
appendix \ref{capapp} to the problem of bounding the capacity of {\em
feedforward neural networks}.

\newglossaryentry{X}{name={\ensuremath{X}}, description={Input Space}}
\newglossaryentry{Y}{name={\ensuremath{Y}}, description={Output Space}}
\newglossaryentry{Z}{name={\ensuremath{Z}}, description={Sample Space (\ensuremath{=X\times Y})}}
\newglossaryentry{P}{name={\ensuremath{P}}, description={Probability measure on \ensuremath{Z}}}
\newglossaryentry{l}{name={\ensuremath{l}}, description={Loss function}}
\newglossaryentry{A}{name={\ensuremath{A}}, description={Action space}}
\newglossaryentry{H}{name={\ensuremath{\H}},description={Hypothesis space}}
\newglossaryentry{h}{name={\ensuremath{h}}, description={Hypothesis}}
\newglossaryentry{l_h}{name={\ensuremath{l_h}}, description={Loss function for hypothesis \ensuremath{h}}}
\newglossaryentry{El_hP}{name={\ensuremath{\E{l_h}{P}}},  description={expected value of \ensuremath{l_h}}, description={Expected value of \ensuremath{l_h}}}
\newglossaryentry{l_H}{name={\ensuremath{l_\H}}, description={Set of loss functions}}
\newglossaryentry{El_hzv}{name={\ensuremath{\E{l_h}{\zv}}}, description={Empirical estimate of \ensuremath{l_h}}}
\newglossaryentry{AA}{name={\ensuremath{\A}}, description={Learner}}
\newglossaryentry{d_nu}{name={\ensuremath{d_\nu}}, description={Metric on \ensuremath{\R^+}}}
\newglossaryentry{PC}{name={PC}, description={Probably consistent}}
\newglossaryentry{d_P}{name={\ensuremath{d_P}}, description={Pseudo-metric on functions}}
\newglossaryentry{Fd_P}{name={\ensuremath{\(\F,d_P\)}}, description={Pseudo-metric space}}
\newglossaryentry{NepFd_P}{name={\ensuremath{\N\(\ep,\F,d_P\)}}, description={Smallest \ensuremath{\ep}-cover}}
\newglossaryentry{MepFd_P}{name={\ensuremath{\M\(\ep,\F,d_P\)}}, description={Packing number}}
\newglossaryentry{sigma_F}{name={\ensuremath{\sigma_\F}}, description={\ensuremath{\sigma}-algebra on \ensuremath{Z} induced by \ensuremath{\F}}}
\newglossaryentry{P_F}{name={\ensuremath{\P_\F}}, description={Probability measures on \ensuremath{Z} induced by \ensuremath{\F}}}
\newglossaryentry{CepF}{name={\ensuremath{\C(\ep,\F)}}, description={\ensuremath{\ep}-capacity of \ensuremath{\F}}}
\newglossaryentry{rP}{name={\ensuremath{r^*(P)}}, description={Best possible loss}}
\newglossaryentry{anu}{name={\ensuremath{(\alpha,\nu)}-EC},description={\ensuremath{(\alpha,\nu)}-empirically correct}}
\newglossaryentry{anuPAC}{name={\ensuremath{(\alpha,\nu)}-PAC}, description={\ensuremath{(\alpha,\nu)}-probably approximately correct}}
\newglossaryentry{Q}{name={\ensuremath{Q}}, description={Environmental measure}}
\newglossaryentry{z}{name={\ensuremath{\z}}, description={\ensuremath{(n,m)}-sample}}
\newglossaryentry{Znm}{name={\ensuremath{\Znm}}, description={\ensuremath{n\times m} matrices over \ensuremath{Z}}}
\newglossaryentry{nmsample}{name={\ensuremath{(n,m)}-sample}, description={\ensuremath{(n,m)}-sample}}
\newglossaryentry{EGfz}{name={\ensuremath{E^*_\G(f,\z)}}, description={Empirical representation loss}}
\newglossaryentry{EGfQ}{name={\ensuremath{E^*_\G(f,Q)}}, description={True representation loss}}
\newglossaryentry{Ehvz}{name={\ensuremath{E(\hv,\z)}}, description={Empirical loss of sequence of hypotheses \ensuremath{\hv}}}
\newglossaryentry{EhvPv}{name={\ensuremath{E(\hv,\Pv)}}, description={True loss of sequence of hypotheses \ensuremath{\hv}}}
\newglossaryentry{Hn}{name={\ensuremath{\H^n}}, description={Set of sequences of hypothesis}}
\newglossaryentry{Hbar}{name={\ensuremath{\Hbar}}, description={Diagonal hypothesis space}}
\newglossaryentry{GnFbar}{name={\ensuremath{\comp{\G^n}{\Fbar}}}, description={Space for representation learning}}
\newglossaryentry{gvfbar}{name={\ensuremath{\comp{\gv}{\fbar}}}, description={Element of \ensuremath{\comp{\G^n}{\Fbar}}}}
\newglossaryentry{EGfPv}{name={\ensuremath{E^*_\G(f,\Pv)}}, description={True loss of representation (2)}}
\newglossaryentry{EfPv}{name={\ensuremath{\E{f}{\Pv}}}, description={Expected value of \ensuremath{f:\Zn\to [0,M]}}}
\newglossaryentry{Efz}{name={\ensuremath{\E{f}{\z}}}, description={Empirical estimate of \ensuremath{f:\Zn\to [0,M]}}}
\newglossaryentry{lbargf}{name={\ensuremath{\lbar_\comp{g}{f}}}, description={Induced function on probability measures}}
\newglossaryentry{lbarGF}{name={\ensuremath{\lbar_{\comp{\G}{\F}}}}, description={Space of induced functions on probability measures}}
\newglossaryentry{lf}{name={\ensuremath{l^*_f}}, description={Function on probability measures}}
\newglossaryentry{lF}{name={\ensuremath{l^*_\F}}, description={Space of functions on probability measures}}
\newglossaryentry{dQ}{name={\ensuremath{d_Q}}, description={Psedudo-metric on functions on probability measures}}
\newglossaryentry{HH}{name={\ensuremath{H}}, description={Hypothesis space family}}
\newglossaryentry{EHz}{name={\ensuremath{E^*\(\H,\z\)}}, description={Empirical loss of hypothesis space}}
\newglossaryentry{EHQ}{name={\ensuremath{E^*(\H,Q)}}, description={True loss of hypothesis space}}
\newglossaryentry{Edxv}{name={\ensuremath{E_d(\xv)}}, description={Reconstruction error}}
\newglossaryentry{rho}{name={\ensuremath{\rho(x,y)}}, description={Canonical distortion measure}}
\newglossaryentry{EFxX}{name={\ensuremath{E_\F(\xv,\Xv)}}, description={Reconstruction error of \ensuremath{\F}}}
\newglossaryentry{rhoG}{name={\ensuremath{\rho_\G(v,w)}}, description={Distortion measure}}
\newglossaryentry{sigma_H}{name={\ensuremath{\sigma_\H}}, description={Sigma algebra on \ensuremath{Z} induced by \ensuremath{\H}}}
\newglossaryentry{P_H}{name={\ensuremath{\P_\H}}, description={Probability measures on \ensuremath{\sigma_\H}}}
\newglossaryentry{h_1h_n}{name={\ensuremath{h_1\oplus\dots\oplus h_n}}, description={Average of \ensuremath{n} functions}}
\newglossaryentry{H_1H_n}{name={\ensuremath{\H_1\oplus\dots\oplus\H_n}}, description={Set of averages of \ensuremath{n} functions}}
\newglossaryentry{H_sigma}{name={\ensuremath{H_\sigma}}, description={Union of hypotheses in family \ensuremath{H}}}
\newglossaryentry{HHn}{name={\ensuremath{H^n}}, description={Set of average hypothesis spaces in family \ensuremath{H}}}
\newglossaryentry{Xrho}{name={\ensuremath{\(X,\rho\)}}, description={Pseudo-metric space}}
\newglossaryentry{NepXrho}{name={\ensuremath{\N\(\ep,X,\rho\)}}, description={Smallest \ensuremath{\ep}-cover}}
\newglossaryentry{MepXrho}{name={\ensuremath{\M\(\ep,X,\rho\)}}, description={Packing number}}
\newglossaryentry{pseudo-dimension}{name={\ensuremath{\dimp(\H)}},description={Pseudo-dimension of the hypothesis space \ensuremath{\H}}}
\newglossaryentry{d_Pl_G}{name={\ensuremath{d_{[P,l_\G]}}}, description={Metric on the first component of a composition}}
\newglossaryentry{dstar_Pl_G}{name={\ensuremath{d^*_{[P,l_\G]}}}, description={Metric on component of a composition (2)}}
\newglossaryentry{C_l_GepF}{name={\ensuremath{C_{l_\G}(\ep,\F)}}, description={\ensuremath{\ep}-capacity of a component of a composition}}
\newglossaryentry{Cstar_l_GepF}{name={\ensuremath{\C^*_{l_\G}(\ep,\F)}}, description={\ensuremath{\ep}-capacity of a component of a composition (2)}}
\newglossaryentry{d_L1}{name={\ensuremath{d_{L^1(P)}}}, description={Metric on functions}}
\newglossaryentry{CepHd_L1}{name={\ensuremath{\C\(\ep,\H,d_{L^1}\)}}, description={Capacity  of function space}}

\printglossaries

\chapter{Ordinary Learning}
\label{ordchap}
The {\em statistical decision theory} 
formulation of machine learning due to Vapnik \cite{Vapi} and
Haussler \cite{Haussler} \label{H3} is introduced and the notion of a {\em
Probably Consistent} or PC learning procedure is defined. 
Haussler's treatment is
then reproduced to give a theorem bounding the number of examples required
for PC learning. A probably consistent learner needs only to perform at the
same level in practice as it does in training, there is no requirement that
it actually perform well. For the learner to be useful it must be both
probably consistent (PC) and {\em Empirically Correct} (EC). An empirically
correct learner is one that performs well in training. Both PC and EC
together imply a version of the well-known PAC (Probably Approximately
Correct) criterion introduced by Valiant (1984)\label{V2}. However it is
shown that for very general learning scenarios, {\em in the absence of prior
information}, it is impossible for a learner to be PAC.

\section{The General Problem}
\label{genprob}
\sloppy Under the statistical decision theory formulation of machine
learning the learner is supplied with a {\it training set} or {\it
  sample} $\zv=\{z_1,z_2,\ldots,z_m\}$, where each element $z_i =
(x_i,y_i)$ consists of an {\em input} $x_i\in X$ and an {\em outcome}
$y_i\in Y$, \glsname{X} and \glsname{Y} being arbitrary sets known as
the {\em input} and {\em outcome spaces} respectively.  The sample is
generated by $m$ independent trials from $\glsname{Z}=X\times Y$
according to an unknown joint probability distribution \glsname{P}.
In addition the learner is provided with an {\em action} space
\glsname{A}, a loss function $\glsname{l}\colon Y\times A \to [0,M]$
and a {\em hypothesis}\jfootnote{In statistical decision theory the
  hypotheses are more commonly called {\it decision rules}.}  space
\glsname{H} containing functions $\glsname{h}\colon X\to A$.  Defining
the {\em expected} or {\em true loss} of hypothesis $h$ with respect
to distribution $P$ as
\begin{equation}
\label{terr}
E(h,P) \de \int\limits_\XY l\(y, h(x)\)\,dP(x,y),
\end{equation}
the goal of the learner
is to produce a hypothesis $h\in\H$ that has expected loss close to zero.
$l(y,h(x))$ is designed to give a measure of the {\em loss} the learner
suffers, when, given an input $x\in X$, it produces an action $h(x)\in A$ and
is subsequently shown the outcome $y\in Y$.
\fussy

If, for each $h\in \H$, a function $\glsname{l_h}\colon Z\to[0,M]$ is
defined by $l_h(z) = l(y,h(x))$ for all $z=(x,y)\in Z$, then $E(h,P)$
can be expressed as the expectation of $l_h$ with respect to $P$,
$$
E(h,P) = \glsname{El_hP} \de \int_Z l_h(z)\, dP(z).
$$
Let $\glsname{l_H} \de \{l_h\colon h\in\H\}$.  The measure $P$ and the
$\sigma$-algebra on $Z$ are assumed to be such that all the $l_h\in
l_\H$ are $P$-measurable\jfootnote{The space of measures $\P_{l_\H}$
  and the $\sigma$-algebra, $\sigma_{l_\H}$ in definition
  \ref{measure} have been designed explicitly to fulfill this
  purpose.}.

The learner does not know the exact distribution $P$, 
it is only given a sample 
$\zv=\{\pr{z}{,}{m}\}$ drawn from $Z$ according to $P$. 
Hence the learner does not have enough 
information to find a function $h\in\H$ minimizing the true loss,
$\E{l_h}{P}$. A typical approach is for the learner to
search for a function that 
minimizes the {\em empirical loss}\footnote{There are more sophisticated 
methods for choosing a hypothesis based on the information in a training set,
for example the technique presented in \cite{BK2}. We will not be 
analysing those methods here.} on the sample $\zv$,
\begin{equation}
\label{eerr}
E(h,\zv) \de \frac1m\sum_{i=1}^m l_h(z_i).
\end{equation}
Let \glsname{El_hzv} be a synonym for $E(h,\zv)$.

Translating, for example, the problem of training a neural network to
recognize the hand-written character ``A'' into this framework: the input space $X$
would be the space of all possible images, the outcome space $Y$ would be
$\{0,1\}$---``1'' for an example of ``A'' and ``0'' otherwise. $P$ would be the
environmental distribution over characters, with $P(y|x)$ being either $1$
or $0$ according to whether the classification $y$ is correct for $x$ or
not. $\H$ would be a class of feed-forward neural networks mapping $X$ into
some interval of the real line, say $[0,1]$, which would be the action space
$A$. Typically, in these problems the loss function $l$ is
the squared error: $l(y,a) = (y-a)^2$ for all $(y,a)\in Y\times A$. The true
loss of a hypothesis $h\in\H$ is then the expected mean-squared difference between
$h(x)$ and $y$ on a sample $(x,y)$ drawn at random according to $P$, and the
empirical loss on a training sample $\zv=(z_1,\dots,z_m), z_i=(x_i,y_i)$ 
is the {\em mean-squared error} of
the hypothesis with respect to the sample: $\E{l_h}{\zv} = 1/m\sum_{i=1}^m
(h(x_i)-y_i)^2$. A hypothesis $h$ with zero empirical loss correctly
classifies all examples in the training set and if $h$ also has zero true loss 
it will correctly classify all examples in practice. 

The purpose of the loss function in this formulation is that it allows the
learner to use the same hypothesis space $\H$ under many different
circumstances, simply by changing the definition of the loss function. This
permits results derived under specific learning scenarios to apply in a
wider variety of situations. In
Haussler (1992)\label{H4}, section 1.1, many more quite diverse examples of machine
learning and classical statistical problems are translated into this framework.

The behaviour of the learner is to take in samples $\zv\in Z^m$ and
produce as output hypotheses in $\H$. Thus the learner can formally be
denoted as a map $\A$ from the space of all samples into $\H$,
$$
\glsname{AA} \colon \bigcup\limits_{m\geq 1} \Zm \to \H.
$$
This notation implies that $\A$ is a function, however it is a trivial
matter to also treat stochastic learners within this framework.

The two main questions to ask of any learner $\A$ are under what
conditions will the learner be able to produce a hypothesis $h$ with
small empirical loss \eqref{eerr}, and with what confidence will the
empirical loss be close to the true loss \eqref{terr}?  The latter
problem is tackled over the next two sections, while discussion of the
former is postponed until section \ref{EC}.

\section{Deviation of Empirical and True Loss}
\label{Dev}
Before the conditions ensuring a close agreement between 
the empirical loss and true
loss can be analysed, it first needs to be decided what ``close'' means in this
context. The empirical loss \eqref{eerr} and true loss 
\eqref{terr} are both positive numbers so 
deciding how to measure their difference is a matter of choosing a metric 
on $\R^+$. Haussler (1992) introduced 
the one parameter family of metrics, \glsname{d_nu}, defined as follows.
For all $x,y\in\R^+$ and all $\nu>0$, let 
$$
\d{x}{y} \de \frac{|x-y|}{\nu+x+y}.
$$
Note that the range of $d_\nu$ is $[0,1)$.  The advantage of this
metric over the usual Euclidean distance $d(x,y) = |x-y|$ is that it
is a relative, rather than absolute measure of distance, and so will
be more appropriate when $x$ and $y$ are large. However, note that
when either $x$ or $y$ is zero the condition $|x-y| < \ep$ is
equivalent to $d_\ep(x,y)<\frac12$ and so results on the deviation
between the true and empirical loss under the $d_\nu$ metric can be
translated into results under the normal Euclidean metric if the
empirical loss is known to be zero.  Some other useful properties of
the $d_\nu$ metric are given in lemma \ref{dnulem}.

It is now possible to define precisely what is meant by PC learning
within this machine learning context.
\begin{defn}
\label{PC}
A learning algorithm $\A\colon \cup_{m\geq 1}Z^m \to \H$ is {\em
  Probably Consistent} or {\em \glsname{PC}} with respect to the set
of probability measures $\P$ on $Z$ if for all $\nu>0$, $0<\alpha<1$,
$0<\delta<1$ and $P\in\P$, there exists a finite integer $N$ such that
$$
\Pr\left\{\zv\in\Zm\colon \d{\E{l_{\A(\zv)}}{P}}{\E{l_{\A(\zv)}}{\zv}} 
> \alpha\right\} < \delta
$$
for all $m\geq N$, where $\zv$ is generated by $m$ independent trials from
$Z$ according to $P$\jfootnote{It is being assumed that $\A$ is not so
pathological in its behaviour that the set in definition \ref{PC} is
unmeasurable}.
\end{defn}

To see how a learner $\A$ 
could fail to be PC, consider the situation in
which $\A$ simply rote-learns any sample it is given. That is, it produces a
hypothesis $h$ that functions by comparing any input $x$ with its training
examples $x_1,\dots,x_m$ and if it finds a match $x_i$ 
outputs the action $a$ minimizing $l(y_i,a)$. If it doesn't find a match it
simply chooses a random action. The behaviour of such a learner off the
sample, no matter how large the sample is, will clearly bear no relation to
its behaviour on the sample and so in general such a learner would not be PC. 
To prevent such a scenario the learner must
be provided with a restricted hypothesis space $\H$ from which to choose its
hypotheses. The next two sections are devoted to demonstrating exactly 
how the PC-ness of a learner is critically governed by the nature of $\H$.

\section{Bounding the Deviation}
\label{devbound}
Making no assumptions about the nature of $\A$ or $P$, the
probability in definition \ref{PC} can be bounded by:
\begin{multline}
\label{bb}
\Pr\left\{\zv\in\Zm\colon  \d{\E{l_{\A(\zv)}}{P}} {\E{l_{\A(\zv)}}{\zv}} > \alpha
\right\} \\
\leq \Pr\left\{\zv\in\Zm\colon \exists l_h\in l_\H\colon  \d{\E{l_h}{P}}{\E{l_h}{\zv}}
>\alpha \right \}.
\end{multline}

This bound is quite crude, as it is simply being said that the deviation
between the true and empirical loss on $\zv$ of the learner's 
hypothesis $\A(\zv)$ can't
be greater than $\alpha$ if there are in fact no functions in the hypothesis 
space $\H$ with such a deviation. This is effectively
a worst-case scenario; the inequality above becomes an equality if the learner
always chooses the hypothesis with the largest deviation between empirical 
and true loss. Clearly no learner will ever achieve this, except possibly
by chance, as the learner doesn't know the true loss and hence cannot bias its 
search towards regions of the hypothesis space with high deviation between 
empirical and true loss (not that it would want to in any case). 
Typically the learner is in fact stochastic: given
a sample $\zv$ the learner produces a {\em probability measure} $\A(\zv)$ on
the function space $\H$, and then selects a function $h\in\H$ according to
$\A(\zv)$. For example, in neural networks trained with simple gradient descent, the
learner starts with an initial distribution (gaussian, uniform, etc) over the
weights in the network, and then iteratively updates them in the direction
of steepest descent of the error function, which, in this
framework, is the empirical
loss of the network on the sample $\zv$. This process usually proceeds until
no further improvement in the error is seen i.e.\ the weights are at a local 
minimum in the error surface. Thus the network (hypothesis) chosen by 
the learner is selected from a distribution over the local minima in the error
surface for sample $\zv$, where the probability of a local minimum is the
probability of its attracting basin in weight space under the initial 
distribution on the weights. In the absence of any other information there is
no reason to suppose that the hypotheses with large deviation between 
empirical loss on $\zv$ and true loss are more concentrated in the local 
minima of the error surface than elsewhere in the 
surface. Hence, 
it can be assumed that the probability of the learner choosing a hypothesis with 
large deviation is simply equal to the measure (under the initial distribution
on the weights) of the hypotheses with large deviation. Unfortunately it is
not clear how to even begin calculating such a measure
and so, rather than weighting each sample 
$\zv$ by the measure of the region of weight space with high deviation 
between empirical loss on $\zv$ and true loss, it is weighted 
by one if the 
set of hypotheses with high deviation on $\zv$ is non-empty, 
and zero otherwise.

To state the main result bounding the right hand side of (\ref{bb}), some
definitions from pseudo-metric space theory are required. These are given in
definitions \ref{pseudodef}, \ref{pmetric} and \ref{capdef} but for
continuity are reproduced here.

Firstly a {\em pseudo-metric} is just a metric without the condition that 
points zero distance apart necessarily be identical. For the present
discussion the following pseudo-metric is significant.
\begin{defn}
\label{ppmetric}
Given any
probability measure $P$ on $Z$ and any space of functions $\F\colon
Z\to[0,M]$, define the pseudo-metric $d_P$ on $\F$ by
$$
\glsname{d_P}(f,g) \de \int_Z |f(z)-g(z)|\,dP(z),
$$
for all $f,g\in\F$. 
\end{defn}
Note that $d_P$ is only a pseudo-metric because $f$ and $g$ could
differ on a set of measure zero and still be equal under $d_P$.  The
pair \glsname{Fd_P} is called a {\em pseudo-metric space}. An {\em
  $\ep$-cover} for $\(\F,d_P\)$ is any subset $F\subseteq\F$ such that
for all $f\in\F$ there exists an $f'\in F$ such that
$d_P(f,f')\leq\ep$.  $G\subseteq\F$ is called an $\ep$-separated set
if $\rho(f,f') > \ep$ for all distinct $f,f'\in F$. Denote the size of
the smallest $\ep$-cover of $\(\F,d_P\)$ by \glsname{NepFd_P} and the
size of the largest $\ep$-separated subset by \glsname{MepFd_P}.
Define \glsname{sigma_F} to be the $\sigma$-algebra on $Z$ generated
by all inverse images under any $f\in\F$ of any Borel set in $[0,M]$.
Define \glsname{P_F} to be the set of all probability measures on
$(Z,\sigma_\F)$.  The {\em $\ep$-capacity} of $\F$ is defined by
$$
\glsname{CepF} \de \sup_{P\in\P_\F}\N\left(\ep,\F,d_P\right).
$$
If the supremum does not exist then the capacity is defined to be $\infty$.

The main theorem bounding the probability of deviation between true
and empirical loss in equation \eqref{bb} can now be stated.
\begin{thm}
\label{impthm}
Let $\H$ be a family of maps from $X$ into $A$ and $l$ be a loss function
$l\colon Y\times A\to [0,M]$. Assume that $\H$ and $l$ are such that $l_\H$ is
permissible\jfootnote{This relatively benign measure theoretic condition is
needed to ensure that sets such as the one appearing in the right hand side
of \eqref{bb} are measurable. The definition of permissibility is given in
appendix \ref{permissapp}.}.
Let $\zv$ be generated by $m$
independent draws from $Z$ according to any probability measure $P$. 
Then for all $\nu>0$, $0<\alpha<1$
$$
\Pr\left\{\zv\in\Zm\colon \exists l_h\in l_\H\colon  \d{\E{l_h}{P}}{\E{l_h}{\zv}}
>\alpha\right\}
\leq 4 \C(\alpha\nu/8,l_\H) e^{-\frac{\alpha^2\nu m}{8M}}.
$$
\end{thm}
This theorem is essentially due to Haussler (1992), Pollard (1984)\label{PE1} 
and others going back even further, but it has been 
derived in this work as corollary \ref{n=1} of a more general result (theorem
\ref{fundthm}) in section \ref{fundsec}.

Using inequality \ref{bb}, theorem \ref{impthm}
implies that to guarantee with probability $1-\delta$ a $d_\nu$ deviation of
less than $\alpha$ between a learner's  true and empirical loss, it
suffices to provide the learner with a sample of size $m$ where
\begin{equation} 
\label{lowerbound}
m > \frac{8M}{\alpha^2\nu}
\ln\left(\frac{4\C\(\alpha\nu/8,l_\H\)}{\delta}\right).
\end{equation}
Thus, as long as $\C\(\alpha\nu/8,l_\H\)<\infty$ the learner is guaranteed to
be PC. Thus the capacity of a hypothesis space $\H$ determines
critically PC-learnability using $\H$. However,
be aware that many approximations
go into establishing theorem \ref{impthm} (see the proof in appendix
\ref{fundapp}) and so in general 
the above bound on $m$ will not be tight.

\section{EC and its incompatibility with PC}
\label{EC}
PC learning ensures consistency between empirical estimates and true errors,
but does not have anything to say about exactly how good the learner's
performance is. This is covered by the {\em EC} criterion.

To begin with, for any probability measure $P$ on $Z$ define
\glsname{rP} to be the infimum of $\E{l_h}{P}$ over {\em all}
functions $h\colon X\to A$ such that $l_h$ is $P$-measurable.  Note
that the function $h$ need not come from the learner's hypothesis
space $\H$.
\begin{defn}
\label{ECdef} 
Let $\nu>0$ and $0<\alpha<1$ be two real parameters.  A learning
algorithm $\A\colon \cup_{m\geq 1}Z^m\to \H$ is {\em
  $(\alpha,\nu)$-Empirically Correct} or \glsname{anu} (or just ``EC''
for short) with respect to the set of probability measures $\P$ if for
all $P\in\P$ and $0<\delta<1$ there exists a positive integer $N$ such
that for all $m\geq N$,
$$
\Pr\{\zv\in Z^m\colon
\d{\E{l_{\A(\zv)}}{\zv}}{r^*(P)} > \alpha\} < \delta.
$$
\end{defn}
Thus for an algorithm $\A$ to be $(\alpha,\nu)$-EC it must, for sufficiently
large samples and with high probability, 
produce hypotheses with empirical loss
within $\alpha$ (under the $d_\nu$ metric) of the true loss of the 
{\em best possible hypothesis}.   
Requiring the learner to match closely the best possible
performance may seem somewhat harsh, as often in machine learning one 
requires only that the learner
approximates the performance of the best possible hypothesis chosen from
within its own hypothesis space $\H$. Clearly that requirement tells
nothing about how well the learner is actually doing---for example if the
learner was performing image recognition with a hypothesis space containing
only the hypothesis that ``everything is a picture of a dog'' then the learner
would always do brilliantly in its own terms but would be of no practical use.

The parameterisation $(\alpha,\nu)$ of the definition of EC allows 
a level of accuracy to be specified 
for the learner's hypotheses---the learner is not
necessarily required to produce arbitrarily good hypotheses, reflecting the
fact that often in practice approximate solutions are sufficient.

To show how EC and PC learning come into conflict ``Probably
Approximately Correct'' or PAC learning is defined, PC and EC together
are shown to imply PAC and then a simple but representative 
situation is given in which PAC implies 
not PC.

\begin{defn}
\label{PAC}
Let $\nu>0$ and $0<\alpha<1$ be two real parameters.  A learning
algorithm $\A\colon \cup_{m\geq 1}Z^m\to \H$ is {\em
  $(\alpha,\nu)$-Probably Approximately Correct} or \glsname{anuPAC}
or just PAC for short with respect to the set of probability measures
$\P$ if for all $P\in\P$ and all $0<\delta <1$, there exists a
positive integer $N$ such that for all $m\geq N$,
$$
\Pr\{\zv\in Z^m\colon
\d{\E{l_{\A(\zv)}}{P}}{r^*(P)} > \alpha\} < \delta.
$$
\end{defn}
So in PAC learning the learner is required to eventually start producing
hypotheses that have true loss arbitrarily close to the best possible. In
Valiant's (1984)\label{V3} original definition of PAC he essentially required that the
sample size $N$ grow at most polynomially in the relevant parameters
$1/\delta$, $1/\alpha$ and $1/\nu$. However, this approach conflates the
problems of {\em computational complexity} and {\em sample complexity} and
for the purposes of this work the issue of sample complexity is far more
important. Thus restrictions on the rate of growth of the sample size are
not given in this definition of PAC.

That
EC learning and PC learning together imply PAC is easily seen through the
triangle inequality for $d_\nu$. If the learner is PC then it is possible to 
sample sufficiently many times so that the probability of
$$
\d{\E{l_{\A(\zv)}}{\zv}}{\E{l_{\A(\zv)}}{P}} > \alpha/2
$$
is arbitrarily small. If the learner is $(\alpha/2,\nu)$-EC then
the probability
$$
\d{\E{l_{\A(\zv)}}{\zv}}{r^*(P)} > \alpha/2
$$
may also be made 
arbitrarily small and hence by the triangle inequality for $d_\nu$
the probability of 
$$
\d{\E{l_{\A(\zv)}}{P}}{r^*(P)} > \alpha,
$$
can be made arbitrarily small. Thus, PC and $(\alpha/2,\nu)$-EC together imply
$(\alpha,\nu)$-PAC. 

To satisfy the PAC condition the learner's hypothesis space must
contain $(\alpha/2,\nu)$ approximations to the best possible for any
$P\in\P$. The following argument shows how, at least for measurable functions
on $\R$ and for $\frac{\alpha\nu}{1-\alpha} \leq \frac18$, 
this is incompatible with PC learning.

Suppose the learner is trying to learn Borel measurable functions from $X=[0,1]$
into $Y=[0,1]$. Denote the set of all such functions by $\F$.
Assume the hypotheses it produces are also Borel measurable functions $[0,1]\to
[0,1]$, so that $A=Y=[0,1]$.
Denote as usual the
learner's hypothesis space by $\H$. Let the loss
function, $l\colon [0,1]\times [0,1]\to [0,1]$ be $l(y,y') = |y-y'|$ for all
$y,y'\in Y$.
Let $P$ be the uniform probability measure on $[0,1]$ and 
given $f\in\F$
denote by $P_f$ the probability measure induced on $\XY$ by generating
samples $(x,f(x))\in\XY$ with $x$ generated according to $P$. This is what
would normally be called ``sampling from $f$''. 
Let $\P_\F=\{P_f\colon f\in\F\}$.
By defining $d_P(f,h) = \int_X|f(x)-h(x)|\, dP(x)$, for all $P$-measurable $f$
and $h$, observe that 
\begin{align*}
\E{l_h}{P_f} &= \int\limits_{X\times Y} l_h(x,y)\, dP_f(x,y)\\
             &= \int_X l(h(x),f(x))\, dP(x) \\
	     &= \int_X |h(x) - f(x)|\, dP(x) \\
	     &= d_P(h,f).
\end{align*}
Note also that for all $P_f\in\P_\F$, $r^*(P_f) = 0$.

What is required of the learner's hypothesis space $\H$ to ensure 
$(\alpha,\nu)$-PAC learning? 
To start with, for all $P_f\in\P_\F$ there must exist at least one $h\in\H$
such that $\d{\E{l_h}{P_f}}{r^*(P_f)} \leq \alpha$. But as $r^*(P_f) = 0$,
$$
\d{\E{l_h}{P_f}}{r^*(P_f)}\leq\alpha\quad \Rightarrow \quad\E{l_h}{P_f} \leq
\frac{\alpha\nu}{1-\alpha}\quad \Rightarrow \quad d_P(h,f) \leq 
\frac{\alpha\nu}{1-\alpha}
$$
Set $\ep=\frac{\alpha\nu}{1-\alpha}$ and it has been 
shown that for $(\alpha,\nu)$-PAC
learning, $\H$ must be an $\ep$-cover for $\(\F, d_P\)$.
Now suppose $H$ is an $\ep$-cover of
$(\H,d_P)$. By
the triangle inequality for $d_P$ $H$ must also be 
a $2\ep$-cover for $\F$ and so $\C\(\ep,\H\)\geq \C\(2\ep, \F\)$. 

We now show that the capacity of $\F$ is infinite.
For all $n=1,2,\dots$ define the subset $I_n$ of $[0,1]$ by
$$
I_n \de \bigcup\limits_{m=1}^{2^{n-1}} 
\left[\frac{2m-1}{2^n},\frac{2m}{2^n}\right].
$$
For any $m,n\geq 0$ let $I_n\triangle I_m$ be the symmetric difference of
$I_n$ and $I_m$ (the set of points $x\in [0,1]$ such that $x\in I_n$ and
$x\not\in I_m$ or $x\in I_m$ and $x\not\in I_n$.) 
By abuse of notation denote the characteristic function 
of $I_n$ by $I_n$ also, and the characteristic function of $I_n\triangle I_m$ by 
$I_n\triangle I_m$.
Let $F=\{I_n\colon n=1,2,\dots\}$. Clearly all $f\in F$ are Borel
functions and so $F\subset\F$. For any $n,m\geq 1$ and $n\neq m$,
\begin{align*}
d_P\(I_n,I_m\) &= \int\limits_X \left|I_n(x) - I_m(x)\right|\,dP(x) \\
&= \int_0^1 \left|I_n\triangle I_m(x)\right|\,dx \\
&= \frac12.
\end{align*}
Thus $F$ is a $\frac12$-separated subset of $(\F,d_P)$ and so for all 
$\delta\leq \frac12$, as $|F| = \infty$,
$\M\(\delta,\F,d_P\) = \infty$ (recall that $\M\(\delta,\F,d_P\)$ is the size of
the largest $\delta$-separated subset of $\F$). So by lemma \ref{Jlem},
for all $\delta\le \frac14$, 
$\N\(\delta,\F,d_P\) = \infty$ and so $\C\(\delta,\F\) = \infty$. Hence
for all $\ep \leq \frac18$, $\C\(\ep,\H\)=\infty$. 

The fact that the learner's hypothesis space has inifinite capacity
does not necessarily mean the learner is not PAC. However this example 
demonstrates something stronger, namely that $\H$ contains a subset with 
infinite {\em VC-dimension}, from which one can easily conclude that there
exists consistent ({\em i.\ e.\ }
empirically correct) learners that are not PC, and 
hence not PAC (see for example \cite{AB}, chapter 8).

Is this example a realistic one? In a sense, the answer is no.
The space of all measurable functions is extremely large and contains many
functions that are very unlikely to be seen in practice, particularly the
more ``wiggly'' members of $F$ above. In fact by assuming that the functions
in $\F$ are limited in their wiggliness (perhaps by putting a bound on the
integral of the square of their second derivative) then it is fairly easy to
see that the capacity of $\F$ will be finite for all $\ep>0$, hence the
learner could get away with using a hypothesis space which is also
finite in capacity.
This is an example of using {\em prior information} of the kinds of
functions likely to be seen by the learner to turn a non-PAC problem into a
PAC one. However that is not the end of
the story. Assumptions about smoothness only serve to bias the learner's
hypothesis space in an absolutely minimal way.
The capacity of the learner's hypothesis space
would still have to be prohibitively large to ensure EC learning in all
situations and hence learning, although not technically
impossible, would still be infeasible. This is where the ``art'' of machine
learning comes in---it is the art of using prior knowledge about the 
problem domain
to further bias the hypothesis space so that sample sizes required for PAC
learning become managable. The problem with this approach is that it is an
``art'' and hence it is very much up to the skill and insights of the
researcher whether sufficiently many correct biases can be found.

In the next chapter it is shown how, at least for some kinds of learning
problems, it is possible to {\em learn} the bias by learning an appropriate
{\em representation} for the learner's environment.

\chapter{Representation Learning}
\label{repchap}
The previous chapter demonstrated that in the absence of any prior
information PAC learning is impossible. In this chapter the concept of the
{\em environment} of a learning process is defined and it is shown how the
environment may be sampled to automatically 
generate prior information for the learner.
Such information can be used to learn an appropriate {\em representation}
for the environment, which can then be used to bias the learner's hypothesis
space and greatly improve learning of
future tasks drawn from the same environment. 
Lower bounds for PC learning of representations are given, 
similar to the ones given for ordinary learning in the previous chapter.
These bounds show that for learning environments consisting of a large number
of similar tasks, such as character recognition and speech recognition, the
number of examples of each task required for PC learning can be greatly
reduced compared to 
that required for ordinary learning. In the final section the
results on representation learning are generalised to cover any procedure
for learning hypothesis space bias.

\section{General Framework}
\label{envlearn}
The ordinary learning scenario of the previous chapter 
is summarised by the following diagram
(``id'' just means the identity function).
\vspace{5mm}

\begin{center}
\begin{picture}(200,50)(20,40)
\thinlines
\put(-18,90){$\P\!\ni\!P$}
\put(5,87){\vector(0,-1){20}}
\put(0,57){$Z =\times \makebox[1.1cm]{} \times \xrightarrow{\ l\ } [0,M]$}
\put(23,70){$X \xrightarrow{h\in\H} A$}
\put(23,40){$Y \xrightarrow{\ \, \text{id}\ \,\, } Y$}
\end{picture}

\end{center}

If the learner receives no information about the kinds of tasks
$P\!\in\!\P$ it is required to learn, it must assume that $\P$ is the
most general possible and it was shown in the previous chapter how in
general this causes the learner to use a hypothesis space $\H$ so
large that PAC-learning using $\H$ is impossible. To enable the
learner to use a smaller hypothesis space it must be given some
information about $\P$. This can be modelled by placing a probability
measure \glsname{Q} on $\P$ so that when the
learner receives a task $P\!\in\!\P$ to learn, it is assumed to have
been drawn from $\P$ according to $Q$. $Q$ will be referred to as the
{\em environment} of the learning process.  For $Q$ to be well defined
there needs to be a $\sigma$-algebra on $\P$. The appropriate one to
use will become clear later in the chapter.

To enable the learner to get some idea of the environment in which it
is learning, and hopefully then extract some of the bias inherent in
the environment, it is provided not just with a single sample $\zv
=(\pr{z}{,}{m})$, sampled according to some probability measure
$P\!\in\!\P$, but with $n$ such samples $\(\pr{\zv}{,}{n}\)$. Each
sample $\zv_i=\(z_{i1},\dots,z_{im}\)$, for $\iton$, is generated by
first sampling from $\P$ according to $Q$ to generate $P_i$, and then
sampling $m$ times from $P_i$ to generate
$\zv_i=\(z_{i1},\dots,z_{im}\)$. Thus the entire sample is generated
by sampling $n$ times from $\P$ according to $Q$ to generate
$P_1,\dots,P_n$, and then sampling $m$ times from $Z$ according to
each $P_i$.  Denote the entire sample by \glsname{z} and write it as
an $n\times m$ ($n$ rows, $m$ columns) matrix over $Z$:
$$
\z =
\begin{matrix}
z_{11} & \hdots & z_{1m} \\
\vdots & \ddots & \vdots \\
z_{n1} & \hdots & z_{nm} 
\end{matrix}
$$ Denote the $n\times m$ matrices over $Z$ by \glsname{Znm} and call
a sample $\z\in\Znm$ generated by the above process an {\em $(n,m)$
  sample}.

To enable the learner to take advantage of
the prior information contained in $\z$,
the hypothesis space $\H\colon X\to A$ is split into 
two sections: $\H = \comp{\G}{\F}$ where $\F\colon X\to V$ and $\G\colon V\to A$, where 
$V$ is an arbitrary set\jfootnote{That is, $\H=\{\comp{g}{f}\colon g\in\G,
f\in\F\}$.}. 
To simplify the notation this will be written in future as  
$$
X \xrightarrow{\F} V \xrightarrow{\G} A.
$$
$\F$ is called the {\em representation space}
and an individual member $f$ of $\F$ is 
called a {\em representation}.
The representation learning scenario is summarised by the following diagram.
\vspace{5mm}

\begin{center}
\begin{picture}(200,70)(20,40)
\thinlines
\put(-18,110){$Q$}
\put(-13,106){\vector(0,-1){15}}
\put(-18,82){$\P\!\ni\! P$}
\put(5,81){\vector(0,-1){15}}
\put(0,57){$Z =\,\,\times\phantom{\xrightarrow{h\in\comp{\G}{\F}}}
\,\,\,\,\,\times\; \xrightarrow{\;\;l\;\;} [0,M]$}
\put(26,70){$X \xrightarrow{h\in\comp{\G}{\F}} A$}
\put(27,40){$Y \xrightarrow{\;\;\;\;{\text{id}}\;\;\;\;\,} Y$}
\end{picture}

\end{center}

Based on the information about the environment $Q$, contained in $\z$,
the learner searches for a good representation $f\in\F$. A good
representation is one with a small {\em empirical loss} \glsname{EGfz}
on $\z$, where this is defined by
\begin{equation}
\label{emploss}
E^*_\G(f,\z) \de \frac1n\sum_{i=1}^n \inf_{g\in\G}\E{l_\comp{g}{f}}{\zv_i},
\end{equation}
where $\zv_i = \(z_{i1},\dots,z_{im}\)$ denotes the $i$th row of $\z$.
The empirical loss of $f$ with respect to $\z\in\Znm$ is a 
measure of how well the learner can learn $\z$ 
using $f$, assuming that the learner 
is able to find the best possible $g\in\G$ for any given sample $\zv\in\Zm$. 
For example, if the empirical loss
of $f$ on $\z=(\zv_1,\dots,\zv_n)$ is zero then it is possible for the 
learner to find a function 
$g_i\in\G$, for each $i$, $1\leq i\leq n$, such that the ordinary empirical
loss, $\E{l_\comp{g_i}{f}}{\zv_i}$, is zero. 
The empirical
loss of $f$ is an estimate of the {\em true loss} of $f$
(with respect to $Q$), where this is defined by
\begin{equation}
\label{exploss}
\glsname{EGfQ} \de \int_\P \inf_{g\in\G} \E{l_\comp{g}{f}}{P}\, dQ(P).
\end{equation}
The true loss of $f$ with respect to $Q$ is the expected best possible
performance of $\comp{g}{f}$---over all 
$g\in\G$---on a distribution chosen at random
from $\P$ according to $Q$. If $f$ has a small true loss then learning using
$f$ on a random ``task'' $P$, drawn according to $Q$, will with high
probability be successful. 

Once the learner has found the best 
representation it can with respect to the sample $\z$, it then learns using 
the {\em restricted} 
hypothesis space $\comp{\G}{f} \de \{\comp{g}{f}\colon g\in\G\}$. That is,
the learner will be fed samples $\zv\in\Zm$ drawn according to some distribution
$P\in\P$, which in turn is drawn according to $Q$, and will search 
$\comp{\G}{f}$ for a hypothesis $\comp{g}{f}$ with small empirical loss on
$\zv$. 
Intuitively, if the capacity of $\F$ is much greater than the capacity of
$\G$, then 
the number of samples required to learn using $\comp{\G}{f}$ will be much less 
than the number of samples required to learn using the full space $\comp{\G}{\F}$,
a fact that is proved later in this chapter. Hence, if the learner can find a
good representation $f$ and the sample $\z$ is large enough,
learning using $\comp{\G}{f}$ should be considerably quicker and more
reliable than learning using $\comp{\G}{\F}$.

\section{Motivation}
Do the environments encountered in practice possess representations? In many 
important cases it is quite plausible that they do.
Take for example the problem of character recognition. It seems a plausible
hypothesis that the kinds of transformations one can apply to characters and
still leave them essentially unchanged (i.e still recognisably the same
character) are the same, regardless of the character concerned (a simple
example is small rotations of the character). 
If this is true then there exists a representation $f$ mapping the space of
images of characters $X$ into some other space $V$ such that if a
``character invariant transformation'' is applied to character $x$, 
the value of
$f(x)$ will remain ``relatively unchanged''\jfootnote{This statement only makes
sense if there is some kind of distance measure for comparing elements of $V$. 
It is shown in chapter \ref{flearnchap} how
one may be induced, by ``pulling back'' 
any distance measure on $\R$, through the loss function $l$ and then through
$\G$.}. The same idea applies to speech recognition. It seems individual
words are invariant under the same kinds of transformations and so a
representation should exist that is insensisitive to such transformations. 
Recognizing faces is yet another example (see in this context the
interesting paper by Weiss and Edelman (1993)\label{E1}). At any rate, the
question of whether a useful representation exists in these cases is an
empirical one---it can be verified either by examining the brains of
biological learners or by attempting to learn a representation for the
environments concerned.
In the next section it is demonstrated how that task is greatly facilitated
by a result concerning the sample complexity of
representation learning.

If the above environments do indeed possess representations, then
there is also strong evidence that the space $\G$ is quite small because
humans are able to learn to recognise spoken words and written characters
from a relatively small number of examples. 
The ability of humans to learn faces from
just one example is evidence of an even smaller $\G$ for this task. Hence
learning of an appropriate representation in all these environments is
likely, by the argument at the end of the previous section, to greatly
enhance learning of future tasks from the same environment.

\section{The Advantage of Learning Many Tasks}
\label{advantage}
In ordinary learning the learner would be learning one ``task'' $P$
from the
environment $\P$ and in the absence of any prior information would be using
the full hypothesis space $\comp{\G}{\F}$. If $\comp{\G}{\F}$ is an
appropriate hypothesis space for the environment then it will contain a
reasonable, say $(\alpha/2,\nu)$-approximation to the best achievable on the
task $P$, and thus as long
as the learner can find such a hypothesis it will be $(\alpha/2,\nu)$-EC.
By the argument following the definition of PAC-learning in the previous
chapter, the learner will be $(\alpha,\nu)$-PAC if it also samples sufficiently
many times to ensure that it is PC. By equation \eqref{lowerbound},
page~\pageref{lowerbound}, this
will be fulfilled if the learner samples 
$$
m = O\(\log\C\(\ep,l_\comp{\G}{\F}\)\) 
$$
times from $Z$ according to $P$, where
$\ep=\alpha\nu/8$ and only the capacity dependence of the
bound has been written. In section \ref{compsec} it is proved that the capacity
of a composition is no more than something 
proportional to the product of the capacities of the
individual spaces,
hence (loosely speaking---this will all be made rigorous shortly),
\begin{equation}
\label{gnbound}
m = O\(\log\C\(\ep,\G\)+\log\C\(\ep,\F\)\).
\end{equation}
In contrast, suppose instead the learner has received $n$ tasks $P_1,\dots,
P_n$ drawn from $\P$ according to the environment $Q$, 
and is learning a representation $f$ 
appropriate to the $n$ tasks, then it turns out that the
learner need only sample 
\begin{equation}
\label{gnfbound}
m = O\(\log\C\(\ep,\G\) + \frac{\log\C\(\ep,\F\)}{n}\)
\end{equation}
times from each task to be $(\alpha,\nu)$-PAC (take the empirical and
true losses of the learner on the $n$ tasks to be the 
the average of the ordinary 
empirical and true losses on each individual task and
require for PC learning that these averages
be close with high probability. Once again
rigorous definitions are given shortly). 

Thus, if the capacity of $\F$ is much greater than the capacity of
$\G$,
then by learning $n$ tasks instead of one  
there will potentially be an {\em $n$-fold reduction 
in the number of examples required
per task for PAC learning}. This suggests a new approach to machine
learning. Instead of only learning the task required, learn as many related
tasks as possible. For instance, if a neural network is being trained to 
recognise handwritten characters then instead of only teaching it to
recognise the roman alphabet, teach it kanji characters, and
hiragana and arabic and the digits and so on.
Many fewer examples of each character will be
required to learn the lot, and if the number of characters is sufficiently
great (greater than $O\(\log\C\(\ep,\F\)\)$, as will be shown later)
then the resulting
representation will be a good one for learning new characters. In
fact learning the kanji characters would be a very good test of this theory
because there are enough of them (thousands) to ensure with high
probability that the representation will be a good one, and to have plenty 
left over for testing the ease of learning using the representation on
unseen characters. 
Note that this is a new form of ``generalisation'', one
level of abstraction higher than the usual meaning of generalisation, for
within this framework a
learner generalises well if, after having learnt many different characters,
it is able to learn new characters easily. Thus, not only is the learner 
required to generalise well in the ordinary sense by generalising well on
the characters in the training set,
but also the learner is expected to have ``learnt to learn'' characters in 
general. 

In a similar way, if a representation exists that is appropriate for spoken
word recognition then the learner should be trained on many words,
perhaps on many languages too, so that the required number of examples of each word
is greatly reduced and so that the resulting representation may be
useful in learning new words. Character recognition and speech
recognition are but two of many problems that should lend themselves to
representation learning.

In chapter \ref{expchap} a gradient-based algorithm for training 
neural networks to learn
representations is given and the results of some simple experiments presented
that fully support the claims made here for the efficacy of
representation learning. The remainder of this chapter is devoted to a
theoretical justification for these claims, and to 
providing rigorous lower bounds on the
number of examples required for PAC-like learning of representations.

\section{Philosophical Aside}
As with ordinary learning, the two main issues 
involved in representation learning 
are whether the learner can find a representation
with small empirical loss \eqref{emploss} on a given $(n,m)$ sample, 
and whether the empirical 
loss is in fact close to the true loss \eqref{exploss}. 
This is the EC/PC question again and a similar argument to the one given at
the end of chapter \ref{ordchap} can be used to show that EC and PC are
also incompatible for representation learning, in the absence of any
prior information about the kinds of environments $Q$ that the
representation learner is likely to be operating within. 
Using the same reasoning that led to the introduction of representation
learning (or at least automated hypothesis space bias) in the first place,
it would be tempting to suppose that $Q$ is drawn from a collection $\Q$ of
environments according to some {\em super-environmental measure} $R$, and
attempt to generate prior information by sampling from $\Q$ according to
$R$.
Unfortunately, the same arguments would show PC and EC learning to still be
incompatible at this new level of abstraction, and so one would be forced to
go higher and higher in a never-ending process. 
The conclusion is that no finite
sample, regardless of whether it is used for representation learning,
ordinary learning, or super-environmental learning or whatever, will ever 
be big enough to solve the PC/EC dilemma. This conclusion should not really be
surprising, it is equivalent to the fact that there is no {\em a priori}
basis for induction, no matter what the circumstance. However this problem
is basic to all of science and philosophy, and hence failing to solve it is
no reason for discarding a particular line of enquiry. 

In fact there are
``superselection'' rules, such as {\em Occam's Razor}, that, although they are 
{\em a priori} unjustified, do seem to work in practice as a means of
breaking the induction dilemma. The modern version of Occam's Razor is 
{\em Kolmogorov Complexity}. The Kolmogorov Complexity of an object is
essentially the length (in bits) of the shortest program that can be fed
into a universal computer causing it to produce a description of
the object as output\jfootnote{See Cover and Thomas (1991), chapter
12\label{CT1}, for a
highly readable discussion of Kolmogorov Complexity.}. It is trivial to show,
up to an additive constant, that the Kolmogorov Complexity of an object is
not affected by the choice of universal computer used to describe it. Thus
Kolmogorov Complexity can be used to compare the complexity of two solutions
to a particular problem. For example, it could be used to choose between two
candidate representations, both with zero empirical loss on a training set
$\z$. In general a representation that simply rote-learns $\z$ will have far
higher complexity than one which takes advantage of regularities in the data,
and hence will be rejected by this criterion (as desired). A general
procedure for learning representations based on complexity is to always
search among those representations with small complexity first. If no
solution can be found, search amongst those representations with higher
complexity and so on until a solution is found. This is equivalent to
providing the learner with a sequence of representation spaces
$\F_1\subset\F_2\subset\F_3\subset\dots$, which in the limit would include
all possible representations\jfootnote{At least all computable ones, and its
not worth bothering about the problem of environments that are generated by
non-computable representations because no solution could ever be computed
for them, no matter what learning procedure was used.}, and hence in the
limit the learner would be guaranteed to be EC\jfootnote{Such a procedure
is closely related to what Vapnik (1982)\label{Vap1} calls {\em structural risk
minimization}.}\jfootnote{The fact that Kolmogorov complexity is not
computable prevents this procedure from being carried to its ultimate limit.
However the necessary finiteness of any real-world computational procedure
means that the limit can never be reached anyway.}. Of course, if the sequence
continues too far the capacity of the 
representation space $\F_n$ will become too big to ensure PC learning and
at that stage the learner would have no choice but to collect further
samples from the environment. However the learner is guaranteed to
eventually find a solution by this procedure.

The obvious question now is why not do away with representation learning
altogether and just apply the same complexity criterion to ordinary
learning? There are two reasons why not. Firstly, ordinary learning is of no
use if the goal is to find a
representation that can be used to improve learning performance on further
tasks drawn from the same environment. Secondly, by sampling from only one task
in an environment containing many similar tasks, the ordinary learner 
ignores information that is freely available and although it is true
that eventually the learner would find a solution to the single task,
convergence would be greatly improved if the information from the other
tasks was incorporated as well. More succinctly put, if the environment
posesses a representation then the learner should not ignore that fact.
Ordinary learning procedures are essentially indifferent to the existence of
a representation in the environment. 

Of course much more work needs to be done to turn these vague ideas into
practical algorithms, but at least {\em in principle} they show that the
PC/EC dilemma is not utterly intractable (provided that ``simplicity'' is
truly a superselection criterion applicable in this universe).   
So for the remainder of this work it will be 
assumed that the 
hypothesis space $\comp{\G}{\F}$ has been chosen so that EC representation
learning is possible, and attention focused instead on the problem
of quantifying the conditions necessary for PC learning.

\section{Bounds for PC Representation Learning}
There are two ways that the generalisation performance of a
representation learner can be measured. The first is how well the
representation performs in practice, as measured by expression
\eqref{exploss}, page~\pageref{exploss}, and the second is how well the
learner generalises only on 
the $n$ tasks from its $(n,m)$ sample. The analysis of both these problems
is performed in this section, the latter leading to the relation
\eqref{gnfbound} on page~\pageref{gnfbound}. 

However, before proceeding, a slight 
complication with the form of 
the empirical loss \eqref{emploss} needs to be discussed, and
some preliminary definitions must also be introduced.

\label{form}
\subsection{Computing the Empirical Loss}
\label{cel}
For the learner to compute \eqref{emploss} it must be able to compute
$\inf_{g\in\G} \E{l_\comp{g}{f}}{\zv}$. If the infimum is not attained in $\G$
for the sample $\zv$ and representation $f$ 
then it is unrealistic to assume that the learner can determine the infimum,
and even if the infimum is attained it is unreasonable to force the learner
to find it. In fact, for EC learning the learner is only ever required 
to get $(\alpha,\nu)$-close to the best possible. Thus, although for
mathematical simplicity definitions \eqref{emploss} and
\eqref{exploss} are used, 
the results presented here can easily be extended to cover
the case in which the learner is only required to find on any sample
$\zv\in\Zm$ a function 
$g^*_\zv$ satisfying
$$
\d{\E{l_\comp{g^*_\zv}{f}}{\zv}}{\inf_{g\in\G} \E{l_\comp{g}{f}}{\zv}} <
\alpha,
$$
for some $\nu>0$, $0<\alpha<1$.

\subsection{Preliminary Definitions}
\label{prelimdef}
\sloppy In searching for a representation minimising the empirical
loss \eqref{emploss}, it is assumed that the learner generates a
sequence of $n$ functions $(\comp{g_1}{f},\dots,\comp{g_n}{f})$, where
each $\comp{g_i}{f}$ is a map from $X$ into $A$ minimizing the
empirical loss $\E{l_\comp{g_i}{f}}{\zv_i}$.  In what follows the loss
of such sequences with respect to $(n,m)$-samples $\z\in\Znm$ needs to
be defined, and also the loss with respect to sequences of
distributions $\Pv=\(\pr{P}{,}{n}\)$.  Toward this end the following
general definitions are made. Given a sequence of functions
$\hv=\(\pr{h}{,}{n}\)$, with $h_i\colon X\to A$, and a sample
$\z\in\Znm$, let $\zv_i = \(z_{i1},\dots,z_{im}\)$ denote the $i$th
row of $\z$ and define the {\em empirical loss of $\hv$ with respect
  to $\z$} by
\begin{equation}
\label{empEloss}
\glsname{Ehvz} \de \frac{1}{n}\sum_{i=1}^n \E{l_{h_i}}{\zv_i} =
\frac1{mn}\sum_{i=1}^n\sum_{j=1}^m l_{h_i}(z_{ij}).
\end{equation}
This is just the average over the $n$ training sets $\zv_1,\dots,
\zv_n$ of the individual empirical losses of the hypotheses
$h_1,\dots,h_n$.  Given a sequence of probability measures
$\Pv=\(\pr{P}{,}{n}\)$, define the {\em true loss of $\hv$ with
  respect to $\Pv$} by
\begin{equation}
\label{trueEloss}
\glsname{EhvPv} \de \frac1n\sum_{i=1}^n \E{l_{h_i}}{P_i}.
\end{equation}
This is just the average of the $n$ {\em true losses} of the hypotheses
$h_1,\dots,h_n$.
\fussy

Given a family of functions $\H\colon X\to A$, define $\H^n\colon
X^n\to A^n$ to be the $n$-fold Cartesian product of $\H$,
$\glsname{Hn} = \{\(h_1,\dots,h_n\)\colon h_i\in \H, \iton\}$ where
$(h_1,\dots,h_n)\linebreak (x_1,\dots,x_n) =
(h_1(x_1),\dots,h_n(x_n))$. Distinguish a special subset of $\H^n$,
consisting of all functions from $X^n$ into $A^n$ of the form
$(h,\dots,h)$ and denote this subset by \glsname{Hbar}. For all
$h\in\H$ set $\bar{h}\de (h,\dots,h)$.  Given
$\gv=(g_1,\dots,g_n)\colon V^n\to A^n \in\G^n$ and
$\fv=(f_1,\dots,f_n)\colon X^n\to V^n \in\F^n$, set
$\comp{\gv}{\fv}\de \(\comp{g_1}{f_1},\dots,\comp{g_n}{f_n}\)$.

With this notation the set of all sequences of the form
$\(\comp{g_1}{f},\dots,\comp{g_n}{f}\)$ where $g_i\in\G$ and $f\in \F$
is denoted by \glsname{GnFbar}. Such sequences will be written
\glsname{gvfbar}.

In addition to the notion of the empirical loss of $f$ with respect to sample
$\z$ (\ref{emploss}), and the true loss of $f$ with respect to the environment
$Q$
(\ref{exploss}), 
the notion of the loss of $f$ with respect to a
sequence of distributions $\Pv=\(\pr{P}{,}{n}\)$ needs to be
defined:
\begin{equation}
\label{pvloss}
\glsname{EGfPv} \de \frac1n\sum_{i=1}^n \inf_{g\in\G}
\E{l_\comp{g}{f}}{P_i}.
\end{equation}

The learner takes as input 
samples $\zv\in\Znm$, for any values $m,n\ge 1$, and produces as output
hypothesis representations $f\in\F$, so it is a map $\A$ from the
space of all possible $(n,m)$ samples into $\F$,
$$
\A\colon  \bigcup_{\substack{n\geq 1\\ m\geq 1}} \Znm \to \F.
$$
As with ordinary learning, the results developed here can be easily
generalised to cover stochastic learners.

\subsection{Generalisation on {\em n} tasks}
\label{gnfsec}
To learn a representation for the environment $Q$, the learner first samples
$n$ times from $\P$ according to $Q$ to generate $\Pv=\(P_1,\dots,P_n\)$,
and then samples $m$ times from $Z$ according to each $P_i$, $\iton$, to
generate the $(n,m)$ sample 
$$
\z=
\begin{matrix} 
z_{11} & \hdots & z_{1m} \\ 
\vdots & \ddots & \vdots \\ 
z_{n1} & \hdots & z_{nm} 
\end{matrix}
$$
The learner then searches for $n$ hypotheses from $\comp{\G}{\F}$,
$\comp{\gv}{\fbar} = \(\comp{g_1}{f},\dots,\comp{g_n}{f}\)$, all using the same representation
$f$ and all with empirical loss $\E{l_\comp{g_i}{f}}{\zv_i}$ as small as
possible, where $\zv_i$ is the $i$th row of $\z$. 
The average empirical loss of 
the learner on the $(n,m)$ sample $\z$ is then $E\(\comp{\gv}{\fbar},\z\)$
where this is defined as in equation \eqref{empEloss}, and the average true
loss is $E(\comp{\gv}{\fbar},\Pv)$ where this is defined as in equation
\eqref{trueEloss}. It may be that the $n$ tasks learnt are all that is ever
going to be required of the learner (for example the $n$ tasks could include
all the characters ever to be recognised, or all the words ever to be
recognised, etc), i.e the $n$ tasks entirely exhaust the environment of the
learner, in which case it is not the performance of the representation $f$ on
future tasks that is of interest, but the performance of each
$\comp{g_i}{f}$ on future examples of the same task. Good generalisation will be
assured if the sample size is large enough to ensure with high probability 
that $E(\comp{\gv}{\fbar},\Pv)$ and $E(\comp{\gv}{\fbar},\z)$ are close.
Following the treatment in chapter \ref{ordchap}, given any 
$\nu>0$ and $0<\alpha<1$, a bound on the probability that 
$$
\d{E(\comp{\gv}{\fbar},\z)}{E(\comp{\gv}{\fbar},\Pv)} > \alpha
$$
is required.
Clearly the probability of this event is bounded by 
\begin{equation}
\label{gnfprob}
\Pr\left\{\z\in \Znm\colon  \exists \comp{\gv}{\fbar}\in \comp{\G^n}{\Fbar}\colon 
\d{E(\comp{\gv}{\fbar},\z)}{E(\comp{\gv}{\fbar},\Pv)}
>\alpha\right\},
\end{equation}
where the probability measure on $\Znm$ is the product measure
$P_1^m\times\dots\times P_n^m$ generated by $\Pv=\(P_1,\dots,P_n\)$.
With a little work and some more definitions, expression \eqref{gnfprob} may
be converted into a form that can be bounded using essentially the same
techniques used in the previous chapter.

\sloppy
If $\Zmn$, the set of all $m\times n$ matrices over $Z$, is equipped with the
measure $\(P_1\times\dots\times P_n\)^m$ (i.e. each {\em row} of a matrix is
selected according to $P_1\times\dots\times P_n$), then the map 
$\psi\colon \Znm\to\Zmn$, $\psi(\z) = \z^T$
where $\z^T$ is the transpose of $\z$ is clearly measure preserving if the
measure on $\Znm$ is $P_1^m\times\dots\times P_n^m$.
Thus the sampling process on $\Znm$ 
according to $P_1^m\times\dots\times P_n^m$ is equivalent 
to sampling from $\Zmn$ according to $(P_1\times\dots\times P_n)^m$, which 
is the same as sampling $m$ times from $\Zn$ according to 
$P_1\times\dots\times P_n$.
This observation, along with a suitable generalization of the loss function
$l$ to $n$ dimensions, allows $E(\comp{\gv}{\fbar},\Pv)$ and
$E(\comp{\gv}{\fbar},\z)$ 
to be interpreted as expectations of loss functions, $l_\comp{\gv}{\fbar}$, 
and hence turns
the probability in equation \eqref{gnfprob} above into a statement about the
probability of large deviation between true and empirical losses.

\fussy
Given a loss function $l\colon Y\times A\to[0,M]$, define $l\colon Y^n\times A^n\to[0,M]$
by 
$$
l(\yv,\av) \de \frac1n\sum_{i=1}^n l(y_i,a_i).
$$
By identifying $X^n\times Y^n$ with $(\XY)^n$ in the obvious
way\jfootnote{$(x_1,\dots,x_n,y_1,\dots,y_n)\leftrightarrow
(x_1,y_1,\dots,x_n,y_n)$.}, and
recalling that $Z=\XY$ so $\Zn = (\XY)^n \equiv X^n\times Y^n$ by this
identification, 
for any function $h\colon X^n\to A^n$, a function $l_h\colon Z^n\to [0,M]$ may be defined 
as follows
$$
l_h(\zv) \de l(\yv,h(\xv)),
$$
for all $\zv\in\Zn, \zv=(\xv,\yv), \xv\in X^n, \yv\in Y^n$. 
If $\H$ is a family of functions from $X^n$ into $A^n$ then denote by $l_\H$
the set $\{l_h\colon h\in\H\}$. 
Given a sequence of probability measures $\Pv=\(\pr{P}{,}{n}\)$, by abuse
of notation denote the product measure $\pr{P}{\times}{n}$ on $\Zn$ by $\Pv$ 
also, and define the 
{\em expectation with respect to $\Pv$} of any function $f\colon \Zn\to [0,M]$ in
the usual way,
$$
\glsname{EfPv} \de \int_\Zn f(\zv)\, d\Pv(\zv).
$$
Define the expectation of $f\colon \Zn\to[0,M]$ with respect to the $m$-fold
sample from $\Zn$,
$\z=\{\zv_1,\dots,\zv_m\} \equiv \begin{smallmatrix} z_{11} & \hdots & z_{1n} \\ \vdots & \ddots & \vdots
\\ z_{m1} & \hdots & z_{mn} \end{smallmatrix}\in\Zmn$ as
$$
\glsname{Efz} \de \frac1m\sum_{i=1}^m f\(\zv_i\).
$$

\sloppy
\begin{defn}
\label{newcap}
For any probability measure $P$ on $\Zn$,
and any family of functions $\H\colon  Z^n\to[0,M]$, 
a pseudo-metric $d_P$ on $\H$ can be defined as in section
\ref{devbound},
$$
d_P(h,h') \de \int_{\Zn} |h(\zv) - h'(\zv)|\, dP(\zv),
$$
for all $h,h'\in\H$. The $\ep$-capacity of $\H$, $\C\(\ep,\H\)$, is then also defined as in
\S\ref{devbound}, namely as the supremum of $\N\(\ep,\H,d_P\)$ over 
all probability measures $P$ such that all $h\in\H$ are $P$-measurable.
$$
\C\(\ep,\H\) \de \sup_{P\in\P_\H} \N\(\ep,\H,d_P\).
$$
\end{defn}

\fussy
If $h\colon X^n\to A^n$ is expressible as a sequence $h=\(\pr{h}{,}{n}\)$, where 
$h_i\colon X\to A$
(that is, $h$ does not mix the components of its argument), and
$\hat{\z}\in\Zmn$
then the expectation of $l_h$ with respect to
$\hat{\z}$ satisfies 
\begin{align*}
\E{l_h}{\hat{\z}}  &= \frac1m \sum_{i=1}^m l_h\bigl(\vec{\zh}_i\bigr) \\
                   &= \frac1{mn}\sum_{i=1}^m\sum_{j=1}^n l_{h_j}(\zh_{ij}) \\
		   &= E(h,\hat{\z}^T)
\end{align*}
(recall equation \eqref{empEloss}, page~\pageref{empEloss}).
Also, for $\Pv=P_1\times\dots \times P_n$,
\begin{align*}
\E{l_h}{\Pv} &= \int_\Zn l_h(\zv)\, d\Pv(\zv) \\
             &= \int_\Zn \frac1n\sum_{i=1}^n l_{h_i}(z_i)\, dP_1(z_1)\dots
dP_n(z_n) \\
             &= \frac1n \sum_{i=1}^n \int_Z l_{h_i}(z)\, dP_i(z) \\
             &= E(h,\Pv)
\end{align*}
(recall equation \eqref{trueEloss}, page~\pageref{trueEloss}), 
where by abuse of notation $\(P_1,\dots,P_n\)$ has been denoted by $\Pv$ also.
Thus \eqref{gnfprob} can be rewritten as, 
\begin{multline}
\label{transprob}
\Pr\left\{\z\in \Znm\colon  \exists \comp{\gv}{\fbar}\in \comp{\G^n}{\Fbar}\colon 
\d{E(\comp{\gv}{\fbar},\z)}{E(\comp{\gv}{\fbar},\Pv)} >\alpha\right\} \\
= \Pr\left\{\z\in\Zmn\colon  \exists l_\comp{\gv}{\fbar}\in l_\comp{\G^n}{\Fbar}\colon 
\d{\E{l_\comp{\gv}{\fbar}}{\z}}{\E{l_\comp{\gv}{\fbar}}{\Pv}} 
>\alpha\right\}.
\end{multline}

$l_\comp{\G^n}{\Fbar}$ is a subset of $l_{\(\comp{\G}{\F}\)^n}$ which can in
turn be written as $l_\comp{\G}{\F}\oplus\dots\oplus l_\comp{\G}{\F}$ as
defined in \ref{htimes}. Therefore, assuming $l_\comp{\G^n}{\Fbar}$ 
is permissible\jfootnote{The conditions ensuring permissibility of 
$l_\comp{\G^n}{\Fbar}$ are given in appendix \ref{permissapp}. However the
motivation for the definitions given there and their relationship to the
current framework is best understood after reading section \ref{genframe} of
the current chapter.}
the conditions of theorem \ref{fundthm} obtain for 
the probability in the right hand side of \eqref{transprob}, hence,
\begin{multline*}
\Pr\left\{\z\in\Zmn\colon  \exists l_\comp{\gv}{\fbar}\in l_\comp{\G^n}{\Fbar}\colon 
\d{\E{l_\comp{\gv}{\fbar}}{\z}}{\E{l_\comp{\gv}{\fbar}}{\Pv}} 
>\alpha\right\} \\
\leq 4 \C\(\alpha\nu/8,l_\comp{\G^n}{\Fbar}\) e^{-\frac{\alpha^2\nu nm}{8M}}.
\end{multline*}
By theorem \ref{compgnf},
$$
\C\(\alpha\nu/8,l_\comp{\G^n}{\Fbar}\) \leq \C\(\ep_1,l_\G\)^n
\C^*_{l_\G}\(\ep_2,\F\),
$$
where $\ep_1+\ep_2 = \frac{\alpha\nu}{8}$ and the capacities in the right
hand side are defined as in definitions \ref{capdef} and \ref{dbardef}.
Thus, if the learner samples 
\begin{equation}
\label{gnfmbound}
m > \frac{8M}{\alpha^2\nu}\left[\ln \C\(\ep_1,l_\G\) + \frac1n
\ln\frac{4 \C^*_{l_\G}\(\ep_2,\F\)}{\delta}\right]
\end{equation}
times from $Z$ according to each distribution $P_i$, $\iton$, then
with probability at least $1-\delta$ the $d_\nu$ distance between the true
and empirical loss of the learner will be less than $\alpha$.
Ignoring the accuracy parameters $\alpha,\nu$ and $\delta$,
$$
m = O\(\log \C\(\ep_1,l_\G\) + \frac1n \log\C^*_{l_\G}\(\ep_2,\F\)\),
$$
which is the rigorous version of the order of magnitude estimate
\eqref{gnfbound} on page~\pageref{gnfbound}.

Exactly the same argument with $n=1$ yields a number of samples sufficient
for learning a single task $P$ using the hypothesis space $\comp{\G}{\F}$ of,
$$
m > \frac{8M}{\alpha^2\nu}\left[\ln \C\(\ep_1,l_\G\) +
\ln\frac{4 \C^*_{l_\G}\(\ep_2,\F\)}{\delta}\right].
$$
This gives an order of magnitude estimate,
$$
m = O\(\log \C\(\ep_1,l_\G\) + \log\C^*_{l_\G}\(\ep_2,\F\)\),
$$ 
which is the rigorous version of \eqref{gnbound} on page~\pageref{gnbound}.

Thus if $\C^*_{l_\G}\(\ep_2,\F\) \gg \C\(\ep_1,l_\G\)$ then there is a virtually
linear (in $n$) reduction in the number of examples per task required for
representation learning as compared with ordinary learning of a single task.

In appendix \ref{nnetapp}, using the same methods as Haussler (1992), it is
shown for neural network $\F$ and $\G$ that the logarithm of the capacities
$\C^*_{l_\G}(\ep_2,\F)$ and $\C\(\ep_1,l_\G\)$ are essentially 
proportional to the number
of weights in each network. Hence the advantage gained by representation
learning using neural networks is essentially proportional to the ratio of
the size of the networks in $\F$ and the networks in $\G$.

In the next section, sufficient sampling conditions are derived 
such that a representation with small empirical loss \eqref{emploss} on an
$(n,m)$ sample also has a small true loss \eqref{exploss}, i.e. it is a
good representation for learning future tasks drawn from the environment.

\subsection{Generalisation on all tasks.}
\label{alltasks}
In representation learning, given a sample $\z\in\Znm$,
the learner attempts to produce a
representation $f\in\F$  with as small an empirical loss \eqref{emploss}
on the sample $\z$ as possible. The empirical loss is an estimate of how
well the learner can be expected to learn using $f$. To be confident that
the empirical loss is a good estimate of the true loss \eqref{exploss}, the
probability 
\begin{equation}
\label{mainprob}
\Pr\left\{\z\in \Znm\colon  
\d{E^*_\G(\A(\z),\z)}{E^*_\G(\A(\z),Q)}
> \alpha \right\}
\end{equation}
must be bounded,
where the samples $\z$ are generated according to the $(n,m)$ sampling
process\jfootnote{The full expression for this probability is:
$$
\int_{\P^n}\int_\Znm
\theta\(\d{E^*_\G(\A(\z),\z)}{E^*_\G(\A(\z),Q)} - \alpha\)\,
dP_1^m\times\dots\times dP_n^m(\z)\,dQ^n\(P_1,\dots,P_n\),
$$
where $\theta$ is the Heaviside step function.}.
Note that with the $(n,m)$ sampling process, in addition to the sample $\z$ 
there is also generated implicitly a sequence of probability measures,
$\Pv=\(\pr{P}{,}{n}\)$
although these 
are not supplied to the learner. Hence the $(n,m)$ sampling process 
can be thought of as a sampling process on $\Znm\times\P^n$. This notion is
used in the following lemma, the proof of which follows directly from the
triangle inequality for $d_\nu$.
\begin{lem}
\label{firstlem}
If
\begin{equation}
\label{ineq1}
\Pr\left\{(\z,\Pv)\in \Znm\times\P^n\colon  
\d{E^*_\G(\A(\z),\z)}{E^*_\G(\A(\z),\Pv)} 
> \frac\alpha2 \right\} \leq \frac\delta2,
\end{equation}
and
\begin{equation}
\label{ineq2}
\Pr\left\{(\z,\Pv)\in \Znm\times\P^n\colon  
\d{E^*_\G(\A(\z),\Pv)}{E^*_\G(\A(\z),Q)} 
> \frac\alpha2 \right\} \leq \frac\delta2,
\end{equation}
then
$$
\Pr\left\{\z\in\Znm\colon  
\d{E^*_\G(\A(\z),\z)}{E^*_\G(\A(\z),Q)} 
> \alpha \right\} \leq \delta.
$$
\end{lem}

Thus in order to determine when (\ref{mainprob}) holds, the conditions under
which (\ref{ineq1}) and (\ref{ineq2}) hold must be found. The two
inequalities are treated separately.

\paragraph{The First Inequality.}
\begin{lem}
\label{firstineq}
\begin{multline}
\label{dunno}
\Pr\left\{(\z,\Pv)\in \Znm\times\P^n\colon  
\d{E^*_\G(\A(\z),\z)}{E^*_\G(\A(\z),\Pv)} > \alpha
\right\} \\
\leq
\Pr\left\{(\z,\Pv)\in \Znm\times\P^n\colon  \exists \comp{\gv}{\fbar}\in \comp{\G^n}{\Fbar}\colon 
\d{E(\comp{\gv}{\fbar},\z)}{E(\comp{\gv}{\fbar},\Pv)} >\alpha\right\}
\end{multline}
\end{lem}
\begin{pf}
Suppose that $(\z,\Pv) \in \Znm\times\P^n$ are such that \\
$\d{E^*_\G(\A(\z),\z)}{E^*_\G(\A(\z),\Pv)} > \alpha$. To simplify things,
denote $\A(\z)$
by $f$.
Suppose first that $E^*_\G(f,\z) \leq E^*_\G(f,\Pv)$.
By definition of the infimum, for all $\ep>0$, $\exists \gv\in\G^n$ such
that
$$
E(\comp{\gv}{\fbar},\z) < E^*_\G(f,\z)+\ep.
$$
Hence by property (3), lemma \ref{dnulem} of the $d_\nu$ metric, 
for all $\ep>0$, $\exists \gv\in\G^n$ such that
$$
\d{E(\comp{\gv}{\fbar},\z)}{E^*_\G(f,\z)} < \ep.
$$
Let $\gv\in\G^n$ satisfy this inequality. By definition,
$$
E^*_\G(f,\Pv) \leq E(\comp{\gv}{\fbar}, \Pv).
$$
Hence
$$
E^*_\G(f,\z) \leq E^*_\G(f,\Pv) \leq E(\comp{\gv}{\fbar},\Pv),
$$
and as $\d{E^*_\G(f,\z)}{E^*_\G(f,\Pv)} > \alpha$, 
by the compatibility of $d_\nu$
with the ordering on the reals (see lemma \ref{dnulem}),
$$
\d{E^*_\G(f,\z)}{E(\comp{\gv}{\fbar},\Pv)} > \alpha = \alpha+\delta,\text{ say.}
$$
By the triangle inequality for $d_\nu$,
\begin{multline*}
\d{E(\comp{\gv}{\fbar},\z)}{E(\comp{\gv}{\fbar},\Pv)} + 
\d{E(\comp{\gv}{\fbar},\z)}{E^*_\G(f,\z)} \\
\geq \d{E^*_\G(f,\z)}{E(\comp{\gv}{\fbar},\Pv)} = \alpha + \delta.
\end{multline*}
Thus,
$$
\d{E(\comp{\gv}{\fbar},\z)}{E(\comp{\gv}{\fbar},\Pv)} > \alpha+\delta-\ep,
$$
and for any $\ep>0$ a $\gv$ satisfying this inequality can be found.
Choosing $\ep=\delta$ shows that there exists $\gv\in\G^n$ such that 
$$
\d{E(\comp{\gv}{\fbar},\z)}{E(\comp{\gv}{\fbar},\Pv)} > \alpha.
$$

If, instead,  $E^*_\G(f,\Pv) < E^*_\G(f,\z)$, then an identical 
argument can be run with the role of $\z$ and $\Pv$ interchanged to show that 
there exists $\gv\in \G^n$ such that 
$$
\d{E(\comp{\gv}{\fbar},\Pv)}{E(\comp{\gv}{\fbar},\z)} > \alpha.
$$
Thus in both cases,
\begin{multline*}
\d{E^*_\G(\A(\z),\z)}{E^*_\G(\A(\z),\Pv)} > \alpha \\
\Rightarrow
\exists \comp{\gv}{\fbar}\in \comp{\G^n}{\Fbar}\colon 
\d{E(\comp{\gv}{\fbar},\z)}{E(\comp{\gv}{\fbar},\Pv)} > \alpha,
\end{multline*}
and so 
\begin{multline*}
\Pr\left\{(\z,\Pv)\in \Znm\times\P^n\colon  
\d{E^*_\G(\A(\z),\z)}{E^*_\G(\A(\z),\Pv)} > \alpha \right\} \\
\leq
\Pr\left\{(\z,\Pv)\in \Znm\times\P^n\colon  \exists \comp{\gv}{\fbar}\in \comp{\G^n}{\Fbar}\colon 
\d{E(\comp{\gv}{\fbar},\z)}{E(\comp{\gv}{\fbar},\Pv)} >\alpha\right\},
\end{multline*}
as required.
\end{pf}

By the nature of the $(n,m)$ sampling process,
\begin{multline}
\label{intprob}
\Pr\left\{(\z,\Pv)\in \Znm\times\P^n\colon  \exists \comp{\gv}{\fbar}\in \comp{\G^n}{\Fbar}\colon 
\d{E(\comp{\gv}{\fbar},\z)}{E(\comp{\gv}{\fbar},\Pv)} >\alpha\right\} \\
= \int\limits_{\Pv\in\P^n}\!\! 
\Pr\left\{\z\in \Znm\colon  \exists \comp{\gv}{\fbar}\in \comp{\G^n}{\Fbar}\colon 
\d{E(\comp{\gv}{\fbar},\z)}{E(\comp{\gv}{\fbar},\Pv)}
>\alpha\right\}\,dQ^n(\Pv),
\end{multline} 
where the probability measure on $\Znm$ used in the integrand on the right hand side is 
the product measure $P_1^m\times\dots\times P_n^m$ generated by 
$\Pv=\(\pr{P}{,}{n}\)\in\P^n$. The probability in the integrand has already
been bounded in \S\ref{gnfsec}, thus replacing $\alpha$ and $\delta$ by
$\alpha/2$ and $\delta/2$ in formula \eqref{gnfmbound} from that section 
gives the
following lemma ensuring the first inequality \eqref{ineq1} is satisfied.
\begin{lem}
\label{ineq1lem}
If
$$
m \geq \frac{32M}{\alpha^2\nu}\left[\ln \C\(\ep_1,l_\G\) + \frac1n
\ln\frac{8\C^*_{l_\G}\(\ep_2,\F\)}{\delta}\right],
$$
where $\ep_1+\ep_2 = \frac{\alpha\nu}{16}$,
then for any representation learner 
$$
\A\colon  \bigcup_{\substack{n\geq 1\\ m\geq 1}} \Znm \to \F,
$$
$$
\Pr\left\{(\z,\Pv)\in \Znm\times\P^n\colon  
\d{E^*_\G(\A(\z),\z)}{E^*_\G(\A(\z),\Pv)} 
> \frac\alpha2 \right\} \leq \frac\delta2.
$$
\end{lem}

\paragraph{The Second Inequality.}
To establish a bound for the probability in the second inequality,
\begin{equation}
\label{secineq}
\Pr\left\{(\z,\Pv)\in \Znm\times\P^n\colon  
\d{E^*_\G(\A(\z),\Pv)}{E^*_\G(\A(\z),Q)} >\ep \right\},
\end{equation}
first note that 
\begin{multline}
\label{exists}
\Pr\left\{(\z,\Pv)\in \Znm\times\P^n\colon  
\d{E^*_\G(\A(\z),\Pv)}{E^*_\G(\A(\z),Q)} >\ep \right\} \\
\leq \Pr\left\{\Pv\in \P^n\colon  \exists f\in\F\colon 
\d{E^*_\G(f,\Pv)}{E^*_\G(f,Q)} >\ep \right\}.
\end{multline}

Recall definitions \eqref{pvloss} and \eqref{exploss} 
for $E^*_\G(f,\Pv)$ and $E^*_\G(f,Q)$,
where $\Pv=\(\pr{P}{,}{n}\)$:
\begin{align*}
E^*_\G(f,\Pv) &= \frac1n\sum_{i=1}^n\inf_{g\in\G}
\E{l_\comp{g}{f}}{P_i}, \\
E^*_\G(f,Q) &= \int_\P\inf_{g\in\G}\E{l_\comp{g}{f}}{P}\,dQ(P).
\end{align*}
For each $\comp{g}{f} \in \comp{\G}{\F}$, define a function
$\glsname{lbargf} \colon \P \to [0,M]$ by
$$
\lbar_\comp{g}{f} \de \E{l_\comp{g}{f}}{P}
$$ and let \glsname{lbarGF} denote the set of all such functions.  For
each $f\in \F$, define a function $\glsname{lf}\colon \P\to [0,M]$ by
$$
l^*_f(P) \de \inf_{g\in\G} \lbar_\comp{g}{f}(P)
$$
and let \glsname{lF} denote the set of all such functions.

It is now possible to introduce the minimal $\sigma$-algebra on $\P$
ensuring all things are measurable. Recalling definition \ref{measure} from
appendix A, $\sigma_{\lbar_\comp{\G}{\F}}$ is the correct $\sigma$-algebra to
use on $\P$, and in fact $\P$ should be the set of probability measures
$\P_{l_\comp{\G}{\F}}$. Furthermore, the environmental measure $Q$ should be
a member of $\P_{\lbar_\comp{\G}{\F}}$ and the $\sigma$-algebra on $Z$ should
be $\sigma_{l_\comp{\G}{\F}}$. This makes everything measurable that needs to
be, including all $l^*_f\in l^*_\F$ (see lemma \ref{bigpermlem} in appendix
A, although the notation there will be more easily understood after reading
section \ref{genframe} of the present chapter). Unless otherwise stated it
will be assumed from now on that $Q$, $\P$, etc.\ all belong to the
appropriate spaces. 

The expectation
of $l^*_f$ with respect to $\Pv = \(\pr{P}{,}{n}\)$ is
\begin{align*}
\E{l^*_f}{\Pv} &= \frac1n \sum_{i=1}^n l^*_f(P_i) \\
	       &= \frac1n \sum_{i=1}^n \inf_{g\in\G} \E{l_\comp{g}{f}}{P_i} \\
	       &= E^*_\G(f,\Pv).
\end{align*}
Similarly,
\begin{align*}
\E{l^*_f}{Q} 	&= \int_\P l^*_f(P)\,dQ(P) \\
		&= \int_\P \inf_{g\in\G} \E{l_\comp{g}{f}}{P} dQ(P) \\
		&= E^*_\G(f,Q).
\end{align*}
Thus,
\begin{multline}
\label{exp2}
\Pr\left\{\Pv\in \P^n\colon  \exists f\in\F\colon 
\d{E^*_\G(f,\Pv)}{E^*_\G(f,Q)} >\ep \right\} \\
= \Pr\left\{\Pv\in \P^n\colon  \exists l^*_f\in l^*_\F\colon 
\d{\E{l^*_f}{\Pv}}{\E{l^*_f}{Q}} >\ep \right\}.
\end{multline}
\begin{defn}
For any probability measure $Q\in \P_{\lbar_\comp{\G}{\F}}$ on
$\P_{l_\comp{\G}{\F}}$, 
define the pseudo-metric \glsname{dQ} on $l^*_\F$ by
$$
d_Q\(l^*_f,l^*_{f'}\) \de \int_\P |l^*_f(P)-l^*_{f'}(P)|\, dQ(P),
$$
and define the $\ep$-capacity of $l^*_\F$ by
$$
\C(\ep,l^*_\F) = \sup_{Q\in\P_{\lbar_\comp{\G}{\F}}} \N(\ep,l^*_\F,d_Q),
$$
with $\C(\ep,l^*_\F) = \infty$ if the supremum is undefined.
\end{defn}

With these definitions the following theorem is obtained immediately from
corollary \ref{n=1}.
\begin{thm}
\label{impthm2}
Let $\l^*_\F$ be a permissible\jfootnote{See section \ref{genframe} of the
present chapter for the conditions ensuring permissibility of $l^*_\F$.}
family of maps from $\P$ into $[0,M]$ and 
let $Q$ be a probability measure on $\P$. 
Let $\Pv$ be generated by $n$ 
independent trials from $\P$ according to $Q$. Then for all $\nu>0$, 
$0<\alpha<1$,
$$
\Pr\left\{\Pv\in\P^n\colon \exists l^*_f\in l^*_\F\colon  \d{\E{l^*_f}{\Pv}}{\E{l^*_f}{Q}} 
>\alpha\right\}
\leq 4 \C(\alpha\nu/8,l^*_\F) e^{-\frac{\alpha^2\nu n}{8M}}.
$$
\end{thm}
With one more definition,
the capacity of $\l^*_\F$ can be bounded in terms of the
capacity of $\F$.
\begin{defn}
\label{QZ}
For any $Q\in\P_{\lbar_{\comp{\G}{\F}}}$, define the probability measure $Q_Z$ on $Z$ by
$$
Q_Z(S) \de \int_\P P(S)\, dQ(P),
$$ 
for all $S$ in the $\sigma$-algebra $\sigma_{l_\comp{\G}{\F}}$ on
$Z$.
\end{defn}
Recalling definition \ref{dbardef} for the metric $d^*_{[P,l_\G]}$ on $\F$
gives the 
following lemma:
\begin{lem}
For all $l^*_f,l^*_{f'}\in l^*_\F$,
$$
d_Q\(l^*_f,l^*_{f'}\) \leq d^*_{[Q_Z,l_\G]}\(f,f'\).
$$
\end{lem}
\begin{pf}
Note that if $h,h'$ are bounded, positive functions on an arbitrary set $X$,
then
\begin{equation}
\label{infeq}
\left|\inf_{x\in X} h(x) - \inf_{x\in X} h'(x)\right| \leq 
\sup_{x\in X} \left|h(x) - h'(x)\right|.
\end{equation}
Now, for all $l^*_f,l^*_{f'}\in l^*_\F$,
\begin{align*}
d_Q\(l^*_f,l^*_{f'}\) &= \int_\P \left|l^*_f(P) - l^*_{f'}(P)\right|\, dQ(P) \\
		  &= \int_\P \left|\inf_{g\in\G}\int_Z
		     l_\comp{g}{f}(z)\,dP(z)  -
		     \inf_{g\in\G} \int_Z
		     l_\comp{g}{f'}(z)\,dP(z)\right|\,dQ(P)\\
		  &\leq \int_\P\sup_{g\in\G} \left| \int_Z l_\comp{g}{f}(z) -
			l_\comp{g}{f'}(z)\,dP(z)\right|dQ(P) 
			\qquad\text{(by \eqref{infeq} above)}\\
		  &\leq \int_\P\sup_{g\in\G} \int_Z \left|l_\comp{g}{f}(z) -
			l_\comp{g}{f'}(z)\right|\,dP(z)\,dQ(P) \\
		  &\leq \int_\P\int_Z\sup_{g\in\G}\left|l_\comp{g}{f}(z) -
			l_\comp{g}{f'}(z)\right|\,dP(z)\,dQ(P)\\
		  &= \int_Z\sup_{g\in\G}\left|l_\comp{g}{f}(z) -
			l_\comp{g}{f'}(z)\right|\,dQ_Z(z)\\
		  &= d^*_{[Q_Z,l_\G]}(f,f').
\end{align*}
\end{pf}
Thus the surjective map $\psi\colon \(\F,d^*_{[Q_Z,l_\G]}\)\to \(l^*_\F,d_Q\)$ 
where $\psi(f) = l^*_f$ is 
a contraction and so by lemma \ref{isom},
$$
\N\(\ep,l^*_\F,d_Q\) \leq \N\(\ep,\F, d^*_{[Q_Z,l_\G]}\).
$$
By lemma \ref{bigpermlem}, $\P_{l_\comp{\G}{\F}}$ is guaranteed to
contain the measure $Q_Z$ for every
$Q\in\P_{\lbar_\comp{\G}{\F}}$, and so 
\begin{lem}
\label{lastclem}
$$
\C\(\ep,l^*_\F\) \leq \C^*_{l_\G}\(\ep,\F\).
$$
\end{lem}
The probability in the second inequality can now be bounded.
\begin{lem}
\label{ineq2lem}
For all $\nu>0$, $0<\alpha<1$ and $0<\delta<1$, assuming $l^*_\F$ is
permissible and 
$$
n \geq \frac{32 M}{\alpha^2\nu}\ln\(
\frac{8 \C^*_{l_\G}\(\frac{\alpha\nu}{16},\F\)}{\delta}\),
$$
then for any representation learner,
$$
\A\colon  \bigcup_{\substack{n\geq 1\\ m\geq 1}} \Znm \to \F,
$$
$$
\Pr\left\{(\z,\Pv)\in \Znm\times\P^n\colon  
\d{E^*_\G(\A(\z),\Pv)}{E^*_\G(\A(\z),Q)} >\frac{\alpha}{2} \right\} 
\leq \frac{\delta}{2}.
$$
\end{lem}

\paragraph{Main Bound.}
Putting lemmas \ref{ineq2lem}, \ref{ineq1lem} and \ref{firstlem} together gives
the following theorem.
\begin{thm}
\label{mainthm}
Let $\F, \G$ be families of functions with the structure  $X \xrightarrow{\F}V \xrightarrow{\G}
A$, and let $l$ be a loss function $l\colon Y\times A\to [0,M]$. Assume $\comp{\G}{\F}$ is such 
that $l_\comp{\G^n}{\Fbar}$ and $l^*_\F$ are permissible\jfootnote{Both these
conditions can be subsumed within one criterion known as {\em
f-permissibility}, see section \ref{genframe} and appendix \ref{permissapp}.}.
Let $\z\in \Znm$ be an $(n,m)$ sample from $Z$ according 
to an environmental measure $Q$. 
For all $0<\alpha,\delta,\ep_1,\ep_2<1$, $\nu>0$, 
$\ep_1+\ep_2=\frac{\alpha\nu}{16}$, 
if 
\begin{align*}
n &\geq \frac{32 M}{\alpha^2\nu} 
\ln\frac{8 \C^*_{l_\G}\(\frac{\alpha\nu}{16}, \F\)}{\delta}, \\
\text{and} \quad m &\geq \frac{32M}{\alpha^2\nu}\left[\ln\C\(\ep_1,l_\G\) + \frac1n
\ln\frac{8\C^*_{l_\G}\(\ep_2,\F\)}{\delta}\right],
\end{align*}
and $\A$ is any representation learner,
$$
\A\colon  \bigcup_{\substack{ n\geq 1\\ m\geq 1}} \Znm \to \F,
$$
then
$$
\Pr\left\{\z\in \Znm\colon \d{E^*_\G(\A(\z),\z)}{E^*_\G(\A(\z),Q)} > \alpha\right\}
\leq \delta.
$$
\end{thm}

\subsection{Discussion}
Theorem \ref{mainthm} shows, unsurprisingly,
that both $n$ and $m$ need to be sufficiently
large to ensure with high probability good generalisation from a
representation. The interaction between the two bounds is best understood by
assuming that $\F$ and $\G$ are neural networks and applying the bounds for
the capacities derived in appendix \ref{nnetapp}. Setting the total number of
weights in $\F$ to be $W_\F$, and similarly $W_\G$ for $\G$, and looking
only at the dependence on $\alpha, \nu, \delta, W_\F$ and $W_\G$ of the above
bounds, gives\jfootnote{The $\tilde{O}$ notation indicates that logarithmic
factors have been suppressed in the same way that constant factors are
suppressed with the $O$ notation. It is borrowed from Haussler et. al. 
(1994)\label{Het1}.} 
\begin{align}
\label{n}
n &= \tilde{O}\(\frac{W_\F}{\alpha^2\nu}\)\\
\label{mn}
m &= \tilde{O}\(\frac1{\alpha^2\nu} \left[W_\G + \frac{W_\F}n\right]\).
\end{align}
The bound on $m$ for ordinary learning would read in this case, 
\begin{equation}
\label{mo}
m = \tilde{O}\(\frac1{\alpha^2\nu} \left[W_\G + W_\F\right]\).
\end{equation}
If the representation is not to be used for learning future
tasks from the same environment then the bound \eqref{n} on $n$ can be
ignored, and then a comparison between \eqref{mn} and \eqref{mo} clearly
demonstrates the superiority of representation learning over ordinary
learning, at least on a number of examples required per task basis and in
the case that $W_\F \gg W_\G$.
However, if the representation is to be used for learning future tasks from
the same environment, then the bound on $n$ must be taken seriously, leading
to the following expression for the 
{\em total} number of examples
$mn$ required for good generalisation,
$$
mn = \tilde{O}\(\frac{W_\F W_\G}{\alpha^4\nu^2}\).
$$
On face value, the fact that this quantity is far greater than that required
for ordinary learning might prompt the conclusion that representation
learning is infeasible. This is not the case, as (1): 
representation learning is
not intended to be performed {\em on-line} as is often true of 
ordinary learning
and (2): for representation learning it is {\em assumed} that there are
many tasks from which to collect examples so the increased burden may be
spread out. Furthermore, once a successful representation $f$ has been
learnt, the hypothesis space used by the learner will be $\comp{\G}{f}$, 
not $\comp{\G}{\F}$. Applying lemma \ref{litcomplem} and using the fact that
the capacity of a space consisting of a single function is one, gives
$\C(\ep,l_\comp{\G}{f}) \leq \C(\ep,l_\G)$. This gives an order of magnitude
estimate for the number of examples required for learning using a
representation of
\begin{equation}
\label{mg}
m = \tilde{O}\(\frac{W_\G}{\alpha^2\nu}\).
\end{equation}
Comparing \eqref{mg} with \eqref{mo} shows that
if $W_\F \gg W_\G$, all the extra work involved in learning the
representation is rewarded when it comes time to learn new tasks
using the representation.

\section{General Framework}
\label{genframe}
Learning a representation is just one example of hypothesis space bias.
Other examples from artificial neural network research include biasing by
limiting the number of hidden nodes and layers posessed by the networks and
limiting the number of non-zero weights or the size of the weights.
In this section, the model of hypothesis space bias used in the previous
section is generalised to cover such learning scenarios, and in fact
any learning scenario in which the learner
receives an $(n,m)$ sample from the environment and uses it to bias the
hypothesis space. Bounds on the $(n,m)$-sample size required for PC learning
will be derived and expressed, as before, in terms of the capacity of certain
classes of functions. 

\subsection{Preliminaries.}

In general, a learner that receives $(n,m)$ samples $\z\in\Znm$ and
biases its hypothesis space in some way, must effectively have a {\em
  family} of hypothesis spaces from which to choose. Let $\glsname{HH}
= \{\H\}$ denote this family.  Given an $(n,m)$-sample
$\z=\(\zv_1,\dots,\zv_n\)$, drawn according to some probability
measure $Q$ on the space of probability measures $\P$ on $Z$, and a
loss function $l\colon Y\times A\to [0,M]$, the learner searches for a
hypothesis space $\H\in H$ with minimal {\em empirical loss} on $\z$,
where this is denoted by \glsname{EHz} and defined by
\begin{equation}
\label{genemploss}
E^*\(\H,\z\) \de \frac1n \sum_{i=1}^n \inf_{h\in\H} \E{l_h}{\zv_i}.
\end{equation}
$E^*\(\H,\z\)$ is an estimate of the {\em true loss} of $\H$
\begin{equation}
\label{gentrueloss}
\glsname{EHQ} \de \int_\P \inf_{h\in\H} \E{l_h}{P}\,dQ(P).
\end{equation}
The true loss of a hypothesis space
$\H$ is a measure of how effective $\H$ is for learning within the
environment $Q$.
To see that representation learning is a special case of this framework, let
$\F$ be the representation space and $\G$ the output function space of the
learner, and let  
$H=\{\comp{\G}{f}\colon f\in\F\}$. The empirical and true loss of a
hypothesis space $\comp{\G}{f}\in H$ is the same as the empirical 
and true loss of the corresponding representation $f$, as defined
in equations \eqref{emploss} and \eqref{exploss} in section \ref{envlearn}.

In determining the empirical loss of a hypothesis space $\H$, the learner
searches for a sequence of $n$ functions $\hv = (h_1,\dots,h_n), h_i\in\H$ 
with minimal empirical loss on the sample $\z$, where this is defined as in
\eqref{empEloss} and reproduced below,
$$
E(\hv,\z) \de \frac1n\sum_{i=1}^n \E{l_{h_i}}{\zv_i}.
$$
If the learner is biasing the hypothesis space with the intention of using
it to learn future tasks from the same environment, then, as with
representation learning, it needs to know how large to make $n$ and $m$ to
ensure with high probability that \eqref{genemploss} and \eqref{gentrueloss}
are close. However, as discussed in section \ref{gnfsec}, the $(n,m)$ sample
$\z$ may contain examples of all the tasks the learner is ever going to be
asked to learn, in which case the learner will only ever be using the
sequence $\hv=\(h_1,\dots,h_n\)$ and will primarily be interested in
bounding the deviation between \eqref{empEloss} and the true loss of $\hv$,
where this is defined as in \eqref{trueEloss} and reproduced below,
$$
E(\hv, \Pv) \de \frac1n\sum_{i=1}^n \E{l_{h_i}}{P_i},
$$
where $\Pv=\(P_1,\dots,P_n\)$ are the probability measures used to generate
$\z$. Both kinds of deviation can be bounded with simple extensions to the
theory developed in previous sections.

\subsection{Generalisation on $n$ tasks.}

\sloppy
To begin with, for any family of hypothesis spaces $H$, define
$$
H_\sigma \de \{h\in\H\colon \H\in H\},
$$
and define
$$
H^n \de \{\H^n\colon \H\in H\},
$$
where $\H^n = \{(h_1,\dots,h_n)\colon  h_i\in\H, 1\leq i \leq n$ and
$(h_1,\dots,h_n)(x_1,\dots,x_n) = (h_1(x_1),\dots,h_n(x_n))$ for all
$(x_1,\dots,x_n)\in X^n$.
It will be assumed that $H$ has been constructed so that 
$H_\sigma$ represents the completely unbiased hypothesis space the learner
would have to use if it had no extra information about the kinds of 
learning problems to be encountered.

\fussy
The operations $\sigma$ and raising to the power of $n$ clearly do not
commute. In fact,
\begin{equation}
\label{noncom}
\[H^n\]_\sigma \subseteq \[H_\sigma\]^n
\end{equation}
always. For example, in representation learning, with
$H=\{\comp{\G}{f}\colon f\in\F\}$ as above,
$\[H^n\]_\sigma = \comp{\G^n}{\Fbar}$ while $\[H_\sigma\]^n =
\(\comp{\G}{\F}\)^n=\comp{\G^n}{\F^n}$, and the two are equal only in the 
trivial case that $\F$ contains a single function. 
The extent of this noncommutativity controls the advantage that can be
achieved through learning $(n,m)$ samples as opposed to ordinary $(m)$
samples, at least in the worst case analysis presented here. 
Recalling equations \eqref{gnbound} and \eqref{gnfbound} in 
section \ref{advantage},  it can be seen that, depending on the relative
capacities of $\G$ and $\F$, anything from no reduction at all to an $n$-fold
reduction in the number of examples per function required for PC learning
can be achieved through learning an $(n,m)$ sample as opposed to an ordinary 
sample. No reduction essentially corresponds to equality in expression
\eqref{noncom} whereas maximum reduction occurs when the two sets are
maximally different. In fact the range
exhibited by representation learning is prototypical of all $(n,m)$-sample
learning, in that an $n$-fold reduction is the maximum possible for {\em any}
learner, as will be demonstrated shortly.

A learner that receives an $(n,m)$ sample $\z$ and produces $n$ hypotheses
$\(h_1,\dots, h_n\)$ in response, where all the $h_i$ belong to the same
hypothesis space $\H\in H$, is effectively a map $\A$ from the space of all
$(n,m)$ samples into $\[H^n\]_\sigma$,
$$
\A\colon  \bigcup_{\substack{ n\geq 1\\ m\geq 1}} \Znm \to \[H^n\]_\sigma,
$$
although, as usual the results derived here apply equally well to a
stochastic learner.

Given a loss function $l\colon Y\times A\to [0,M]$ and a hypothesis space family
$H$ consisting of hypothesis spaces whose elements are all functions mapping
$X\to A$, define $l_H = \{l_\H\colon  \H\in H\}$ (recall that $l_\H =
\{l_h\colon h\in\H\}$).
The definitions and arguments of section \ref{gnfsec} can be adpated 
to this more general learning scenario by simply replacing all occurences of
$\comp{\G^n}{\Fbar}$ in that section with $\[H^n\]_\sigma$.
This yields the following 
theorem on the sample size $m$ required for PC learning using the hypothesis
space family $H$\jfootnote{Note that $l_{[H^n]_\sigma} =
\bigcup\limits_{\H\in H}\{l_\H\oplus\dots\oplus
l_\H\}\subseteq l_{H_\sigma}\oplus\dots\oplus l_{H_\sigma}$ and so theorem
\ref{fundthm} applies to $l_{\[H^n\]_\sigma}$.}.
\begin{thm}
\label{gengnfbound}
Let $H$ be a family of hypothesis spaces such that each hypothesis space
$\H\in H$ is a set of functions mapping $X$ into $A$. Let $l$ be a loss
function $l\colon  Y\times A\to [0,M]$ and suppose that $l_H$ is
f-permissible\jfootnote{Among other things this ensures $l_{[H^n]_\sigma}$ is 
permissible---see appendix \ref{permissapp}.}. 
Let $\Pv=\(P_1,\dots,P_n\)\in\P_{l_{H_\sigma}}$ be $n$ probability
measures on $Z$. Let $\z\in \Znm$ be an $(n,m)$ sample generated by
sampling $m$ times from $Z$ according to each $P_i$. For all
$0<\alpha<1$, $0<\delta<1$, $\nu>0$ and any $(n,m)$ sample learner 
$$
\A\colon  \bigcup_{\substack{ n\geq 1\\ m\geq 1}} \Znm \to \[H^n\]_\sigma,
$$
if 
$$
m > \frac{8M}{\alpha^2\nu n}\ln\frac{4 \C\(\ep,l_{\[H^n\]_\sigma}\)}{\delta},
$$
then
$$
\Pr\{\z\in\Znm\colon  \d{E(l_{\A(\z)},\z)}{E(l_{\A(\z)},\Pv)} > \alpha\}
< \delta.
$$
\end{thm}
The $n=1$ case of this theorem gives the sample size required for ordinary
learning using the unbiased space $H_\sigma$,
$$
m > \frac{8M}{\alpha^2\nu}\ln\frac{4 \C\(\ep,l_{H_\sigma}\)}{\delta}.
$$
Using the results on capacity from appendix \ref{capapp} 
it is a fairly trivial exercise to show
$$
\C\(\ep,l_{H_\sigma}\) \leq \C\(\ep,l_{\[H^n\]_\sigma}\) \leq
\C\(\ep,l_{H_\sigma}\)^n,
$$
which, in conjuction with theorem \ref{gengnfbound}, 
shows that at worst an $(n,m)$ sample
learner will require as many examples per function for PC learning as an
ordinary learner, and at best a factor of $n$ less than the ordinary learner.
The ratio
$$
I\(H,n,\ep\) \de \frac1n\frac{\log\C\(\ep,l_{\[H^n\]_\sigma}\)}
{\log\C\(\ep,l_{H_\sigma}\)}
$$
measures the usefulness of $H$ for learning $(n,m)$ samples, and will be
called the {\em learning impedance} of $H$. A learning impedance of $1$
indicates that the learner can do no better than ordinary learning with 
$H$, i.e. the learner would be just as well off learning a single task at a
time with $H_\sigma$, 
while an impedance of $\frac1n$ (the best possible) indicates that 
PC learning of an $(n,m)$ sample using $H$ can be achieved with $m$ a factor
of $n$ smaller than that required for ordinary learning.

For representation learning, $H = \{\comp{\G}{f}\colon  f\in\F\}$, and
the impedance of $H$ obeys (roughly speaking)
$$
I(H,n,\ep) = \frac{\frac1n + \frac{\log\C(\ep_1,l_\G)}{\log\C^*_{l_\G}(\ep_2,\F)}}
                  {1       + \frac{\log\C(\ep_1,l_\G)}{\log\C^*_{l_\G}(\ep_2,\F)}}
$$
where $\ep_1+\ep_2 = \ep$. Thus the smaller the ratio of the logarithm of the 
capacities of $\G$ and $\F$, the smaller the learning impedance of $H$. The
limiting cases of $I(H,n,\ep)=1$ and $I(H,n,\ep)=\frac1n$ correspond to
total domination of $H$ by $\G$ and $\F$ respectively.

\subsection{Learning the hypothesis space.}
In this section bounds on $n$ and $m$ ensuring small deviation between
\eqref{genemploss} and \eqref{gentrueloss} are given, similar to the bounds
in theorem \ref{mainthm}. The learner is now taking an $(n,m)$ sample $\z$
as input and producing a hypothesis space $\H\in H$ as output, so it is a map,
$$
\A\colon  \bigcup_{\substack{ n\geq 1\\ m\geq 1}} \Znm \to H.
$$
\begin{defn}
For all $h\in H_\sigma$, define $\lbar_h\colon \P_{l_{H_\sigma}}\to [0,M]$ by 
$$
\lbar_h(P) = \E{\lbar_h}{P}.
$$
For all $\H\in H$, define $l^*_\H\colon \P_{l_{H_\sigma}}\to [0,M]$ by 
$$
l^*_\H(P) \de \inf_{h\in\H} \lbar_h(P).
$$
Let $l^*_H$ denote the set of all such functions.
\end{defn}
Note that for representation learning, $l^*_H = l^*_\F$.

The same arguments leading to theorem \ref{mainthm} can be used in this more
general context to yield the following theorem.
\begin{thm}
\label{genmainthm}
Let $H$ be a hypothesis space family such that each $\H\in H$ is a set of
functions mapping $X$ into $A$. 
Let $l$ be a loss function $l\colon Y\times A\to [0,M]$, and assume that $l$ is
such that $l_H$ is f-permissible\jfootnote{This guarantees permissibility of
both $l_{[H^n]_\sigma}$ and $l^*_{H}$. See lemma \ref{bigpermlem}.}
Let $\z\in \Znm$ be an $(n,m)$ sample from $Z$ generated according 
to some $Q\in\P_{\lbar_{H_\sigma}}$. For all $0<\alpha,\delta<1$, $\nu>0$,  
if 
\begin{align*}
n &\geq \frac{32 M}{\alpha^2\nu} 
\ln\frac{8 \C\(\frac{\alpha\nu}{16}, l^*_H\)}{\delta}, \\
\text{and} \quad m &\geq \frac{32M}{\alpha^2\nu n}
\ln\frac{8 \C\(\frac{\alpha\nu}{16},l_{[H^n]_\sigma}\)}{\delta},
\end{align*}
and $\A$ is any hypothesis space learner,
$$
\A\colon  \bigcup_{\substack{ n\geq 1\\ m\geq 1}} \Znm \to H,
$$
then
$$
\Pr\left\{\z\in \Znm\colon \d{E^*(\A(\z),\z)}{E^*(A(\z),Q)} > \alpha\right\}
\leq \delta.
$$
\end{thm}

\chapter{Algorithms and Experiments}
\label{expchap}
In this chapter algorithms are given for learning representations
with Artificial Neural Networks, and the results of several experiments are
presented which verify the theoretical conclusions of the previous chapter. 

All the examples considered here concern perhaps the most familiar of learning
scenarios---that of learning Boolean functions over some input domain $X$.
Thus the outcome space $Y$ is the set $\{0,1\}$, or equivalently $\bool$, 
the probability 
measures $P\in\P$ consist of a measure $P$ on $X$ and a function $f\colon X\to Y$,
so that samples $z=(x,y)\in Z=\XY$ are always of the form $(x,f(x))$. 
For the particular learning scenarios treated in this section $P$ is the
same regardless of the function $f$.  
Thus, denoting the space 
of all Boolean functions on the domain $X$ by $\F_2(X)$,  
the environmental measure $Q$ is effectively a measure on $\F_2(X)$. 
An $(n,m)$
sample $\z\in\Znm$ is produced by first sampling $n$ times from $\F_2(X)$
according to $Q$ to
generate a list of functions $\fv = (f_1,\dots,f_n)$,
and then sampling
$m$ times from $X$, $n$ times over according to $P$ to generate
$x_{11},\dots,x_{1m},\dots,x_{n1},\dots,x_{nm}$. The final $(n,m)$ sample is
$$
\z =
\begin{matrix}
(x_{11},f_1(x_{11})), & (x_{12},f_1(x_{12})), &\hdots & (x_{1m},f_1(x_{1m}))\\ 
(x_{21},f_2(x_{21})), & (x_{22},f_2(x_{22})), &\hdots & (x_{2m},f_2(x_{2m}))\\ 
\vdots & \vdots & \ddots & \vdots \\
(x_{n1},f_n(x_{n1})), & (x_{n2},f_n(x_{n2})), &\hdots & (x_{nm},f_n(x_{nm}))
\end{matrix}
$$
The learner's hypotheses are functions $f\colon X\to [0,1]$
so $\z$ is intended to give the learner some idea of the kinds of
functions it will be asked to learn.

Modulo the fact the analysis in chapter \ref{repchap} is only for the
worst-case situation, we have seen that the
three main advantages of representation learning over ordinary machine 
learning are:
\begin{itemize}
\item The number of examples $m$ required
per function for PC learning decreases as the number of functions $n$ being
learnt increases.
\item If $n$ is sufficiently large---determined 
principally by the 
capacity of the representation space $\F$---then the representation learnt
will, with high probability, be a good one for learning future functions drawn
from the same environment $Q$.
\item Learning with a representation will in general be far more efficient
than learning without.
\end{itemize}

In order to take advantage of these theoretical results in practical learning
problems an algorithm must be found for learning representations from 
$(n,m)$ samples. 
If the representation space $\F$ and the space 
$\G$ consist of feedforward networks (as defined in appendix \ref{nnetapp}), 
then 
it is a simple matter to extend existing gradient-descent
``backpropagation'' type 
algorithms for training those networks to training  
algorithms for representation learning. Such an algorithm is used to learn
representations in 
two experimental situations. In the first the environment
$Q$ is restricted to have 
support only on a kind of ``translation invariant'' Boolean function, 
and in the second
$Q$'s support is restricted to a subset of the symmetric Boolean functions.
In both cases the three theoretical results summarised above are well
supported.
These experiments are discussed in detail in the last two sections of this
chapter. However the first section gives the details of a somewhat simpler
experiment using binary neural networks and exhaustive search rather than
gradient descent to learn representations.

All experiments were performed on the CM5 machine at the
South Australian Centre for Parallel Supercomputing, and all programs were
written in $C^*$, Thinking Machine's parallel version of $C$.

\section{Learning Binary Neural Network Representations}

In this first experiment the second central tenet 
of representation learning theory is verified, namely 
if $n$ and $m$ are sufficiently 
large, a representation $f$ with small empirical loss on $\z\in\Znm$ will
be ``useful'' for learning future functions drawn from the same environment.
The general idea 
of the experiment is to first choose an arbitrary representation space 
$\F$ and output function space $\G$ for the representation learner to use, and
then the support of the environmental measure $Q$ is chosen to be $\comp{\G}{f^*}$
where $f^*$ is some representation in $\F$. By doing this it is guaranteed that 
for any $(n,m)$ sample chosen according to $Q$, there will exist at least one 
representation (namely $f^*$) with zero empirical loss on the sample, and also 
if $\F$ and $\G$ are sufficiently simple it will be  
computationally feasible to find a representation with zero-empirical loss on
any $(n,m)$ sample simply by exhaustively searching through all representations 
$f\in\F$. The main experiment consisted of generating $(n,m)$ samples for many 
different values of $n$ and $m$, finding all representations with zero empirical
loss on the sample and then measuring the performance of a learner that used the 
resulting representation to learn. The details of the experiment follow. 

The input space $X$ of the learner was chosen to be $X=\bool^5$ and
all examples $x\in X$ were generated according to the uniform distribution
on $X$, which will be denoted by $P$. 
With this restriction on the distribution on $X$, the only
thing that the environmental measure $Q$ has control over are the functions 
seen by the learner.
The support of $Q$, a subset $F$ of $\F_2(X)$, was constructed
by firstly selecting at random
a {\em binary neural network} 
$f^*$ with five input nodes, no hidden layers and three output nodes, and
then composing any Boolean function with the output of $f^*$.
A binary neural 
network is one with $\bool$ valued weights, no thresholds and 
the hard-limiting sign function 
$$
\theta(x) = \left\{ \begin{array}{ll}
                       1 & \mbox{if $x>0$} \\
                      -1 & \mbox{if $x \leq 0$}
                      \end{array} \right.
$$
for a squashing function. Thus $f^*$ is a function mapping\jfootnote{There is
nothing special about the numbers 
being used in this example, they were chosen merely to ensure the tractability 
of the experiment.} $X=\bool^5$ into 
$\bool^3$. For later reference, denote the space of all such binary
neural 
networks by $N_{5,3}$. $Q$'s support $F$ thus consisted of all 
Boolean functions from $X$ into $\bool$ that can be formed by the composition 
of any function from $\F_2(\bool^3)$ with $f^*$. $F$ is a 
highly restricted subset of all possible Boolean functions on $X$ because 
$F$ contains at most 256 functions whereas  $|\F_2(X)| = 2^{32}$. 
Functions were
selected from $F$ by choosing a function from $\F_2(\bool^3)$ uniformly at random
and then composing it with $f^*$.

To ensure 
that the learner could achieve zero empirical loss on any $(n,m)$ sample 
the choices $\G=\F_2(\bool^3)$ for the learner's output function space
and $\F = N_{5,3}$ for its representation space were made.
Hence the output space $Y$
and the action space $A$ were the same $\bool$.
The loss function $l$ was chosen to be simply $l(a,y) = 0$ if $a = y$ and
$l(a,y) = 1$ otherwise, i.e 
$$
l(a,y) = 1-\delta(a,y)
$$
where $\delta$ is the Kronecker delta function.
With this choice of $l$ the empirical loss of any function
$\comp{g}{f} \in \comp{\G}{\F}$ on a sample
$\zv = \((x_1,y_1),\dots,(x_m,y_m)\)$ is given by
$$
E(\comp{g}{f}, \zv) = 1 - \frac1m\sum_{i=1}^m \delta(\comp{g}{f}(x_i), y_i).
$$
The empirical loss of a representation $f\in\F$ on an $(n,m)$ sample
$\z = \(\zv_1,\dots,\zv_n\)$ is given by (recall equation \eqref{emploss} in
chapter \ref{repchap}),
$$
E^*_\G(f,\z) = \frac1n\sum_{i=1}^n \inf_{g\in\G} E(\comp{g}{f},\zv_i).
$$
The fact that $\G$ consisted of every Boolean function on the range of $\F$
simplified the task of computing the empirical loss of a representation $f$.
To see this, note that given any sample 
$\zv_i = \((x_{i1},y_{i1}),\dots,(x_{im},y_{im})\)$, each input vector 
$x_{ij}$ will be mapped 
by $f$ to one of the eight elements of $\bool^3$. Denoting the elements of 
$\bool^3$ by $s_1,\dots,s_8$, $\zv_i$ can be partitioned by $f$ 
into a disjoint union of sets $S^i_1,\dots,S^i_8$ where 
$S^i_a = \{(x_j,y_j)\in\zv_i\colon f(x_j) = s_a\}$, $1\leq a\leq 8$.
Note that some of the $S^i_a$ may be empty. For each set $S^i_a$ let
$\plus(S^i_a)$ be the number of points $(x_j,y_j)$ in $S^i_a$ such that 
$y_j=1$ 
and similarly let $\minus(S^i_a)$ be the 
number of points for which $y_j=-1$. Clearly the function $g$ that minimizes the
empirical loss of $\comp{g}{f}$ on $\zv_i$ is the one $g(s_i) = 1$ if 
$\plus(S^i_a) > \minus(S^i_a)$ and 
$g(s_i) = -1$ if $\minus(S^i_a) > \plus(S^i_a)$ (the choice of 
$g(s_i)$ does not affect the value of the empirical loss if 
$\plus(S^i_a) = \minus(S^i_a)$). With this choice of $g$ the empirical loss of 
$\comp{g}{f}$ on $\zv_i$ is $\frac1m\sum_{a=1}^8
\min(\plus(S^i_a),\minus(S^i_a))$ and
hence the empirical loss of the representation $f$ on the sample
$\z=\(\zv_1,\dots,\zv_n\)$ is given by
\begin{equation}
\label{form3}
E^*_\G(f,\z) = \frac1{mn}\sum_{i=1}^n\sum_{a=1}^8
\min\(\plus(S^i_a),\minus(S^i_a)\).
\end{equation}
The important thing about this formula is that it can be computed directly from
the behaviour of $f$ on the sample $\z$, so that it is not necessary 
to explicitly search the space $\G$ to determine $f$'s empirical loss.

\subsection{The Experiment}

$(n,m)$ samples $\z$ were generated according to the environment $Q$ described 
above for values of $m$ in the set $\{2,6,10,14,18,22\}$  and values of $n$ in the
set $\{1,2,3,4,5,6,7,8,9\}$. Ten independent $(n,m)$ samples were generated
for each $(n,m)$ pair, and the results were averaged across all ten samples. 
For each $(n,m)$ sample $\z$, 
$\F$  was exhaustively searched to find {\em all} 
representations with {\em zero} empirical loss on $\z$,
utilising formula (\ref{form3}). The environment Q was then sampled ten more 
times to generate ten new Boolean functions
$\comp{g_1}{f^*},\dots,\comp{g_{10}}{f^*}$, and training sets with sizes of 
$2$ up to $22$ again  
were generated for each of the ten functions. Thus there 
were 60 new training sets 
$\zv^1_2, \zv^1_6, \zv^1_{10}, \dots, \zv^{10}_{14}, \zv^{10}_{18}, 
\zv^{10}_{22}$. Each zero-loss 
representation $f$ was then tested for its learning ability against each of the ten 
new test functions
as follows. For each training set $\zv^i_m$, $1\leq i\leq 10,
m\in\{2,6,10,14,18,22\}$, the function $g\in\G$ giving the smallest 
value of $E(\comp{g}{f},\zv^i_m)$ was found and then the total error between
$\comp{g}{f}$ and the original function $\comp{g_i}{f^*}$ was determined
(simply by going through all $32$ ($2^5$) vectors in the input space $X$ and
counting up the number of times $\comp{g}{f}$ and $\comp{g_i}{f^*}$
disagreed). For each training set size and each zero-loss representation 
the total error was averaged 
across all ten functions, and then further averaged across all of the 
zero-loss representations. 
This rather cryptic process yields a number $x$ for each triple $n$, $m$ and
$m_1$ which is an estimate of the expected generalisation error that a learner
would see if it received an $(n,m)$ sample $\z$ 
drawn according to $Q$, selected a representation $\fhat$ 
uniformly at random
with zero loss on $\z$, then received a further $m_1$ examples of a new
function drawn according to $Q$ and found a function $\comp{g}{\fhat}$ with
smallest empirical loss on the new sample. $x$ is an estimate of the expected 
true error (i.e generalisation error) of $\comp{g}{\fhat}$. Plotting $x$ as
a function of $m_1$ for each value of $m$ and $n$ in the
range specified above yields a set of expected 
{\em generalisation curves}. 
For comparison purposes two more generalisation curves were also generated.
The first is the curve that would be produced in an 
ordinary learning scenario, i.e 
using the same training sets $z^1_2, z^1_6, \dots$ above, the full
space $\comp{\G}{\F}$ was searched to find all networks with zero loss on 
each training set and then the total error for each network was computed and 
averaged across all ten
functions and across all networks with zero loss. The second curve was the one that 
would be generated by a representation learner that had first learnt the exact 
representation $f^*$. Note that the exact representation curve is the best curve
(on average) that a representation learner could achieve.

The generalisation curves for representative values of $m$ and $n$ are plotted 
in figures \ref{binfig1} and \ref{binfig2}, along with a plot of the
ordinary  learning curve and 
the exact representation curve. Notice that the 
generalisation behaviour of the exact representation curve is superior to
the ordinary learning curve---the exact representation learner 
has been given a big head start 
by having its input layer weights already set to their correct values whereas 
the ordinary learner has to determine both the input layer weights and the 
output function $g$. 
For small values of $n$ and $m$ the generalisation curves for the 
representation learner are considerably worse than the ordinary machine learning 
curve (the cause of the dip in the early part of the ordinary learning curve
is a mystery, but may simply be an artefact of insufficient sampling). 
This is to be expected because there is insufficient information in
a small $(n,m)$ sample to determine a good representation and so, after 
the representation learning phase, the representation 
learner in most cases will be locked into using a bad representation when 
learning new functions. However, as $n$ and $m$ increase,
the generalisation curves of the representation learner improve, and
eventually the representation learner 
begins to perform better than the ordinary learner. As $n$ increases even
further, for large enough values of $m$, the learner's
performance approaches that of
the best possible (the exact) representation curve. Thus,
for large enough values 
of $n$ and $m$, a representation $f$ with zero empirical loss on the 
data is pretty close to the true representation $f^*$, and hence exhibits
generalisation behaviour close to that achieved by $f^*$.

\begin{figure}
\epsfbox{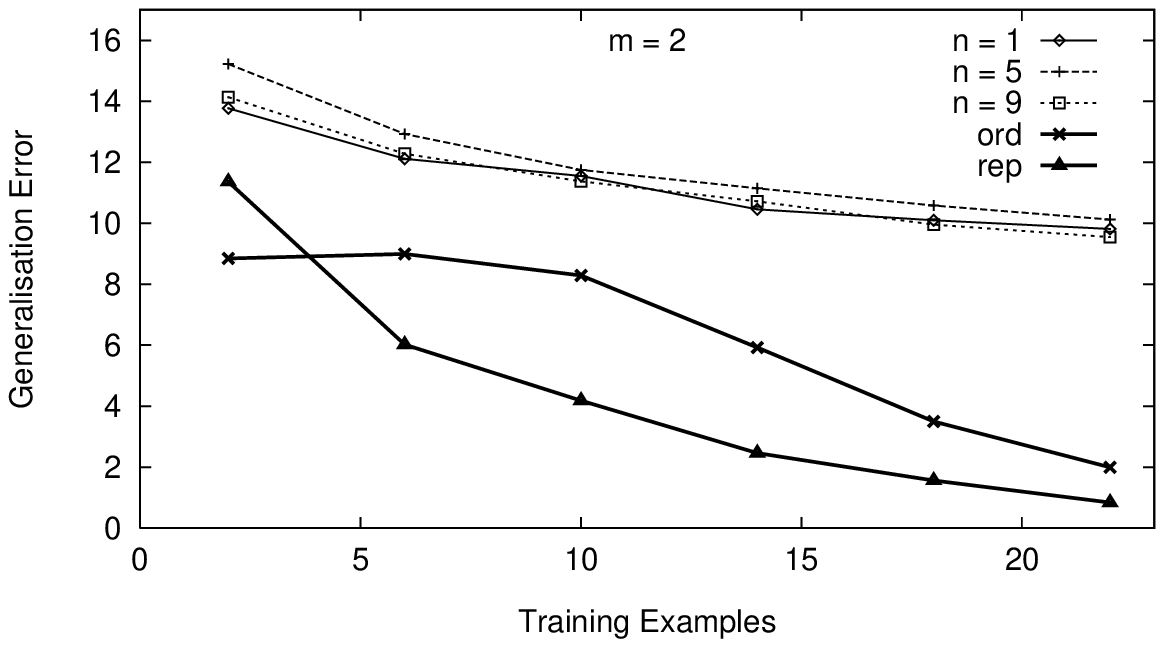}
\vspace{1mm}
\epsfbox{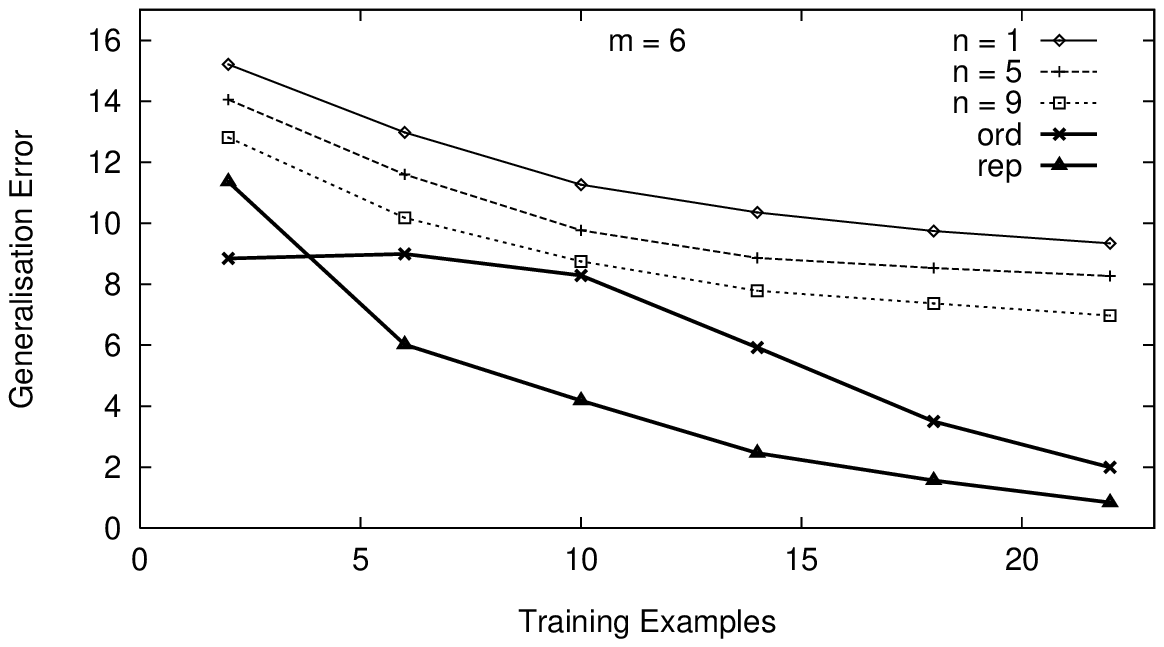}
\jcaption{Binary representation generalisation curves ($m=2$ and $m=6$). 
$n$ is the number of functions sampled from the environment and $m$ is the
number of times each function was sampled.
The curves labelled ``rep'' and ``ord'' are the best representation learning
curve and ordinary learning curve respectively.}
{binfig1}
\end{figure}

\begin{figure}
\epsfbox{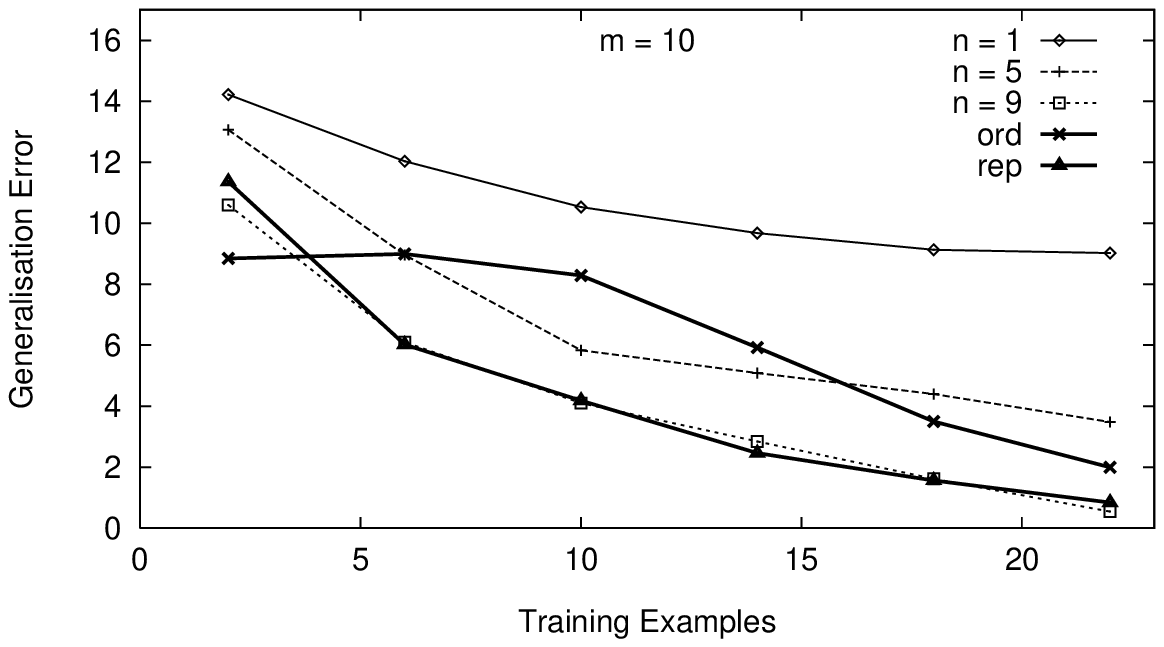}
\vspace{1mm}
\epsfbox{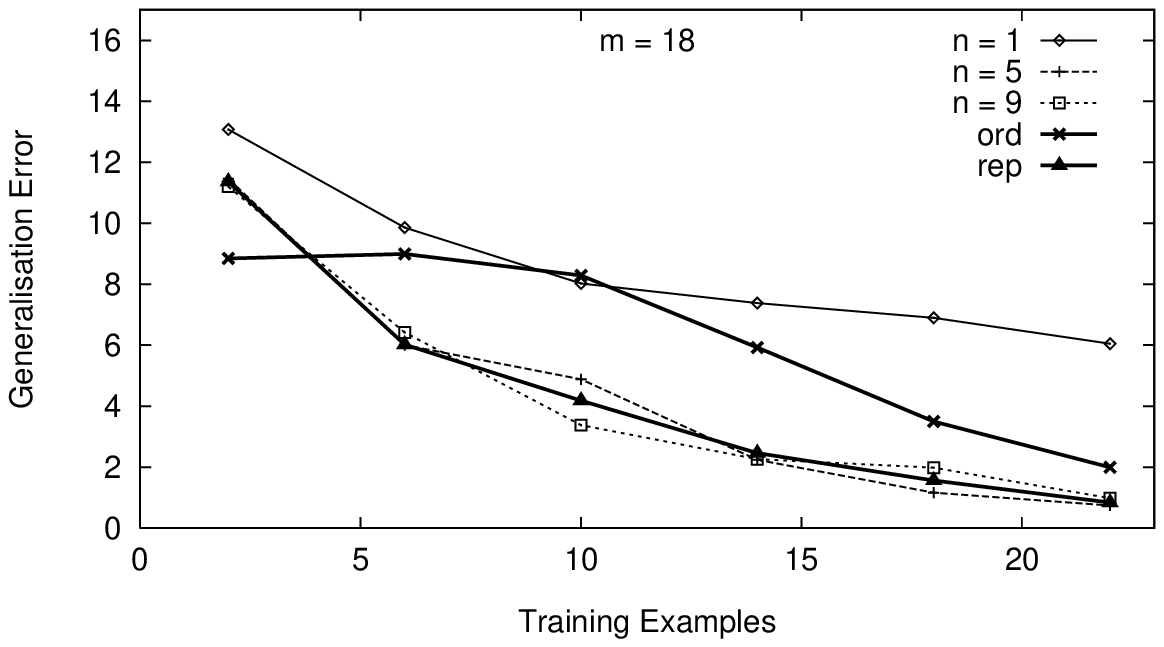}
\jcaption{Binary representation generalisation curves ($m=10$ and $m=18$).
$n$ is the number of functions sampled from the environment and $m$ is the
number of times each function was sampled.
The curves labelled ``rep'' and ``ord'' are the best representation learning
curve and ordinary learning curve respectively.}
{binfig2}
\end{figure}

The conclusion from this experiment is clear: for sufficiently large $n$ and
$m$ the generalisation performance of a learner is significantly improved if
it first learns an internal representation.

To some extent this experiment is quite artificial;
in particular, the environment 
of the learner was hardly a very realistic one: Boolean functions formed by 
composition of an arbitrary function with a binary neural network. Also the 
learning algorithm employed---exhaustive search---is clearly not a feasible 
one in practice. However as an in principle demonstration of the
effectiveness of representation learning it is fine. 
The next section is devoted to a discussion of two more
experiments that are closer to the kind of problems one sees in real-life 
machine learning situations (although they are still toy problems). 

\section{Learning Continuous Neural Network Representations}

In this section it is shown how gradient-descent may be used to train neural
networks to learn representations from $(n,m)$ 
samples\jfootnote{The methods presented here can be 
generalised quite easily to cover learning with any kind of continuous
function network, not just neural networks.}.
Two sets 
of experiments are performed. The first is a kind of very basic machine vision 
problem where the neural network is attached to a one-dimensional ``retina'' 
and has to learn an appropriate representation for an environment in which 
all the functions are invariant under {\em translations} in their inputs. 
In the second the functions in the environment are all 
invariant under permutations of the inputs, i.e they are all {\em symmetric} 
functions. These experiments were chosen because they are simple enough to be
performed within a reasonable amount of computer time, but complex enough to
exhibit the phenomena desired, and because it is easily verified that a
neural network representation exists for both experiments. 

Both experiments support the three main theoretical results of 
chapter \ref{repchap}. Firstly, in both experiments as the number of functions 
being learnt 
increases there is a corresponding decrease in the number of
examples required per function for good generalisation. Secondly, if the
number of functions learnt is sufficiently large then the resulting 
representation is a good one to use for learning further functions drawn 
from the same environment, and thirdly, the number of examples required to
ensure good generalisation is far less if one initially learns a good
representation.

\subsection{Gradient Descent for Representation Learning}

Given an
$(n,m)$ sample
$$
\z = 
\begin{matrix}
z_{11} & \hdots & z_{1m}\\
\vdots & \ddots & \vdots\\
z_{n1} & \hdots & z_{nm} 
\end{matrix}
$$
where $z_{ij} = (x_{ij},y_{ij})$, a representation space $\F$ and an output
function space $\G$, it is required that 
the learner find
a representation $f\in\F$ minimizing the {\em mean-squared} representation 
error, that is
$$
E^*_\G(f,\z) = \frac1n\sum_{i=1}^n \inf_{g\in\G} 
\frac{1}{m}\sum_{j=1}^m \(\comp{g}{f}(x_{ij}) - y_{ij}\)^2
$$
must be minimal.

$\F$ and $\G$ are assumed to consist of {\em differentiable families 
of neural networks}\jfootnote{The algorithm presented here does not actually
depend in any way on $\F$ and $\G$ being neural networks---it applies equally
well if $\F$ and $\G$ are any parameterised families of differentiable
functions.}. The most common procedure for training differentiable neural 
networks is to use some form of gradient descent algorithm (vanilla backprop,
conjugate gradient, etc) to minimize the error of the network on the sample
being learnt. For example, in ordinary learning the learner would receive
a single sample $\zv = ((x_1,y_1),\dots,(x_m,y_m))$ and would 
perform some form
of gradient descent to find a function $\comp{g}{f}\in\comp{\G}{\F}$ such that
\begin{equation}
\label{orderr}
E(\comp{g}{f},\zv) = \frac1m\sum_{i=1}^m\(\comp{g}{f}(x_i)-y_i\)^2
\end{equation}
is minimal. This procedure works because it is a relatively simple matter 
to compute the gradient, $\nabla_\w E(\comp{g}{f},\z)$, where $\w$
are the parameters (weights) of the networks in $\F$ and $\G$ (one simply
iteratively applies the chain rule to each layer in the network, moving from
the output layer back through to the first hidden layer. This is the
backpropagation algorithm). 

Applying this
method directly to the problem of minimising $E^*_\G(f,\z)$ above would mean
calculating the gradient $\nabla_\w E^*_\G(f,\z)$ where now $\w$
refers only to the parameters of $\F$. However, due to the presence of the
infimum over $\G$ in the formula for $E^*_\G(f,\z)$, calculating 
the gradient in this
case is much more difficult than in the ordinary learning scenario. 
Fortunately this problem has a simple solution. Recalling the
definition of the function space $\comp{\G^n}{\Fbar}$ from section
\ref{prelimdef} (it is just the set of
all sequences of functions $\comp{\gv}{\fbar} =
(\comp{g_1}{f},\dots,\comp{g_n}{f})$ where $g_1,\dots,g_n\in\G$ and $f\in\F$), 
define the {\em mean-squared error} of 
$\comp{\gv}{\fbar}\in\comp{\G^n}{\Fbar}$ with respect to an $(n,m)$ sample 
$\z$ as 
\begin{equation}
\label{gnferr}
E(\comp{\gv}{\fbar},\z) = \frac1n\sum_{i=1}^n\frac1m\sum_{j=1}^m
\(\comp{g_i}{f}(x_{ij})-y_{ij}\)^2
\end{equation}
where $\gv=(g_1,\dots,g_n)$. Note that $\frac1m\sum_{j=1}^m
\(\comp{g_i}{f}(x_{ij})-y_{ij}\)^2$ is just the mean-squared error of the 
network $\comp{g_i}{f}$ on the sample
$\zv_i=((x_{i1},y_{i1}),\dots,(x_{im},y_{im}))$, 
so that the mean-squared error
of $\comp{\gv}{\fbar}$ on $\z=(\zv_1,\dots,\zv_n)$ is nothing more than the 
average of the mean-squared error of each $\comp{g_i}{f}$ on $\zv_i$.
Clearly,
\begin{equation}
\label{gradlem}
\inf_{\comp{\gv}{\fbar}\in\comp{\G^n}{\Fbar}} E(\comp{\gv}{\fbar},\z) =
\inf_{f\in\F} E^*_\G(f,\z).
\end{equation}
If $\comp{\gv}{\fbar}$ is such that 
$$
\left|E(\comp{\gv}{\fbar},\z) - 
\inf_{\comp{\gv}{\fbar}\in\comp{\G^n}{\Fbar}} E(\comp{\gv}{\fbar},\z)\right|
\leq \ep, 
$$
equation \eqref{gradlem} implies, along with the fact that, by definition,
$E(\comp{\gv}{\fbar},\z) > E^*_\G(f,\z)$,
$$
\left|E^*_\G(f,\z) - \inf_{f\in\F} E^*_\G(f,\z)\right| \leq \ep.
$$
Thus if the learner finds a function $\comp{\gv}{\fbar}$ such that
$E(\comp{\gv}{\fbar},\z)$ is close to minimal then so too will
$E^*_\G(f,\z)$ be close to minimal. So instead of minimising
$E^*_\G(f,\z)$ the learner can equivalently minimise
$E(\comp{\gv}{\fbar},\z)$.
The advantage of this approach is that essentially the 
same techniques used for computing the gradient in ordinary learning can 
be used to compute the gradient of $E(\comp{\gv}{\fbar},\z)$.

A neural network of the form $\comp{\gv}{\fbar}$ is illustrated 
in figure \ref{gnnet}.
In the figure, $f$ belongs to a family  $\F$ consisting of neural networks 
with ten inputs, one hidden layer
with three nodes and one output layer with two nodes. $g_1, g_2$ and $g_3$
belong to a family $\G$ consisting of
networks with two input nodes, two hidden layers each with two hidden
nodes and one output node. Note that as the networks in $\G$ are composed 
with networks from $\F$, the dimensionality of the input of $\G$ (i.e the
number
of input nodes for the networks in $\G$ ) must equal the dimensionality of
the output of $\F$. The language introduced in the theoretical chapters is
also illustrated in the diagram, with the input space denoted by $X$,  
the internal space by $V$ and the output space of the network
denoted by $A$. 

\begin{figure}
\begin{center}
\leavevmode
\epsfbox{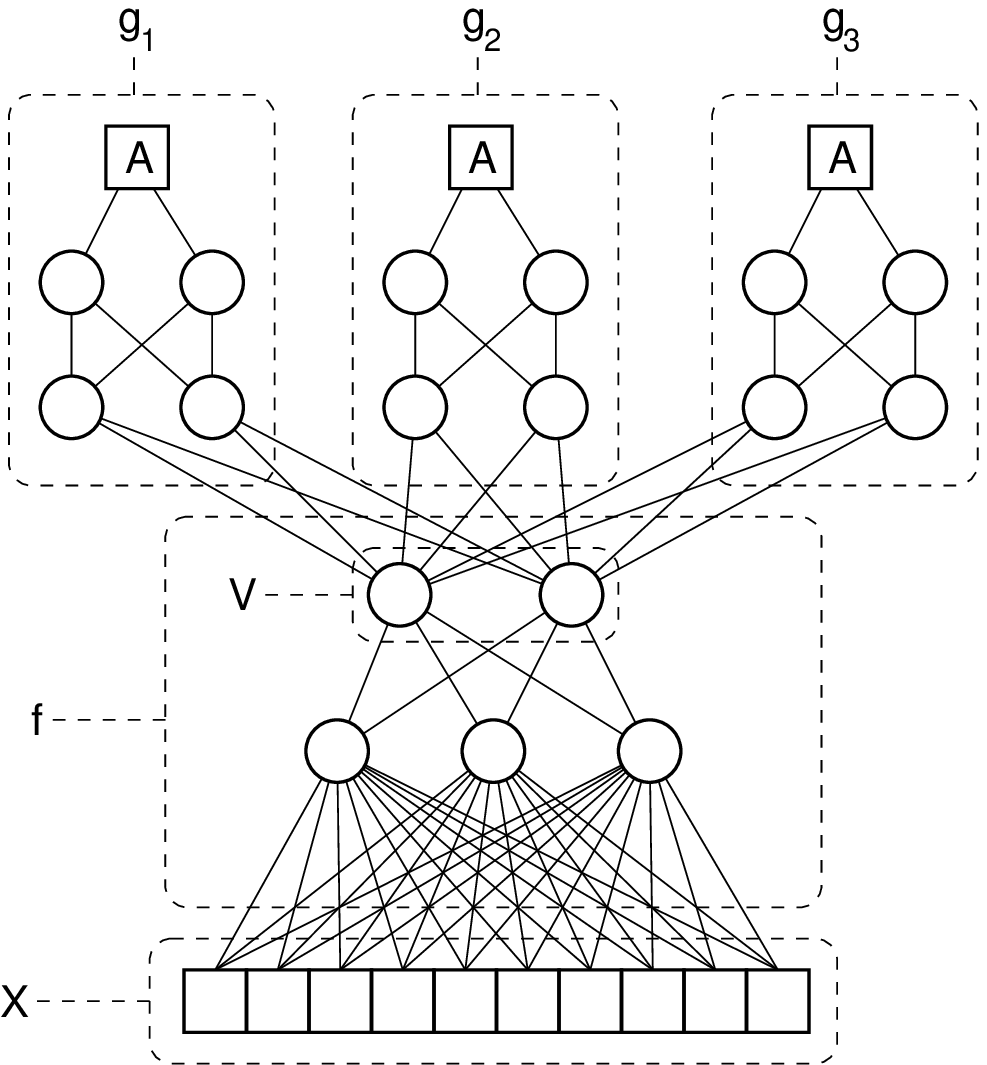}
\jcaption{A neural network for representation learning.}
{gnnet}
\end{center}
\end{figure}

Consider now the problem of 
computing the derivative of $E(\comp{\gv}{\fbar},\z)$
with respect to a weight in the $i^{th}$ output network $g_i$. Denoting the 
weight by $w_i$ and recalling equation \eqref{gnferr},
$$
\frac{\partial}{\partial w_i} E(\comp{\gv}{\fbar},\z) = 
\frac1n\frac{\partial}{\partial w_i}\frac1m\sum_{j=1}^m 
\(\comp{g_i}{f}(x_{ij}) - y_{ij}\)^2
$$
which is just $\frac1n$ times 
the derivative of the ordinary learning error \eqref{orderr}
of $\comp{g_i}{f}$
on sample $\zv_i=\((x_{i1},y_{i1}),\dots,(x_{im},y_{im})\)$ with respect to
the weight $w_i$. This can be computed using any standard formula for the
derivative (e.g backpropagation). 

Alternatively, if $w$ is a weight in the representation network $f$ then
$$
\frac{\partial}{\partial w} E(\comp{\gv}{\fbar},\z) =
\frac1n\sum_{i=1}^n \frac{\partial}{\partial w}\frac1m\sum_{j=1}^m
\(\comp{g_i}{f}(x_{ij})-y_{ij}\)^2
$$
which is simply the {\em average} of the derivatives of the ordinary 
learning errors over all the samples $(\zv_1,\dots,\zv_n)$ 
and hence can also be computed using standard formulae.

\subsection{Representation Learning Using Backpropagation}
\label{backproprep}
In this section it is shown how the
backpropagation learning algorithm (Rumelhart et.\ al.\ (1986)\label{R1}) 
can be turned into an algorithm for learning
representations. It is assumed that the architectures of the networks 
in $\F$ and $\G$ are known\jfootnote{Architecture selection is still a 
major unsolved problem in Neural Network research and will not be addressed
here. However it is argued in chapter \ref{flearnchap} 
that the information provided by $(n,m)$ samples is
much more appropriate for hierarchical construction of neural networks than is
the information provided by a single sample. Some ideas are also presented
for how this information may be used to ``grow'' a representation of an
appropriate size.}
and that the learner has been supplied with
an $(n,m)$ sample 
$$
\z = 
\begin{matrix}
(x_{11},y_{11}) & (x_{12},y_{12}) & \hdots & (x_{1m},y_{1m})\\
(x_{21},y_{21}) & (x_{22},y_{22}) & \hdots & (x_{2m},y_{2m})\\
\vdots & \vdots & \ddots & \vdots\\
(x_{n1},y_{n1}) & (x_{n2},y_{n2}) & \hdots & (x_{nm},y_{nm})\\
\end{matrix}
$$
Note that it is not necessary for the architectures of all the output networks
to be the same.
\begin{enumerate}
\item Construct the network $\comp{\G^n}{\Fbar}$ as shown for example in figure 
\ref{gnnet} and randomly initialise all the weights in the representation and
output networks.
\item For each subsample $(x_{ij},y_{ij})$:
\begin{itemize}
\item Initialise the input nodes 
of $f$ to the components of
$x_{ij}$ and perform a {\em forward pass} through $f$ to yield $f(x_{ij})$
at $f$'s output nodes. 
\item Set the input nodes of the $i^{th}$ output network, $g_i$ to the 
value of $f$'s output nodes (note that the number of output nodes from
$f$ must equal the number of input nodes to each of the $g's$).
\item Perform a forward pass through the $i^{th}$ output network $g_i$ to
yield the network's output $o_{ij} = g_i(f(x_{ij}))$ on input $x_{ij}$.
\item Compute the error $(o_{ij} - y_{ij})^2$.
\item Backpropagate this error information in the usual way through the output
network $g_i$ {\em and} the  representation network $f$. That is, for the
purpose of backpropagating the error values, use the same formulae that
would be used if the network being trained was $\comp{g_i}{f}$.
\item Update the gradient for each weight in {\em both} $f$ and $g_i$ 
based on the backpropagated errors.
\end{itemize}
\vspace{2mm}
\item The total error is the sum of all the individual $(o_{ij}-y_{ij})^2$
errors. This should be divided by $m\times n$ to yield the average error.
\item Update the weights using any gradient procedure (vanilla
gradient descent, conjugate gradient descent, etc) and start all over
again at (2).
\end{enumerate}

Note that for each input $x_{ij}$, error information is only backpropagated
through the output network $g_i$. It may well be the case that the desired
output can easily be computed for the other output networks too, in which
case ordinary backpropagation can be used to train the entire network.

\subsection{Experiment One: Learning Translation Invariance}
\label{exp1}
In this first experiment a neural network was trained to perform a
very simple ``machine vision'' task in which it had to learn to
recognise different objects drawn on its one dimensional retina, 
no matter where they were drawn. Thus all the tasks in the environment 
of the neural network were {\em translationally invariant}.

The input space $X$ consisted of a ten-pixel, one-dimensional ``retina''
in which all the pixels could be either on (1) or off (0) (so in fact
$X=\{0,1\}^{10}$). However the network did not see all possible input vectors
during the course of its training, the only vectors with a non-zero probability
of appearing in the training set 
were those consisting of from one to four active adjacent pixels placed
somewhere in the retina (see figure \ref{objects}). Thus there were 40 input
vectors in all:
$4$ objects times $10$ possible positions
(for asthetic reasons the objects were allowed to ``wrap around'' at the 
edge so that the retina should more appropriately be viewed as a circle 
rather than a line.)

\begin{figure}
\begin{center}
\leavevmode
\epsfbox{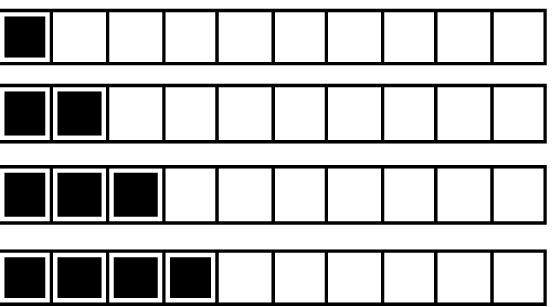}
\jcaption{The inputs seen by the learner in experiment one consisted of
these four and any of their translates (including wrapping around at the
edge).}
{objects}
\end{center}
\end{figure}

The functions in the environment of the network consisted of
all possible {\em translationally invariant} Boolean functions over the 
input space (except the trivial ``constant 0'' and ``constant 1''
functions). The requirement of translation invariance means that the
environment consisted of just 14 different functions---all the Boolean
functions on four objects (of which there are $2^4=16$) less the two 
trivial ones (the functions are shown in figure \ref{funfig}). Thus the
environment was highly restricted, both in the 
number of different input vectors seen---40 out of a possible 1024---and
in the number of different functions to be learnt---14 out of a possible 
$2^{1024}$. It is this kind of principled restriction of
the environment of a learner that makes generalisation possible (and of 
course without it generalisation is impossible). $(n,m)$ samples were 
generated from this environment by firstly choosing $n$ functions (with 
replacement, so a function could be chosen more than once) uniformly from 
the fourteen possible, and then choosing $m$ input vectors (with replacement 
again), for each function, uniformly from the 40 possible input vectors.

The architecture of the network was similar to the one shown in figure 
\ref{gnnet}, the only difference being that the output networks $g\in\G$ for
this experiment had only one hidden layer, not two. The network in the
figure is for learning $(3,m)$ samples (it has $3$ output networks), in
general for learning $(n,m)$ samples the network will have $n$ output
networks. The network architecture 
is more complex than is needed to solve the learning problem,
however it was deliberately made that way to demonstrate
that the learning algorithm works on complex networks.

Each node in the network computed its activation by applying the standard sigmoid 
squashing function to a weighted sum of its inputs. Thus if $a$ is the 
activation of the node, $\wv=(w_1,\dots,w_n)$
its weight vector, $w_T$ its threshold, and 
$\xv=(x_1,\dots,x_n)$ is the vector of inputs to the node, then
$$
a = \frac{1}{1+e^{-\(\sum_{i=1}^n w_ix_i+w_T\)}}
$$

\begin{figure}
\begin{center}
\leavevmode
\epsfbox{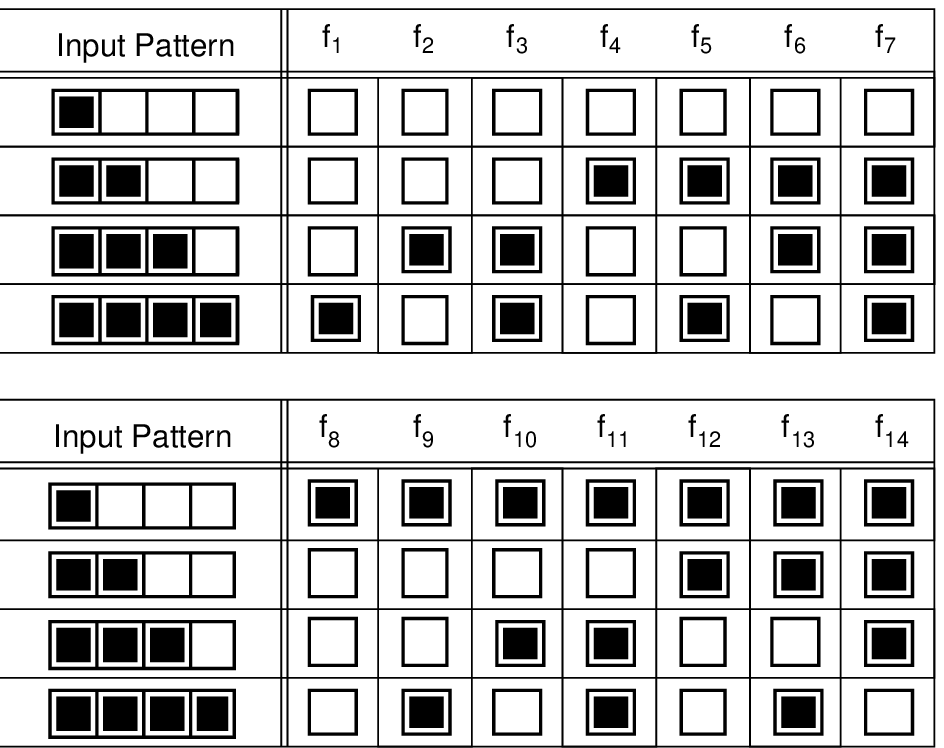}
\jcaption{The functions in the environment of the learner: all
{\em translationally invariant} Boolean functions on the restricted set of
inputs shown in figure \ref{objects}, excluding the two constant functions.}
{funfig}
\end{center}
\end{figure}

The network was trained on $(n,m)$ samples with $n$ ranging from $1$ to
$21$ in steps of four and $m$ ranging from $1$ to $151$ in steps of
$10$. Conjugate-gradient descent was used with
exact line search with the gradients for each weight computed 
according to the algorithm outlined in the previous section. Training 
was halted if the mean-squared error (equation \eqref{gnferr}) 
fell below $10^{-6}$ {\em or} the $L^\infty$ error
(i.e the maximum absolute difference between the desired output 
and the actual output for any of the training examples and any of the
output networks) fell 
below $0.01$.  These are fairly stringent halting 
requirements but as a 
solution is known to exist they are not unreasonable.
If the relative reduction in error over $5$ training iterations was less 
than $0.01\%$, the network was deemed to have struck a local minimum
and training was restarted with a new randomly generated set of weights.
Weights were chosen uniformly in the interval $[-1,1]$.
All weights were clipped when their absolute values exceeded $20$, except 
for the thresholds which were permitted to grow to $80$, although they never 
got that far in any of the simulations. At any iteration of the algorithm,
conjugate gradient descent was only performed on the subspace of weights
excluding the clipped weights.
However, if the gradient 
vector at any stage indicated that a reduction in absolute value of 
a clipped weight would occur if a step was taken in the downhill direction,
then the clipped weight was reintroduced 
into the optimization process. This method prevented the network heading 
blindly for the local minimum that often exists ``at infinity'' due 
to the asymptotic behaviour of the sigmoid squashing function. The clipping
of a weight for several iterations and then its subsequent reintroduction
into the optimisation process was not an uncommon event, and occasionally
led to the discovery of a global minimum that would not otherwise have been
discovered. However the main utility of this technique was its early
termination of unfruitful directions of search.

Once the network had sucessfully learnt the $(n,m)$ sample its generalization
ability was tested on all $n$ functions in the training set.
In this case the generalisation
error (i.e {\em true} error) could be computed exactly by calculating the
network's output (for all $n$ functions) for each of the $40$ input vectors,
and comparing the result with the desired output. Both the mean-squared and
$L^\infty$ generalisation errors were computed.

In an ordinary
learning situtation the generalisation error of a 
network would be plotted as a function of $m$, the number of examples 
in the training set, resulting in a {\em generalisation curve}.
For representation learning there are two parameters $m$ and $n$ so the curve 
becomes a {\em generalisation surface}.
Plots of the generalisation surface are shown in figure \ref{tplot}
for three independent simulations. 
All three cases support the theoretical 
result that the number of examples $m$ required for good generalisation
decreases with increasing $n$. This is most easily seen by observing the
behaviour of the surface for fixed values of $m$. 
Although it is out of range and not shown in the plots, the number of examples
required for good generalisation was around $100$--$150$ when only one
function from the environment was learnt ($n=1$, the ordinary learning
scenario), while the plots show that less than $21$ examples per function
are required if $n\geq 17$. 

\begin{figure}
\begin{center}
\leavevmode
\vspace{-7mm}
\epsfbox{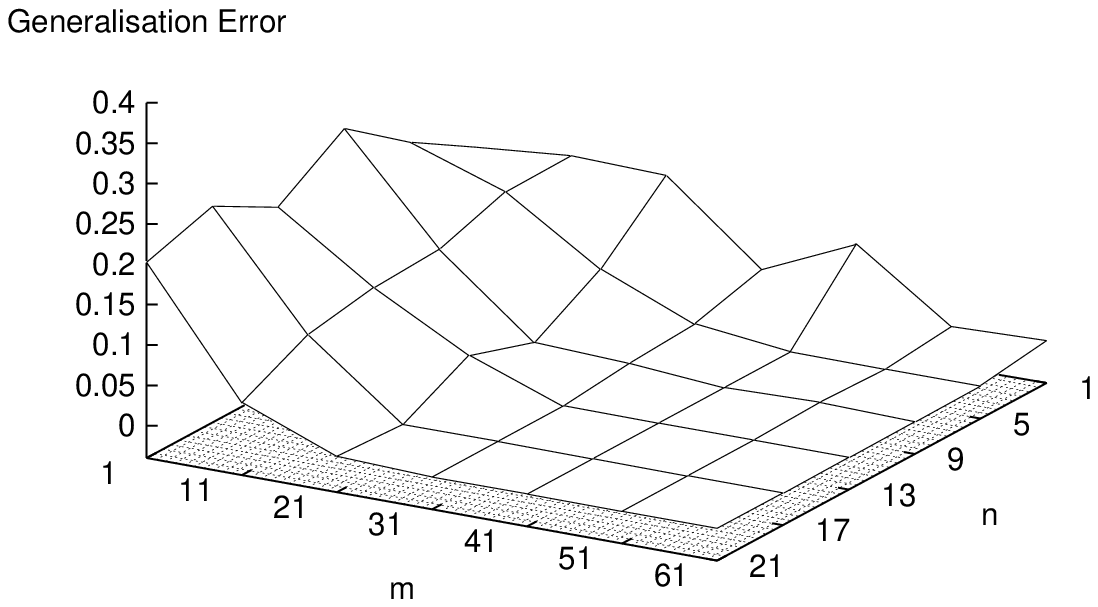}
\vspace{-2mm}
\epsfbox{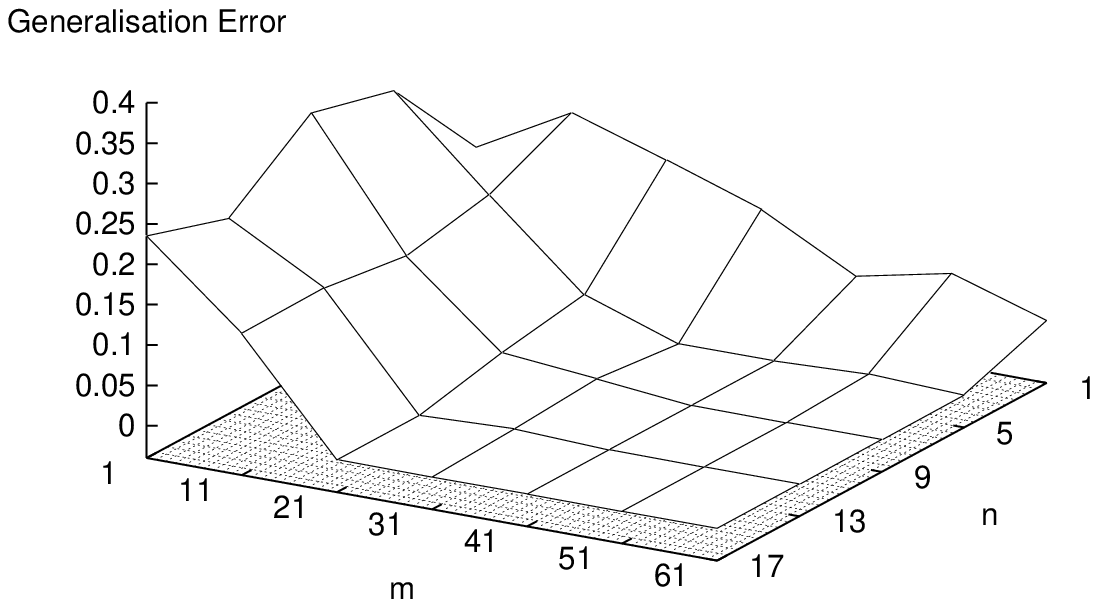}
\vspace{-7mm}
\epsfbox{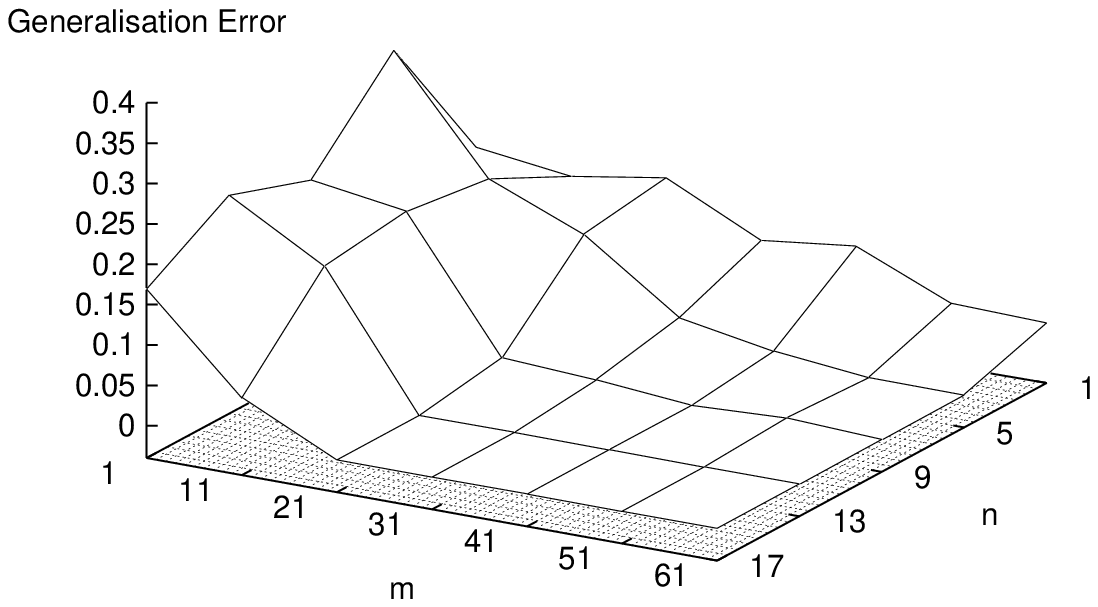}
\jcaption{Mean-squared generalisation error as a function of 
$n$---the number of functions
learnt from the environment---and $m$---the number of examples per function,
for three independent simulations from the first experiment.}
{tplot}
\end{center}
\end{figure}

For $(n,m)$ samples that led to a generalisation error of less
than  $0.01$, the representation network $f$ was extracted and tested for 
its {\em true error}, where this is defined as in
equation \eqref{exploss} and in the current framework translates to
$$
E^*_\G(f,Q) = \frac1{560}\sum_{i=1}^{14}\inf_{g\in\G} \sum_{j=1}^{40}
\(\comp{g}{f}(x_j) - f_i(x_j)\)^2 
$$
where $x_1,\dots,x_{40}$ are the $40$ input vectors seen by the learner
and $f_1,\dots,f_{14}$
are all the functions in the environment. 
$E^*_\G(f,Q)$ measures how useful the representation $f$ is for learning all
functions in the environment, not just the ones used in generating the
$(n,m)$ sample $f$ was trained on. To measure $E^*_\G(f,Q)$, entire training
sets consisting of 40 input-output pairs were generated for each of the 14
functions in the environment. The architecture used for learning
$(n,m)$ samples was reused,
only this time with just one output network $g$ and $f$ as
the representation network. The network was taught each of the fourteen
functions in turn by keeping the weights of $f$ fixed and using
conjugate-gradient descent to set the weights of $g$ so that the combination
$\comp{g}{f}$ had minimal error on the training set. To be (nearly) certain
that a minimal solution had been found for each of the functions, learning
was started from $32$ different random weight initialisations for 
$g$ (this number was chosen so that the CM5 could
perform all the restarts in parallel) and the best result from all 32
recorded. For each
$(n,m)$ sample giving perfect generalisation, $E^*_\G(f,Q)$ was calculated and then
averaged over all $(n,m)$ samples with the same value of $n$, 
and finally averaged
over all three simulations, to give an indication
of the behaviour of $E^*_\G(f,Q)$ as a function of $n$. This is
plotted in figure \ref{E*}, along with the $L^\infty$ representation error
for the three simulations (i.e, the maximum error over all 14 functions and
all 40 examples and over all three simulations). Qualitatively the curves
support the theoretical conclusion that the representation error should
decrease with an increasing number of tasks $n$ in the $(n,m)$ sample.
However,  
note that the representation error is very small, even
when the representation is derived from learning only one function from the
environment. This can be explained as follows. 
For a representation to be a good one for learning in this
environment it must be translationally invariant and distinguish all
the four objects it sees (i.e.\ have different values on all four objects). 
For small values of $n$, to achieve perfect
generalisation the representation is forced to be translationally invariant
and so half of the problem is already solved. However, depending upon the
particular functions in the $(n,m)$ sample, the representation may not have
to distinguish all four objects, for example it may map two objects
to the same element of $V$ if none of the functions
in the sample distinguish those objects (a function distinguishes two
objects if it has a different value on those two objects). However, because the
representation network is continuous it is very unlikely that it will
map different objects to
{\em exactly} the same element of $V$---there will always be slight
differences. When the representation is used to learn a function that does
distinguish the objects mapped to nearly the same element of $V$,
often an output network $g$ with sufficiently large weights can be found to
amplify this difference and produce a function with small error. This is why
a representation that is simply translationally invariant does quite well in
general. This argument is supported by a plot in figure \ref{outplot}
of the representation's behaviour for
$(n,m)=\{(1,1),(1,31),(1,61),(1,91),(1,101),(1,111)\}$
for the first simulation. The four different symbols marking the points in the
plots correspond to the four different input objects. The ordinary 
generalisation error of the network for  $m=91,101$ and $111$ was zero. For
the $(1,91)$ plot the three and four pixel objects are well separated by the
representation while the one and two pixel objects are not separated at all
(they are both crammed into the bottom left-hand corner of the plot). This
representation will not be a good one for learning any function that
distinguishes the one and two pixel objects. A similar conclusion holds for
the $(1,101)$ and $(1,111)$ samples, except that a closer look reveals that
there is a slight separation between the representation's output for the one
and two pixel objects. This separation can be exploited to learn a function
that distinguishes the two objects.

For comparison, the behaviour of representations learnt on $(5,m)$ and
$(21,m)$ samples is shown in figure \ref{outplot1}. For large enough $m$
these representations do distinguish well all four objects.

The behaviour of a typical perfect
representation is shown pictorially in figure \ref{repfig}, and its weights 
in table \ref{repweights}. Notice that the
representation is effectively operating as an ``object detector'' for the
four objects in the environment.

\begin{figure}
\begin{center}
\leavevmode
\epsfbox{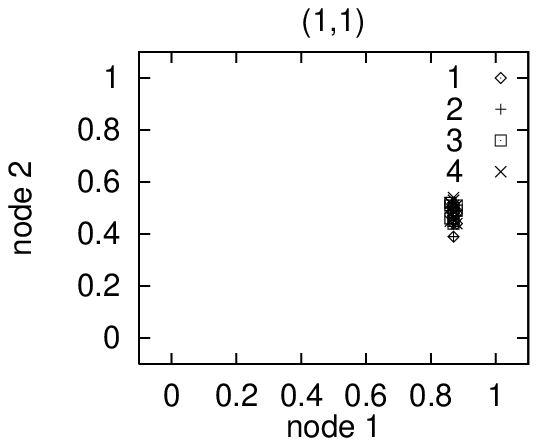}
\epsfbox{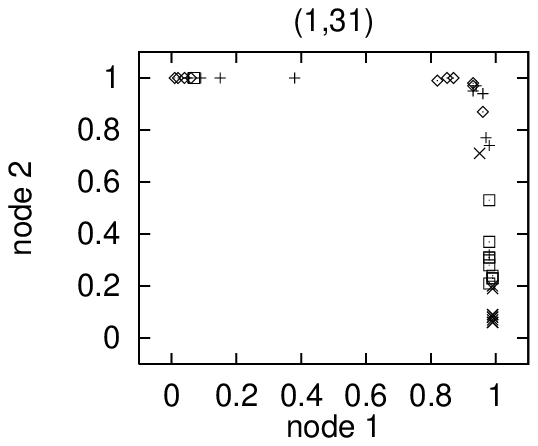}
\vspace{2mm}
\epsfbox{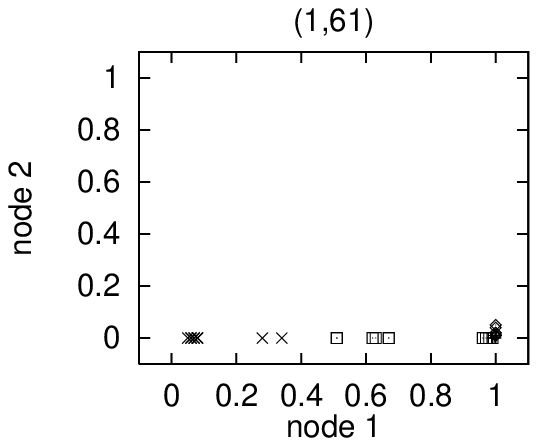}
\epsfbox{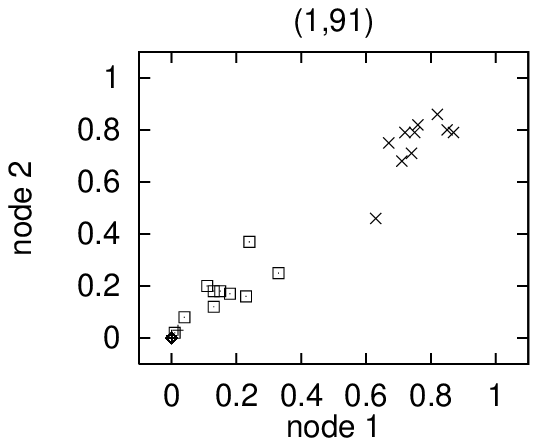}
\vspace{2mm}
\epsfbox{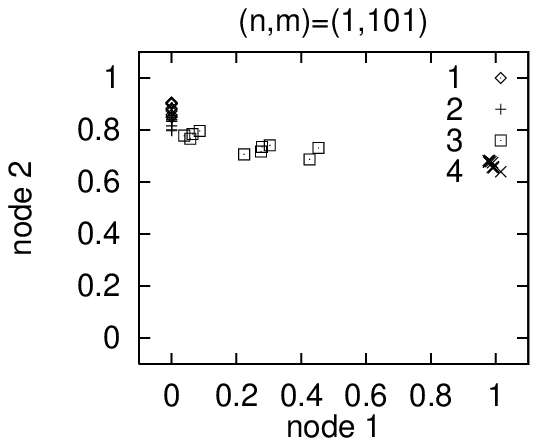}
\epsfbox{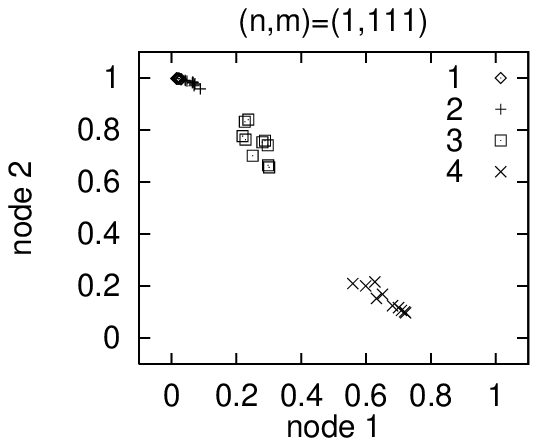}
\jcaption{Plots of the representation's output for $(1,m)$ samples from
the first experiment.}
{outplot}
\end{center}
\end{figure}

\begin{figure}
\begin{center}
\leavevmode
\epsfbox{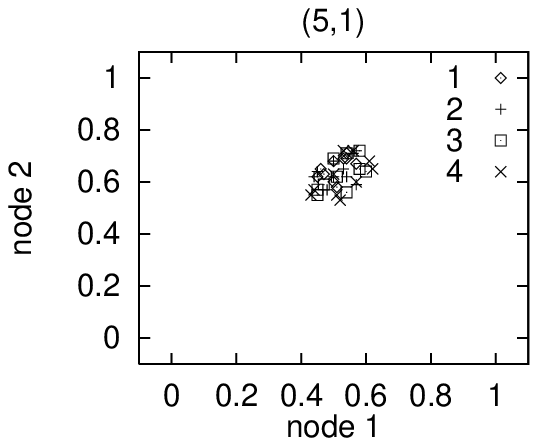}
\epsfbox{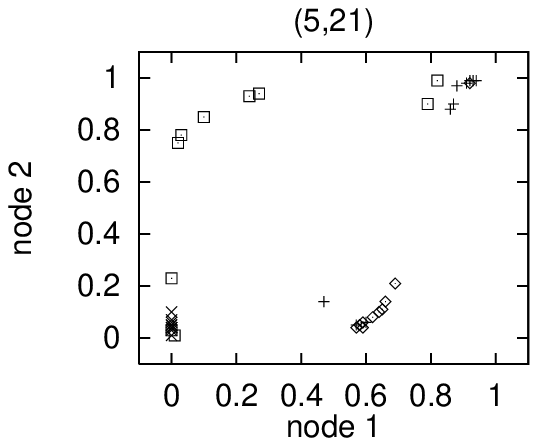}
\vspace{2mm}
\epsfbox{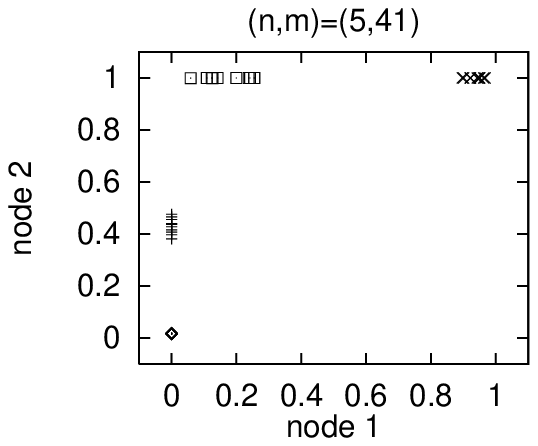}
\epsfbox{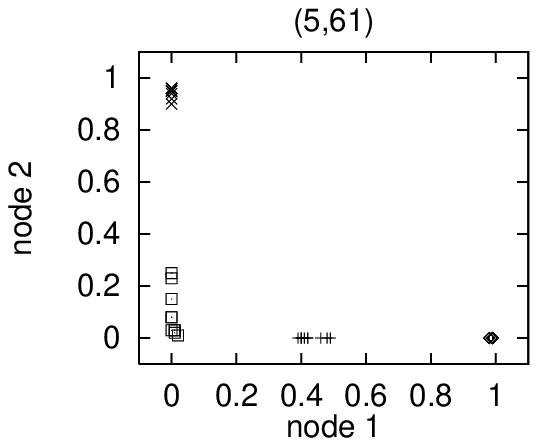}
\vspace{2mm}
\epsfbox{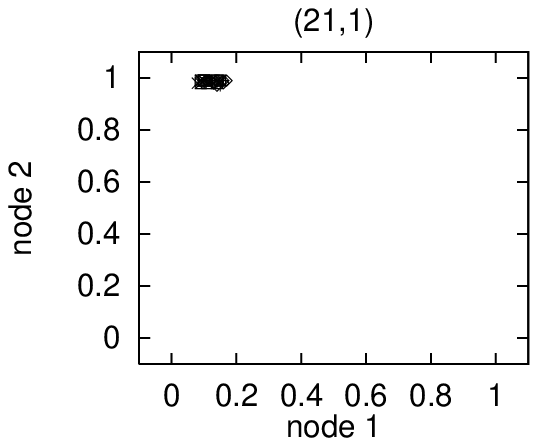}
\epsfbox{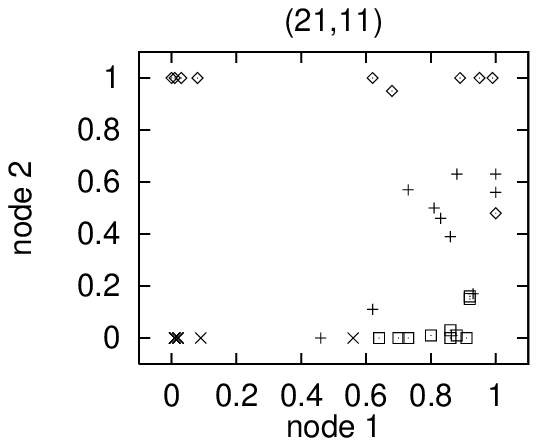}
\vspace{2mm}
\epsfbox{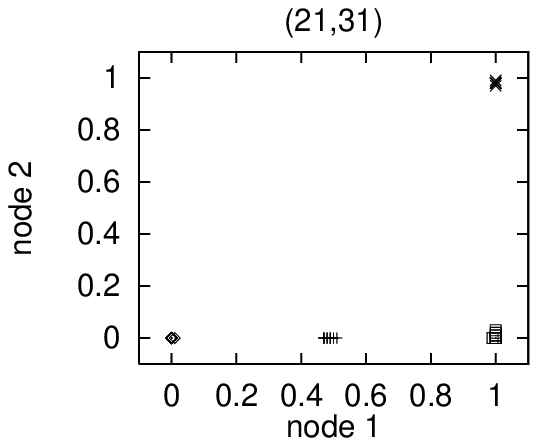}
\epsfbox{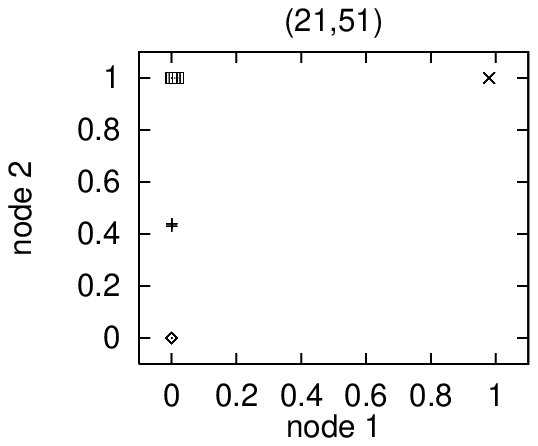}
\jcaption{Plots of the representation's output for $(5,m)$ and $(21,m)$
samples from the first experiment.}
{outplot1}
\end{center}
\end{figure}

\begin{figure}
\begin{center}
\leavevmode
\epsfbox{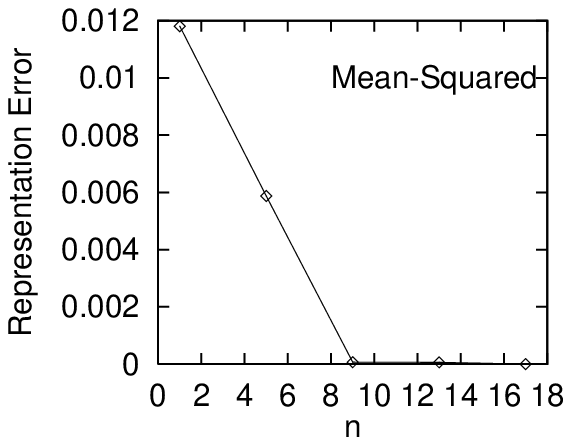}
\epsfbox{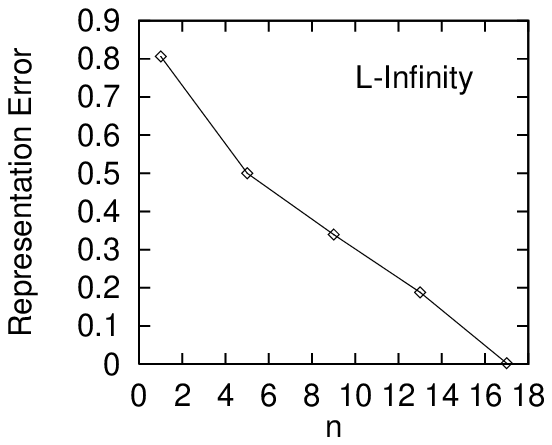}
\jcaption{Mean-Squared and $L^\infty$ representation error curves for the
first experiment.}
{E*}
\end{center}
\end{figure}

\begin{figure}
\begin{center}
\leavevmode
\epsfbox{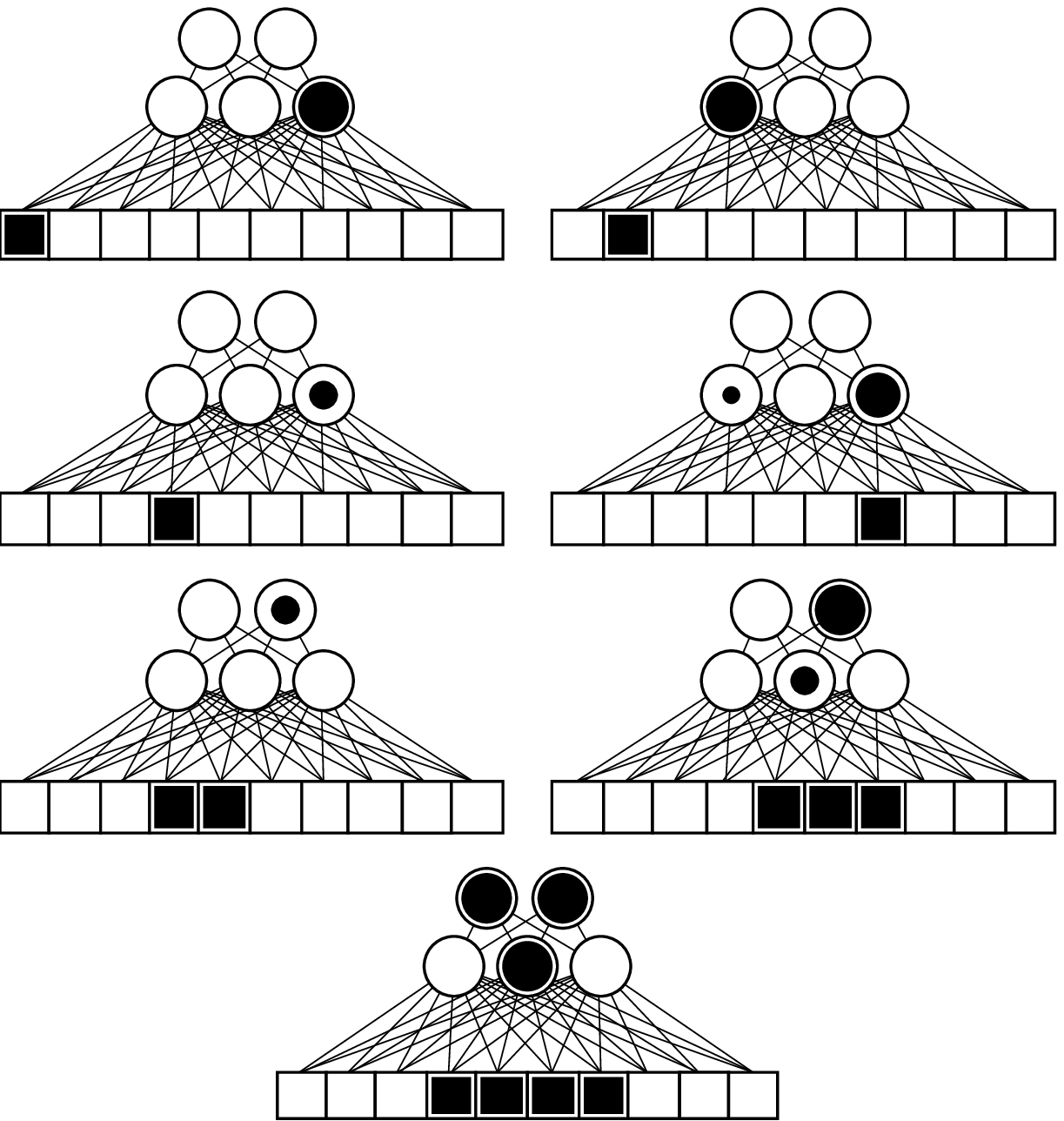}
\jcaption{Behaviour of a perfect representation for the first experiment.
The representation's output
node activations were invariant under translations
in the input object (to within about $1\%$). The hidden layer activations
were also invariant for the two, three and four pixel objects, and there
were four different hidden layer activation vectors for the one pixel
object, depending on the position of the pixel in the input. Representative
input vectors and the corresponding activations are shown here. This
representation was the result of learning a $(21,21)$ sample.}
{repfig}
\end{center}
\end{figure}

\begin{table}
\begin{center}
\begin{tabular}{|c|rrrrrrrrrrr|}
\multicolumn{12}{c}{Layer 1} \\ [2pt] \hline
Node & $w_0$ & $w_1$ & $w_2$ & $w_3$ & $w_4$ & $w_5$ & $w_6$ & $w_7$ & $w_8$ &
$w_9$ & $w_T$ \\[2pt]\hline
0  & -7.7 & 0.8 & -7.0 & -5.4 & -3.2 & -4.3 & -1.8 & -4.5 & -4.8
& -3.1 & 0.5 \\ [2pt] 
  1  & 5.4 & 5.6 & 5.8 & 5.3 & 5.7 & 6.0 & 5.6 & 5.5 & 5.5 & 5.5 &
-16.8 \\ [2pt]
  2  & -9.2 & -13.1 & -8.4 & -10.8 & -8.8 & -8.0 & -9.6 & -8.8 &
-9.3 & -9.2 & 10.5 \\ [2pt] \hline
\end{tabular}
\vspace{10mm}
\begin{tabular}{|c|rrrr|}
\multicolumn{5}{c}{}\\ 
\multicolumn{5}{c}{Layer 2} \\[2pt]\hline
Node & $w_0$ & $w_1$ & $w_2$ & $w_T$ \\[2pt]\hline
0 & -2.3 & 20.0 & -20.0 & -16.2 \\ [2pt]
1 & -5.9 & 15.4 & -17.6 & -0.3 \\[2pt]\hline
\end{tabular}
\jcaption{Weight values for the representation in figure \ref{repfig}.}
{repweights}
\end{center}
\end{table}

\paragraph{Representation vs. No Representation.}

The other principal
purpose of learning a representation is that it should greatly
reduce the number of examples required to learn future tasks from the
same environment. This was experimentally verified by taking a
representation $f$ known to be perfect for the environment in experiment one, 
and using it to
learn all the functions in the environment in turn. Hence the hypothesis
space of the learner was 
$\comp{\G}{f}$, rather than the full space
$\comp{\G}{\F}$. By way of comparison all the functions in the environment
were also learnt using the full space. The generalisation curves (i.e. the
generalisation error as a function of the number of examples in the training
set) were calculated for all 14 functions in each case. The generalisation
curves for all the functions were very similar; the first four are
plotted in figure 
\ref{repcomp}, for learning with a representation (Gof in the graphs) 
and without (GoF). The full curves for learning without a representation are
also plotted in figure \ref{repcompj}.
These curves are the average of 32
different simulations obtained by using different random starting points 
for the weights in $g$ (and $f$ when using the full space to learn). 
Learning with a
good representation is clearly far superior to learning without.

\begin{figure}
\begin{center}
\leavevmode
\epsfbox{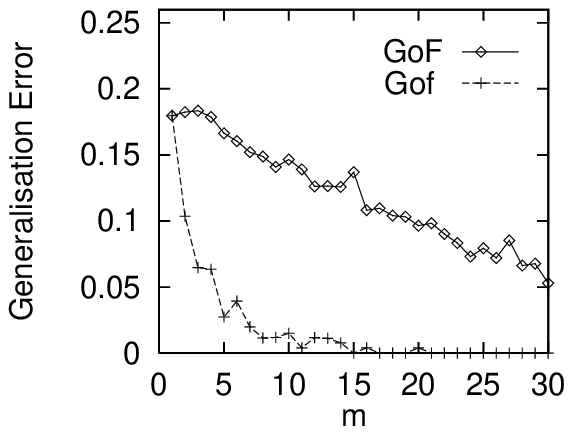}
\epsfbox{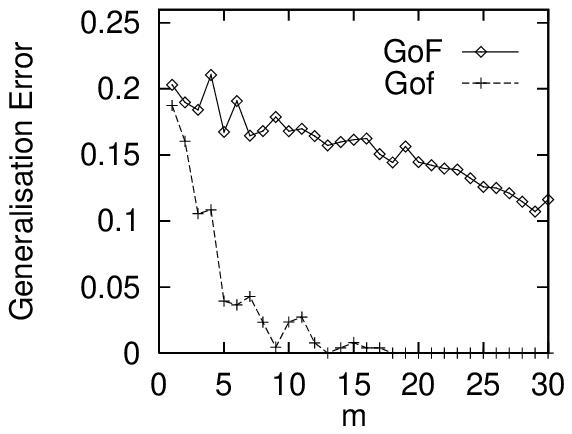}
\vspace{5mm}
\epsfbox{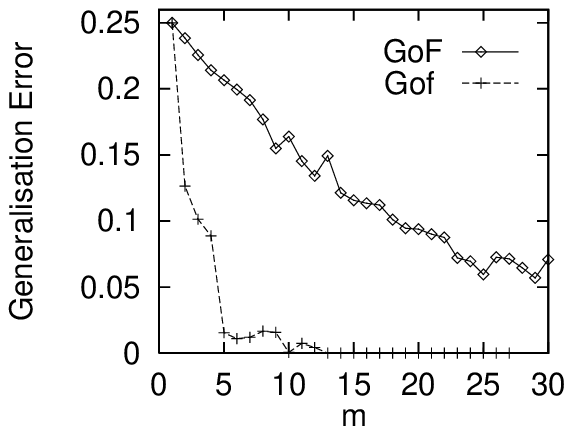}
\epsfbox{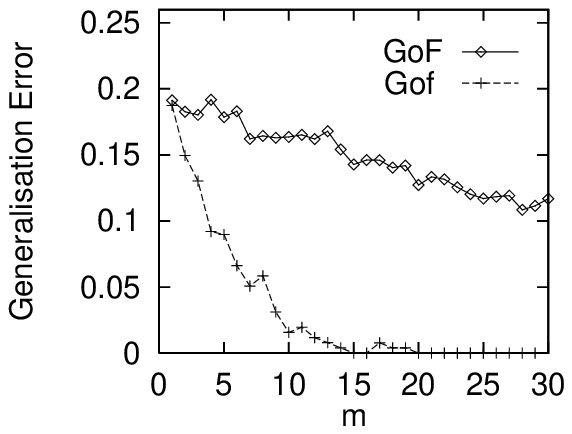}
\jcaption{Generalisation curves for learning with a representation (Gof) 
vs. learning without (GoF) for
the first four functions from the environment in the first experiment.}
{repcomp}
\end{center}
\end{figure}

\begin{figure}
\begin{center}
\leavevmode
\epsfbox{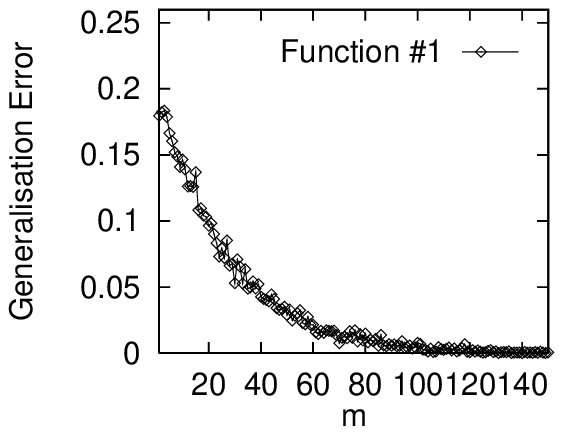}
\epsfbox{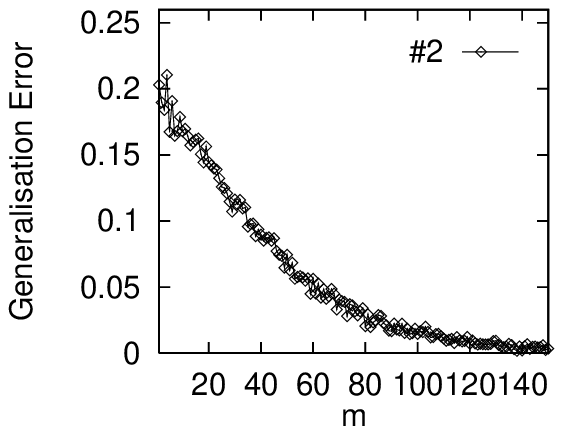}
\vspace{5mm}
\epsfbox{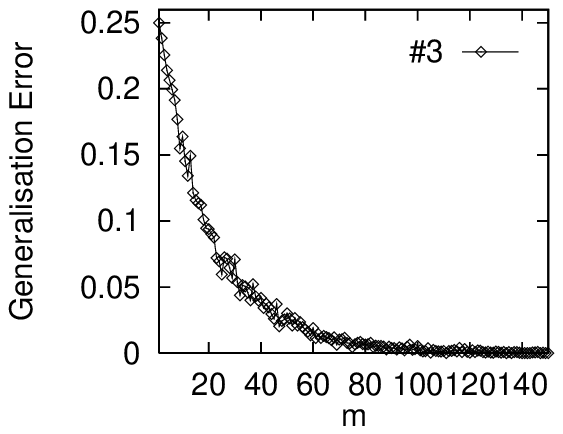}
\epsfbox{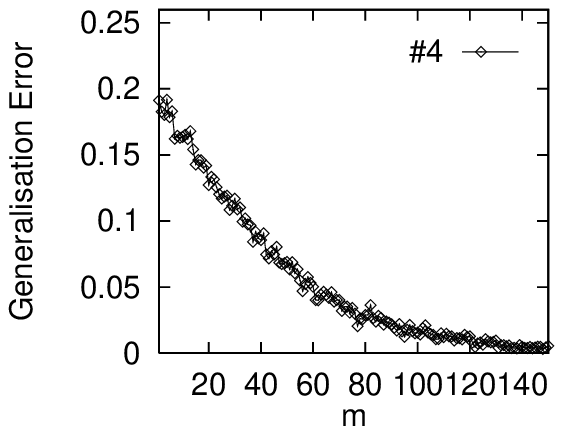}
\jcaption{Full generalisation curves for learning without a representation
for the first four functions from the environment of the first experiment.}
{repcompj}
\end{center}
\end{figure}

\subsection{Experiment Two: Learning Symmetric Functions}

The second neural network representation learning experiment was very
similar to the first, the only difference being in the architectures of the
networks and in the environment. This time the environment was chosen to
consist of a subset of all the {\em symmetric} Boolean functions on $10$
variables. The symmetric Boolean functions are those invariant under changes
in the order of their inputs, or equivalently, the functions that depend
only upon the number of ``1's'' in the input, not where the ``1's'' are
located. The
number of different input vectors seen by the learner was also restricted to
consist only of those containing between one and four ``1's''.
The probability distribution $P$ on the input
space was the same for all functions in the environment: input vectors were
selected
by firstly choosing uniformly a number from 1 to 4, and then placing that many
active pixels randomly within the input vector. Once again the two trivial
Boolean functions were excluded so that the environment consisted of 14
functions in all. Clearly, a representation exists for this environment---one
that simply counts the number of active pixels in its input will do. 
 
The neural networks in the representation space $\F$ had no hidden layers,
ten input nodes and three output nodes. The output networks $\G$
also had no hidden layers, three input nodes (matching the number of the
output nodes for networks in $\F$) and one output node.
The networks were trained  on $(n,m)$ samples with $n$ ranging from 1 to 21
in steps of four and $m$ ranging from 1 to 171 in steps of 10. The
same training  and testing 
regime was followed as in the previous experiment. The
(mean-squared) generalisation error as a function of $m$ and $n$ is plotted
in figure \ref{symfig} for three independent simulations. 
Once again these graphs clearly show that
fewer examples are required for good generalisation if many functions are
learnt simultaneously from the same environment. 

\begin{figure}
\begin{center}
\leavevmode
\vspace{-7mm}
\epsfbox{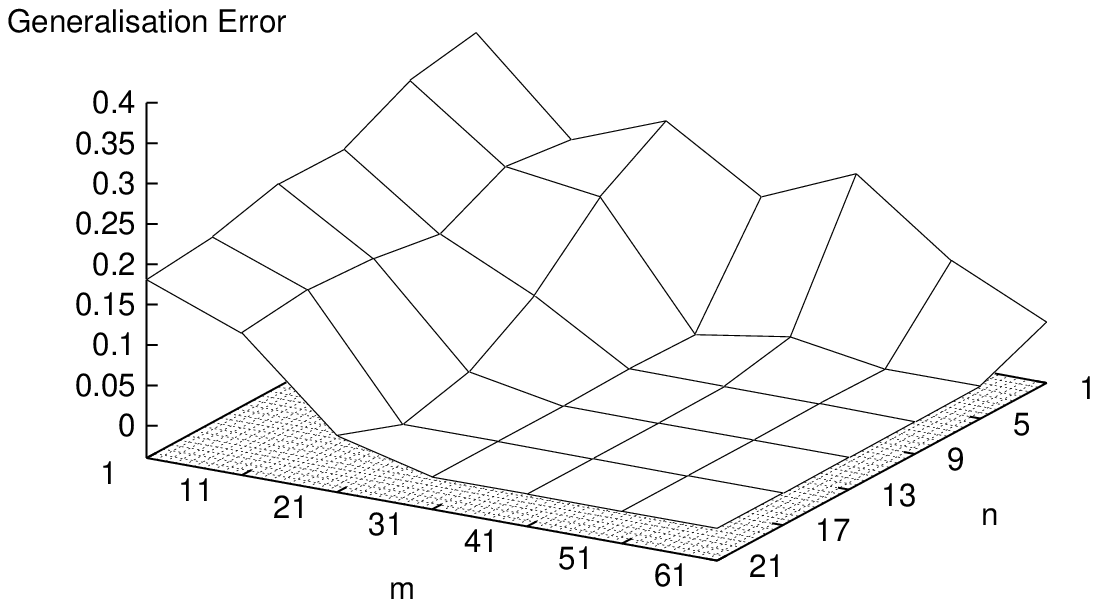}
\vspace{-2mm}
\epsfbox{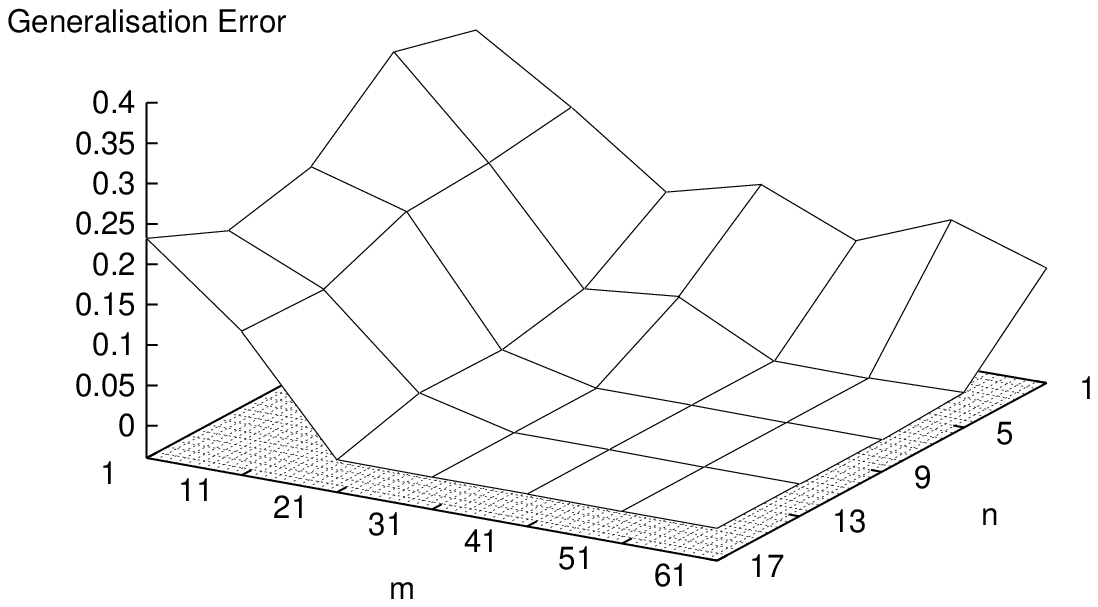}
\vspace{-7mm}
\epsfbox{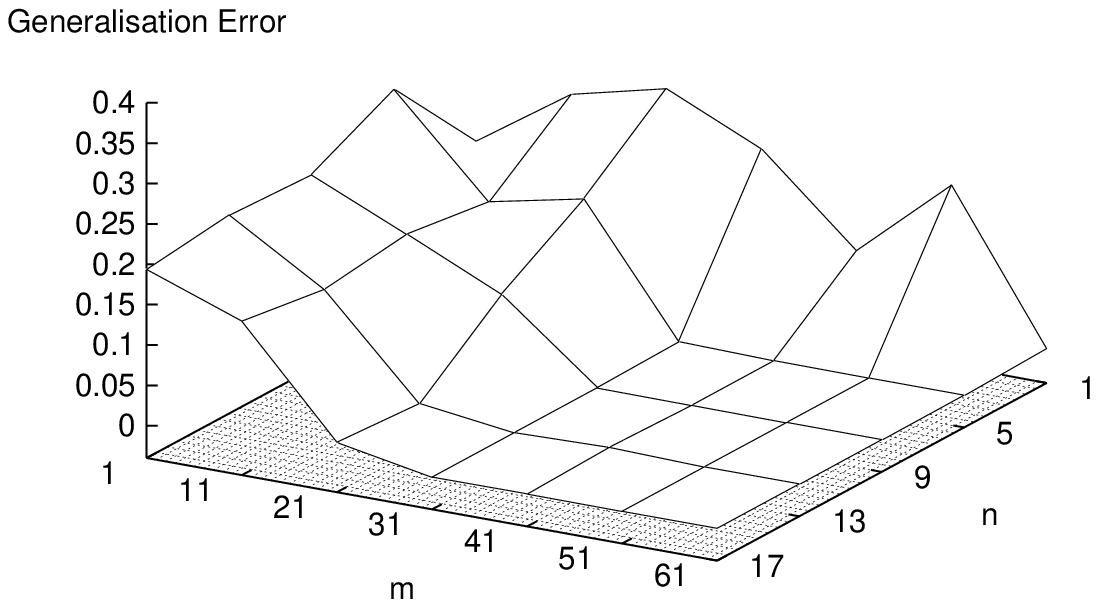}
\jcaption{Generalisation error as a function of $n$---the number of functions
learnt from the environment---and $m$---the number of examples per 
function, for three independent simulations from the second experiment.}
{symfig}
\end{center}
\end{figure}

As in the first experiment, $(n,m)$ samples resulting in networks with perfect 
generalisation had their representations tested, and it was found
that no more than five functions needed to be sampled from the environment
before a perfect representation was learnt. Plots of the
mean-squared and $L^\infty$ representation error
are given in figure
\ref{srepplot}. The remarkably low representation error has essentially the
same  explanation as in the previous experiment. The behaviour of a typical
perfect representation is shown in figure \ref{srepfig}.
Note that the
representation maps the input vectors into points in $\R^3$ in such a way
that all the functions in the environment can be implemented as linearly
separable maps from $\R^3$ into $\{0,1\}$ (look at the cube in figure
\ref{srepfig}---any subset of the four labelled vertices can be sliced off
with a plane). Given that the networks in $\G$ are only zero-hidden-layer
``perceptrons'', the representation is forced to behave in this way if it is
to be a useful one for learning all functions from the environment. 

\begin{figure}
\begin{center}
\leavevmode
\epsfbox{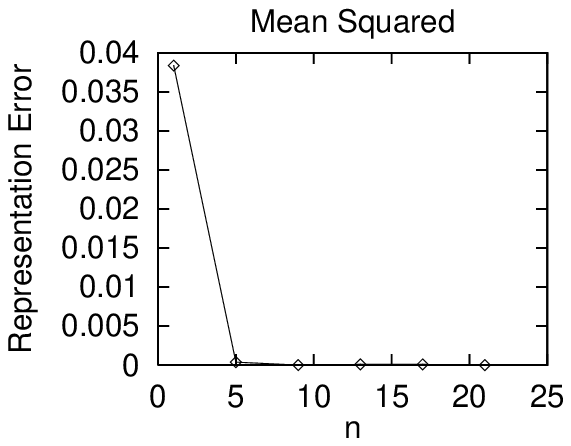}
\epsfbox{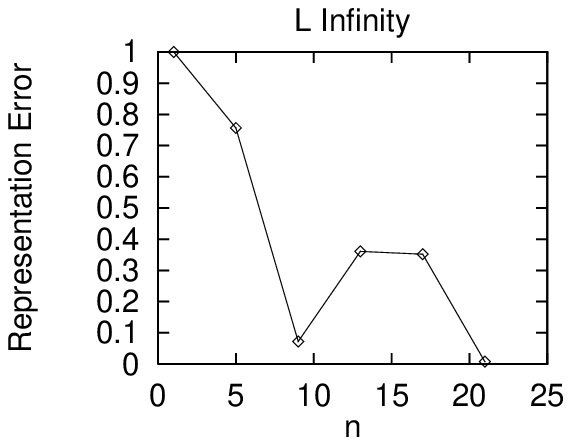}
\jcaption{Mean-squared and $L^\infty$ representation error curves for the
second experiment.}
{srepplot}
\end{center}
\end{figure}

\begin{figure}
\begin{center}
\leavevmode
\epsfbox{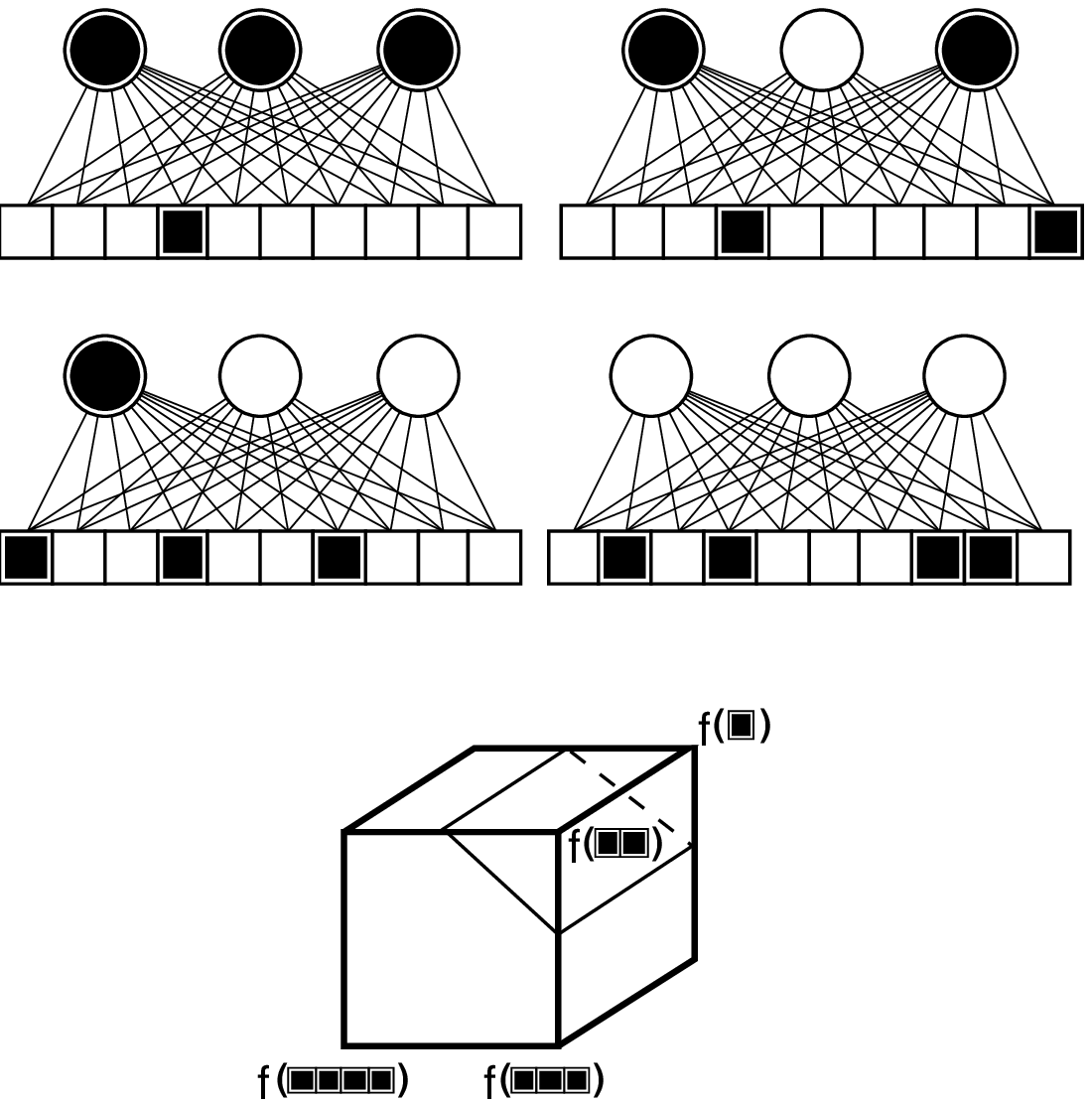}
\jcaption{Behaviour of a perfect representation $f$ for the second experiment.
The output layer of $f$ is
invariant (within about 1\%) with respect to changes in the order of its
inputs. Four example input vectors and the corresponding output vectors are
shown here. The cube illustrates how any function from the
environment can be implemented as a linearly separable map applied to the
output of $f$ (one example is shown).}
{srepfig}
\end{center}
\end{figure}

Again, the learning ability of a network using a perfect representation was
compared with a network that had to learn with the full space, and the
generalisation curves of both for four different functions from the
environment are plotted in figure \ref{srepcomp}. 
The full generalisation curves
for learning with the full space $\comp{\G}{\F}$ are also shown in figure
\ref{srepcomp1}. These curves were
generated in the same manner as the corresponding curves in experiment one.
Learning with a
representation is clearly far better than learning without. 

\begin{figure}
\begin{center}
\leavevmode
\epsfbox{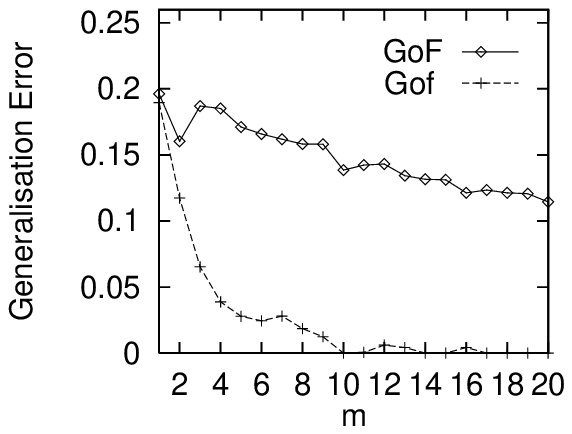}
\epsfbox{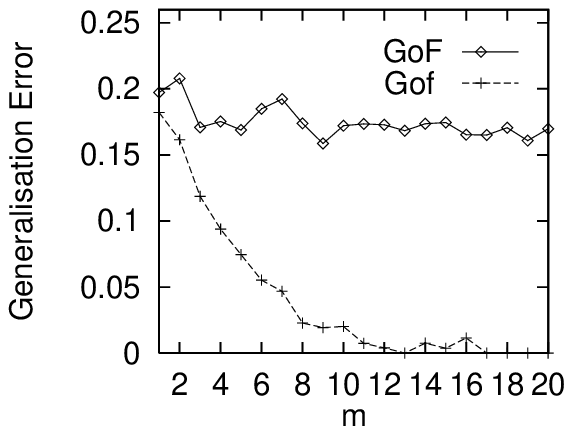}
\vspace{5mm}
\epsfbox{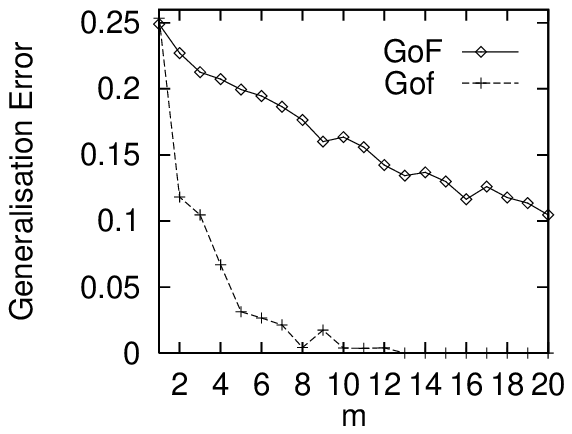}
\epsfbox{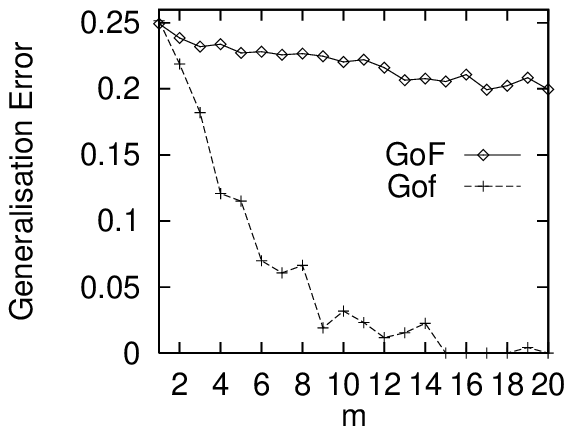}
\jcaption{Generalisation curves for learning with a representation (Gof) 
vs. learning without (GoF) for
four different functions from the environment in the second experiment.}
{srepcomp}
\end{center}
\end{figure}

\begin{figure}
\begin{center}
\leavevmode
\epsfbox{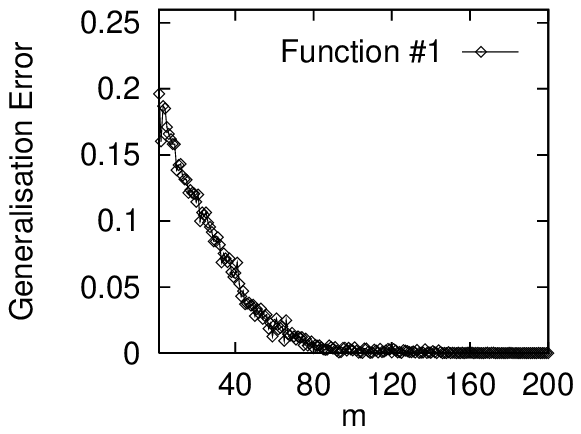}
\epsfbox{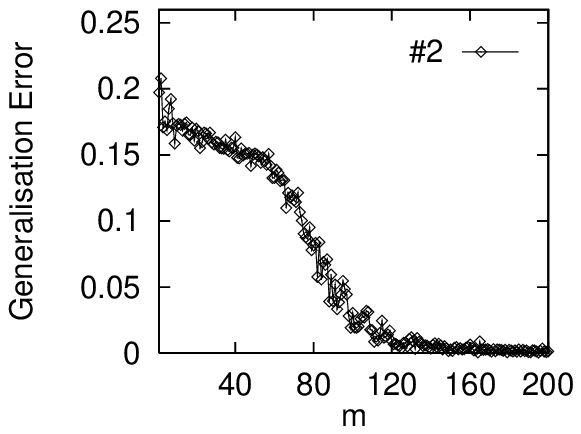}
\vspace{5mm}
\epsfbox{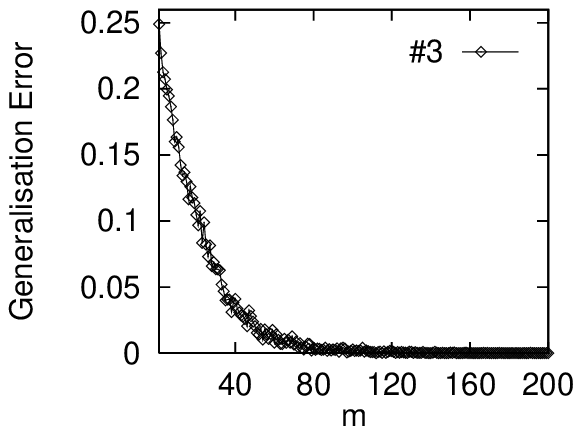}
\epsfbox{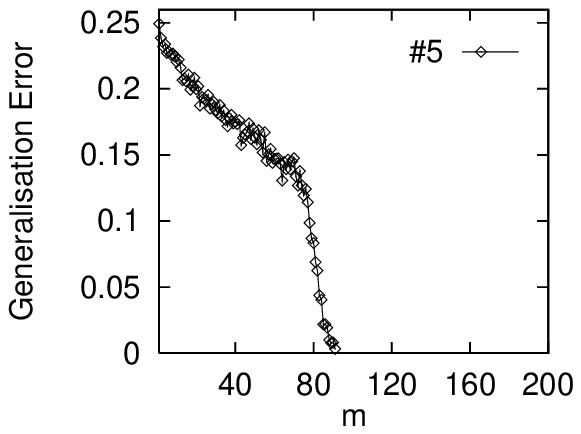}
\jcaption{Full generalisation curves for learning without a representation
for four different functions from the environment in the second experiment.} 
{srepcomp1}
\end{center}
\end{figure}

\subsection{Local Minima.}
The number of restarts needed (or equivalently the number of 
local minima encountered) in typical simulations from both
experiments are plotted as a function of $n$ and $m$ in figures \ref{tcount}
and \ref{scount}. 
Because the CM5 used for
these experiments had $32$ nodes, all simulations were automatically started
with $32$ different random initial weight vectors. The number of restarts
shown in the graphs do not include the initial $32$.

The behaviour of the first experiment is not particularly illuminating.
Essentially all that can be concluded from the graphs in figure \ref{tcount}
is that the number of local minima tended to increase monotonically 
with the number
of examples. However the behaviour of the second experiment has an interesting
interpretation. I have no explanation for why the two experiments behaved so
differently. 

Observe for the second experiment that the
number of restarts required increased fairly monotonically with $m$ until
the value of $m$ corresponding to the transition to perfect generalisation
was reached (compare figure \ref{symfig}), at which point the graph drops
rapidly back to zero. This behaviour can be understood as follows:
firstly, the ease of training for values of $m$ far below and far above 
the transition to perfect generalisation 
indicates that for these values of $m$ there are mostly global, not local,
minima on the error surface. For
values of $m$ on the upside of the transition from poor to good
generalisation, the fact that all solutions turn out to be true
solutions to the problem indicates that most of the global minima
correspond to true solutions. These points are
clearly also global minima for values of $m$ on the lower side of the
transition point. However as most solutions for small values of $m$ turn out
to have
poor generalisation, these global minima must constitute only 
a small fraction of
all the global minima on the error surface---the other ones being points with 
low error on the training set but poor generalisation ability. As $m$
increases it is these points of poor generalisation that must discontinue
being global minima of the error surface, i.e. their error values must
``lift''. Thus they become local minima for $m$ around the transition point
and eventually get ironed out of the error surface altogether at high enough
values of $m$. This explains the preponderance of local minima around the
transition point. With such a large number of deep
local minima it takes many restarts before the initial weight values (by 
chance) land in the attracting basin of a global
minima.

As a final point, part of the cause of the local minima in these
representation learning simulations is no doubt the fact that error
information for $f$ is filtered first by the output functions $g$. It will
be shown in chapter \ref{flearnchap} how this can be avoided in the case
that the environment of learner consists of classification problems. 

\begin{figure}
\begin{center}
\leavevmode
\epsfbox{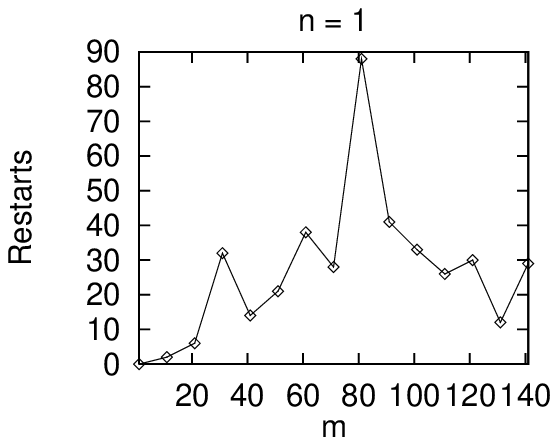}
\epsfbox{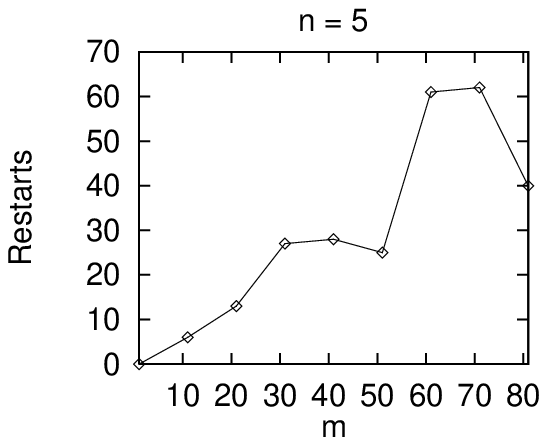}
\vspace{5mm}
\epsfbox{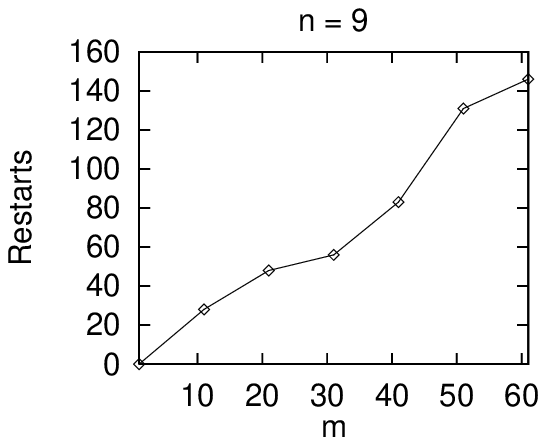}
\epsfbox{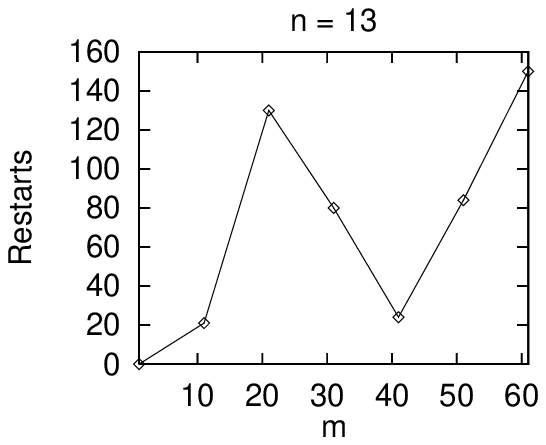}
\vspace{5mm}
\epsfbox{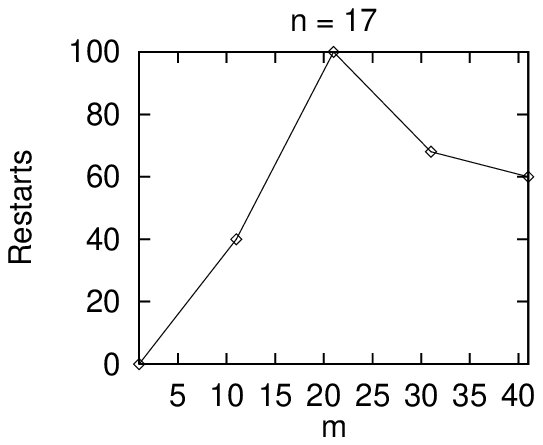}
\jcaption{Restarts needed (or local minima encountered) 
before finding a solution for a typical simulation from the first
experiment.}
{tcount}
\end{center}
\end{figure}

\begin{figure}
\begin{center}
\leavevmode
\epsfbox{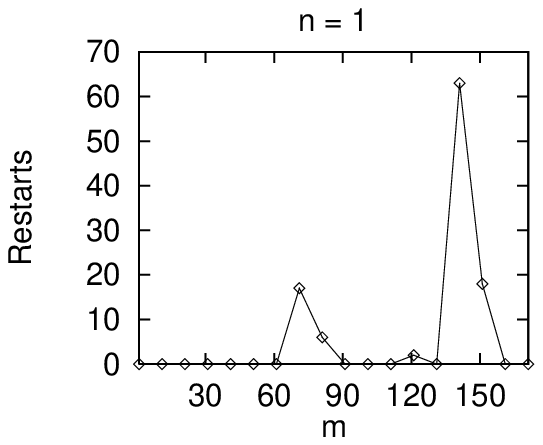}
\epsfbox{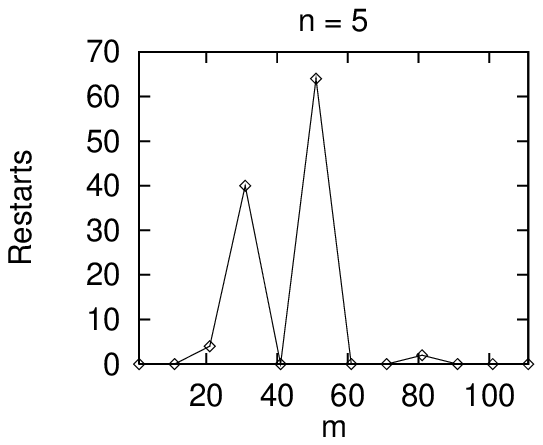}
\vspace{5mm}
\epsfbox{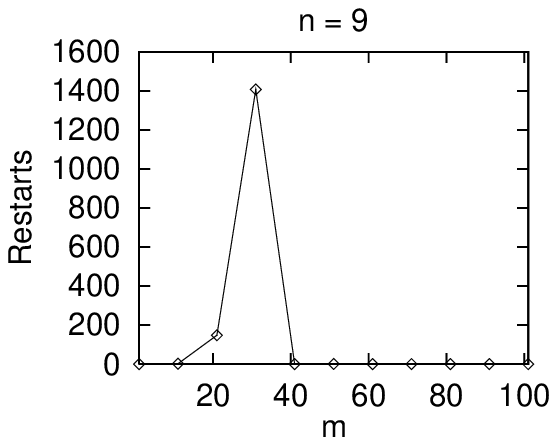}
\epsfbox{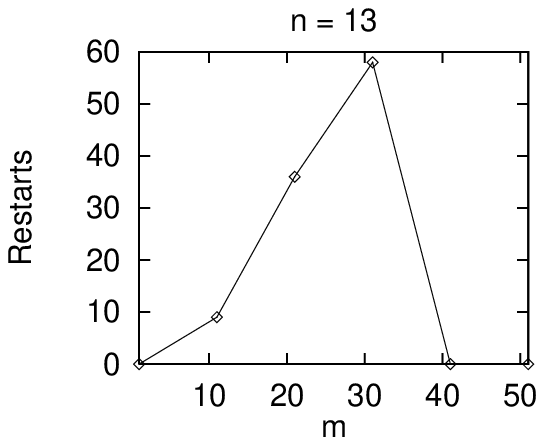}
\vspace{5mm}
\epsfbox{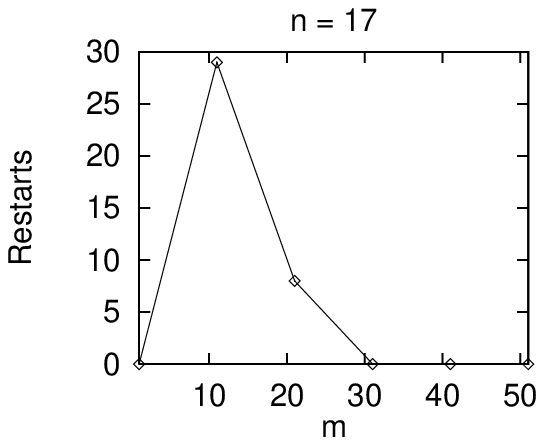}
\epsfbox{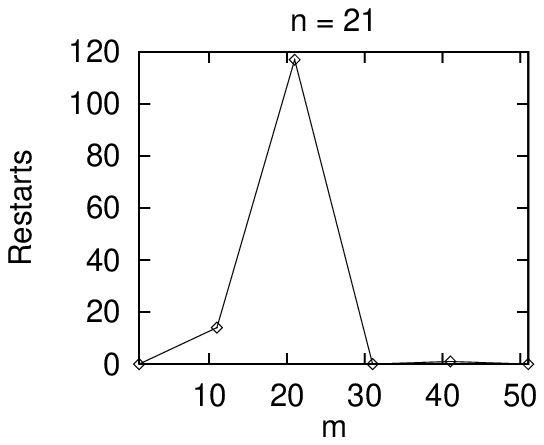}
\jcaption{Restarts needed (or local minima encountered) 
before finding a solution for a typical simulation from the second
experiment.}
{scount}
\end{center}
\end{figure}

\section{Zip-Code Network}
One of the more famous examples of a successful application of neural
networks is the network developed by Le Cun et.\ al.\ (1989)\label{L1}
for classification of
handwritten digits in zip-codes. The final hidden layer of their network
consisted of 30 nodes and these were fully connected to 10 output nodes, one
for recognising each of the digits. Within the framework of representation
learning discused here, the final hidden layer of their network can be
viewed as the output of a representation, while the 10 output nodes (and
their respective incoming weights) can be viewed as
10 output functions $g_1,\dots,g_{10}$. It would be interesting to extract the
representation part of the network and use it to learn some new characters,
such as `A', `B', etc. My guess is that the network's ability to
learn new characters will be quite limited. However
if it {\em can} successfully learn many new
characters using the existing representation then that would be
evidence {\em against} the thesis being advocated here 
that the information necessary to learn a
good representation is spread across many examples of similar learning
tasks. On the other hand, if the network is incapable of learning a
significant number of new characters with the existing representation then
it would be interesting to see if the network's {\em architecture} is
sufficient to learn a good representation by training it on many more
character classification problems than just the digits (this can be achieved
by simply adding an extra output node to the network for each new character
and retraining the entire network\jfootnote{Note that for classification
problems such as character recognition the {\em entire} network
$\comp{\gv}{\fbar}$ can be trained all at once because the value of every
output function $g_i$ is always known for each input. This means the network can
be trained with ordinary backpropagation in this case.}).
If the theory here is correct, the number of examples of each character
required should decrease as the number of characters increases. The other
two principles of representation learning (speed and effectiveness of
learning new characters using a good representation) could also be tested.

\chapter{Vector Quantization}
\label{quantchap}
In this chapter it is shown how the idea of the {\em environment} of a
learning process can be applied to the 
problem of vector quantization and how this leads to a natural choice
for the distortion measure in the quantization process. This distortion
measure turns out to be optimal in a very general sense.

\section{The Problem}

\sloppy As any real digital communication channel has only finite
capacity, transmitting continuous data (e.g speech signals or images)
requires firstly that it be transformed into a discrete
representation.  Typically, given a probability space
$(X,P,\sigma_X)$, one chooses a {\em quantization} of $X$,
$\{x_1,\dots,x_k\}\subset X$ and instead of transmitting $x\in X$, the
index of the ``nearest'' quantization point $q_d(x) = \min_{x_i}
d(x_i,x)$ is transmitted, where $d$ is some function (not necessarily
a metric) measuring the distance between points in $X$. $d$ is called
a {\em distortion measure}. The quantization points are chosen so that
the expected distance between $x$ and its quantization is minimal, i.e
$\xv = \{x_1,\dots,x_k\}$ are chosen to minimize the {\em
  reconstruction error}
\begin{equation}
\label{recerr}
\glsname{Edxv} \de \int_X d(x,q_d(x))\, dP(x).
\end{equation}
There are simple methods, for
iteratively generating good quantization sets $\{x_1,\dots,x_k\}$ (See Cover
and Thomas (1991) chapter 13\label{CT2} and references therein).

\fussy
Some examples of distortion measures are the {\em Hamming metric},
$$
d(x,x') = 1 - \delta(x,x'),
$$
where $\delta$ is the Kronecker delta function, and the {\em squared Euclidean
distortion} for vector-valued $X$,
$$
d(x,x') = \|x-x'\|^2.
$$
The use of these distortion measures has more to do with their convenient
mathematical properties than their applicability to a particular problem
domain. For example, suppose $X$ is a space of images and it is images of
characters that are being transmitted over the channel. An image of the
character ``A'' and another translated image of the same character would be
considered ``close'' in this context, although the squared Euclidean
distance between the two images would be large, quite likely larger
than the distance between an image of ``A'' and an image of ``B'' in the
same location. Thus the Euclidean distortion measure does not at all capture
the idea of two images being ``close'' in this context. Another example is
that of speech coding---there is a large squared Euclidean distance between
a speech signal and a small translation of it in time, although both sound
very similar to a human observer. Although improved distortion measures
based on entropy considerations have been found, the situation is still
quite unsatisfactory. To quote from Cover and Thomas (1991),
page 340,\label{CT3}
\begin{quote}
In image coding, however, there is at present no real alternative to using
the mean squared error as the distortion measure.
\end{quote}
The next section  shows how this problem may be solved, at least {\em
in principle}, by using the idea of the {\em environment} of a learning
process.

\section{The Solution}

What makes the translated image of an ``A'' close to the original, while an
untranslated image of a ``B'' quite distant? And what makes two speech
signals nearly identical even though they are miles apart from a Euclidean 
perspective? In both cases it is because there is an {\em environment} of
classifiers that behave as if the distinct images of ``A'' are close, and
as if the speech signals are close. 
For example, although there is nothing {\em intrinsic}
about ``A'' and its translate that makes them close, there is
something about the set of all character classifiers that makes them close.
A classifier for the character ``A'' is just a function $f_A\colon X\to
\{0,1\}$
such that if $x$ is an image of an ``A'' then $f_A(x)=1$ and if not, then
$f_A(x)=0$. Classifiers for ``B'', ``C'', ``\#'' and any other character-like
objects  similarly exist in the environment of X. All these classifiers have
the property that they are {\em constant} across images of the same
character, either ``0'' if the character is not theirs, or ``1'' if it is.
Thus it is this space of classifiers that determines what makes images of
characters close together. Similarly for speech signals. There exist word
classifiers in the environment of the space of all speech signals that are
constant across those sets of signals corresponding to a particular word.
The word classifiers induce the distance measure on the speech signals. 

In order to formalize this idea, suppose that the {\em environment} of
a probability space $(X,P,\sigma_X)$ consists of a set of functions
$\F$ mapping $X$ into a space $(Y,\sigma)$, where $\sigma\colon
Y\times Y\to \R$ ($\sigma$ may be a metric), and an {\em environmental
  probability measure} $Q$ on $\F$. An environment so defined induces
the following natural distortion measure on $X$:
\begin{equation}
\label{rho}
\glsname{rho} \de \int_\F \sigma(f(x),f(y))\, dQ(f),
\end{equation}
for all $x,y\in X$\footnote{In this chapter it will be assumed that 
everything that needs to be measurable in fact is.}. 
Note that if $\sigma$ is a metric on $Y$ then 
$\rho$ is a pseudo-metric on $X$.
In relation to the character
transmission problem, $\F$ would consist of all character-like classifiers, $Y$
would be the set $\{0,1\}$ and $\sigma$ would be the Hamming metric. Then if
$x$ and $y$ are two images of the same character, $f(x)=f(y)$ for all
$f\in\F$ and so $\rho(x,y) = 0$, as required. If $x$ and $y$ are images of
different characters, two of the functions in $\F$ would have
$f(x) \neq f(y)$ which would give $\rho(x,y) > 0$, assuming the functions
have positive probability under $Q$. Thus $\rho$ behaves exactly as desired
for a distortion measure in this context. Note that $\rho$ depends only upon
the environment $(\F,Q)$ and not upon $X$ or its probability measure $P$.
Thus problems with the same $X$ but different environments (for example
character classification and face recognition---different environments for
the space of images) will generate different
distortion measures. The following pair of simple
examples further illustrate the behaviour of $\rho$ as a
distortion measure. 

\begin{exmp}
Let $X=Y=[0,1]$, $\sigma(y,y')=|y-y'|$ for all
$y,y'\in Y$ and
$\F$ be the set of all linear functions mapping $[0,1]$ into $[0,1]$ with
slope $a$ uniformly distributed also in the range $[0,1]$. Thus
$\F=\{f\colon x\mapsto ax\colon a\in [0,1]\}$. For any $x,y\in X$,
\begin{align*}
\rho(x,y) &= \int_\F \sigma(f(x),f(y))\, dQ(f) \\
          &= \int_0^1 |ax-ay|\, da \\
	  &= \frac12|x-y|.
\end{align*}
Thus $\F$ induces the Euclidean metric on $[0,1]$ in this case. 
\end{exmp}
\begin{exmp}
\label{ex3}
Let $X=Y=[-1,1]$, $\sigma(y,y')=|y-y'|$ for all
$y,y'\in Y$ and
$\F=\{f\colon x\mapsto a x^2\}$ with $a$ uniformly distributed in the
range $[-1,1]$. This time,
\begin{align*}
\rho(x,y) &= \frac12\int_{-1}^1 |a x^2 - a y^2|\, da \\
          &= \frac12|x-y| |x+y|.
\end{align*}
$\rho$ is plotted in figure \ref{rhoplot}. Note that $\rho(x,y)=0$ if $x=y$
and if $x=-y$, so that $x$ and $-x$ are zero distance apart under $\rho$.
This reflects the fact that $f(x)=f(-x)$ for all $f\in\F$. Notice also that
$\rho(x,y)$ is the ordinary Euclidean distance between $x$ and $y$, {\em
scaled} by $\frac12|x+y|$. Thus two points with fixed Euclidean distance become
further and further apart under $\rho$ 
as they are moved away from the origin. This
reflects the fact that the quadratic functions in $\F$ have larger variation
in their range around large values of $x$ than they do around small values
of $x$. This can also be seen by calculating the $\ep$-ball around a point
$x$ under the $\rho$ metric (i.e the set of points $x'\in X$ such that
$\rho(x,x') \leq \ep$). To first order in $\frac\ep{x}$ this is 
$$
\[-x-\frac{\ep}{x}, -x + \frac{\ep}{x}\] \bigcup
\[x-\frac{\ep}{x}, x + \frac{\ep}{x}\].
$$
Note that  the Euclidean diameter of the $\ep$-ball around $x$ decreases
linearly with $x$'s---Euclidean again---distance from the origin. Although not
as illuminating as the metric for a quadratic environment, the metric for a
cubic environment is particularly attractive and so is plotted in figure
\ref{rhoplot1}.

\begin{figure}
\begin{center}
\leavevmode
\epsfbox{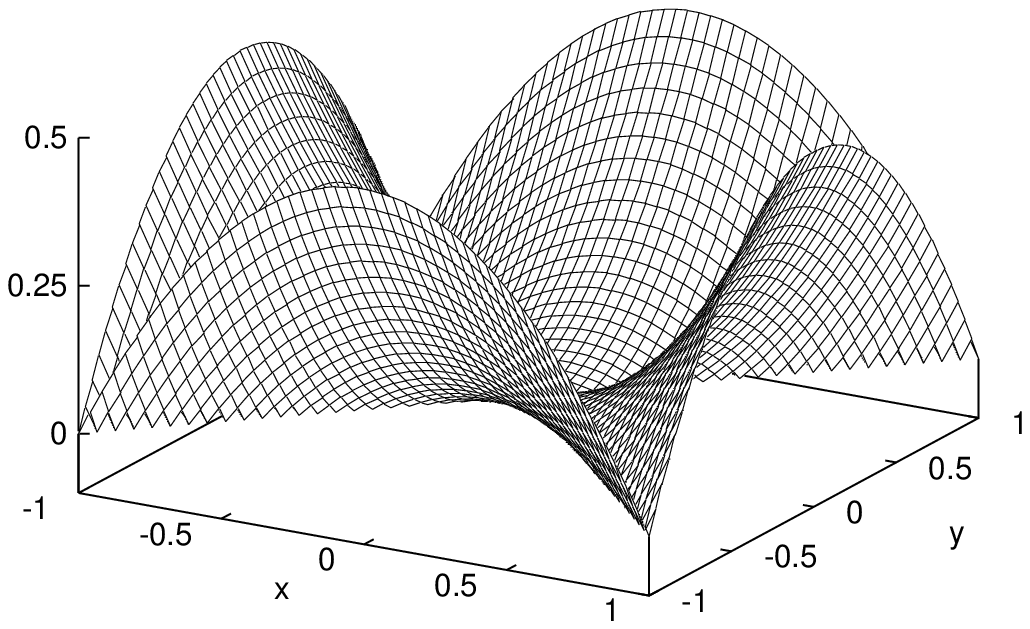}
\jcaption{$\rho(x,y)$ for a quadratic environment.}
{rhoplot}
\end{center}
\end{figure}

\begin{figure}
\begin{center}
\leavevmode
\epsfbox{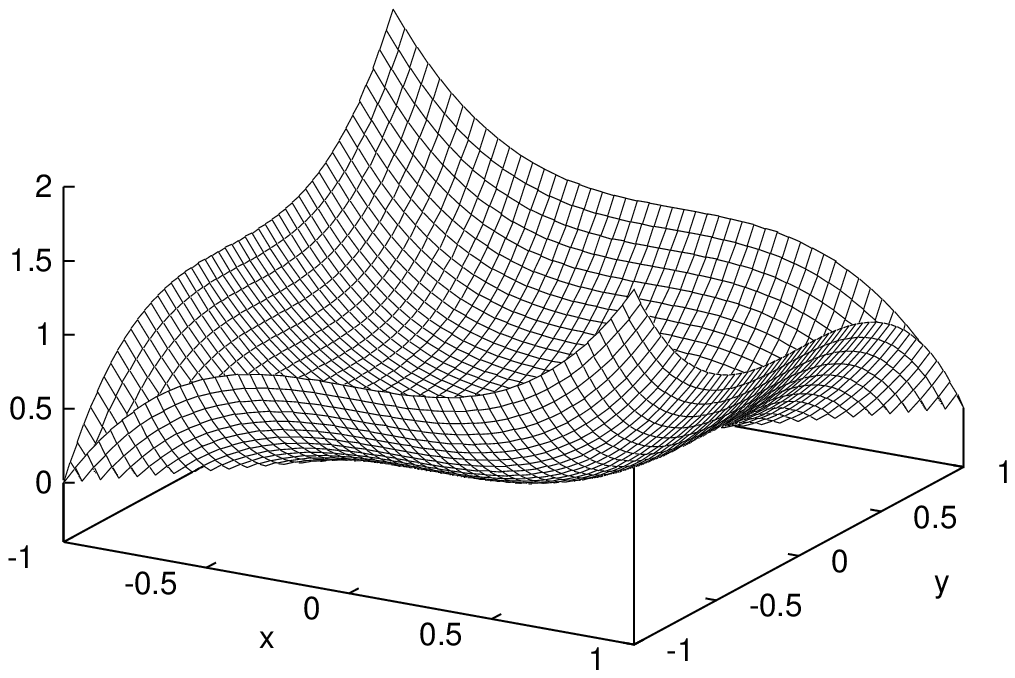}
\jcaption{$\rho(x,y)$ for a cubic environment.}
{rhoplot1}
\end{center}
\end{figure}

\end{exmp}

\section{Optimal Function Approximation}

In this section it is shown that $\rho$ is the optimal distortion measure to
use in vector quantization if the goal is to find piecewise constant
approximations to the functions in the environment. 

Piecewise constant approximations to $f\in\F$ are generated by specifying a
quantization $\xv=\{x_1,\dots,x_k\}$ of $X$ and a partition 
$\Xv=\{X_1,\dots,X_k\}$ of
$X$ that is {\em faithful} to $\{x_1,\dots,x_k\}$ in the sense that
$x_i\in X_i$ for $1\leq i\leq k$.
The piecewise constant approximation $\fhat$ to any function $f$ 
is then defined by $\fhat(x) = f(x_i)$ for all
$x\in X_i$, $1\leq i\leq k$. 
The most natural way to measure the deviation between $f$ and
$\fhat$ in this context is with the pseudo-metric $d_P$,
$$
d_P(f,\fhat) = \int_X\sigma(f(x),\fhat(x))\,dP(x).
$$
$d_P(f,\fhat)$ is the expected difference between $f(x)$ and $\fhat(x)$ 
on a sample $x$ drawn at random from $X$ according to $P$. The {\em
reconstruction error} of $\F$ with respect to the pair $\xv=\{x_1,\dots,x_k\}$ and
$\Xv=\{X_1,\dots,X_k\}$ is defined to be the expected deviation between $f$ and
its approximation $\fhat$, measured according to the distribution $Q$ on $\F$:
$$
\glsname{EFxX} \de \int_\F d_P(f,\fhat)\, dQ(f).
$$
The quantization and partition should be chosen so as to minimize $E_\F$.

Given any quantization $\xv=\{x_1,\dots, x_k\}$ and distortion measure
$\rho$ as in equation \eqref{rho}, define the partition 
$\Xv^\rho = \{X^\rho_1,\dots, X^\rho_k\}$ by $X^\rho_i = \{x\in X\colon \rho(x,x_i) 
\leq \rho(x,x_j),\,\text{for all}\, j\neq i\}$ (break any ties by rolling an appropriately
sided dice).
Call this the partition of $X$ {\em induced} by $\rho$ and
$\xv$. 

\begin{thm}
The reconstruction error $E_\F$ of $\F$ is minimal with respect to a 
quantization $\xv = \{x_1,\dots,x_k\}$ minimizing the
reconstruction error $E_\rho(\xv)$ of $X$, and the partition $\Xv^\rho$ 
induced by 
$\rho$ and $\{x_1,\dots,x_k\}$.
\end{thm}

\begin{pf}
This is proved in two steps. First it is shown that if $\xv=\{x_1,\dots, x_k\}$
is any quantization of $X$, then the partition $\Xv$ minimizing
$E_\F(\xv,\Xv)$ is the one
induced by $\rho$. The proof is completed by showing that 
$E_\F(\xv,\Xv^\rho) = E_\rho(\xv)$.

Thus, let $\xv=\{x_1,\dots,x_k\}$ be any quantization of $X$ and let
$\Xv^\rho =\{X^\rho_1,\dots,X^\rho_k\}$ be the corresponding partition induced by
$\rho$. Denote the approximation of $f\in\F$ with respect to this partition
by $\fhat_\rho$. Let $\Xv=\{X_1,\dots,X_k\}$ be any other partition of $X$ that is
faithful to $\xv$ and let $\fhat$ denote the approximation of
$f$ with respect to this second partition. Define $X_{ij} = X_i \cap
X^\rho_j, 1\leq i \leq k, 1\leq j\leq k$. Note that the $X_{ij}$'s are also
a partition of $X$. The reconstruction error of $\F$ with respect to the
partition $\Xv$ satisfies:
\begin{align*}
E_\F(\xv,\Xv) &= \int_\F d_P(f,\fhat)\, dQ(f) \\
     &= \int_\F\int_X \sigma(f(x),\fhat(x))\,dP(x)\,dQ(f) \\
     &= \sum_{i,j=1}^k\int_{X_{ij}}\int_\F\sigma(f(x),f(x_i))\,dQ(f)\,dP(x) \\
     &= \sum_{i,j=1}^k\int_{X_{ij}}\rho(x,x_i)\,dP(x) \\
     &\geq \sum_{i,j=1}^k\int_{X_{ij}}\rho(x,x_j)\,dP(x) \\
     &= \int_\F d_P(f,\fhat_\rho)\, dQ(f) \\
     &= E_\F(\xv,\Xv^\rho),
\end{align*}
and so the first part of the proof is established.
Now observe that
\begin{align*}
E_\F(\xv,\Xv^\rho) &= \int_\F d_P(f,\fhat_\rho)\,dQ(f) \\
     &= \int_\F\int_X \sigma(f(x),\fhat_\rho(x))\,dP(x)\,dQ(f) \\
     &= \int_X\int_\F \sigma(f(x),f(q_\rho(x)))\,dQ(f)\,dP(x) \\
     &= \int_X \rho(x,q_\rho(x))\, dP(x) \\
     &= E_\rho(\xv),
\end{align*}
which completes the proof.
\end{pf}

\begin{exmp}
For the quadratic environment of example \ref{ex3}, the
optimal quantization for $k=6$ is shown in figure \ref{fhat}
along with $f$ and $\fhat_\rho$ for $f(x)=x^2$.
Note how the optimal quantization reduces the deviation between $f$ and its
approximation by spacing its points closer together for larger values of $x$.

Even in such a simple environment as this one the optimal quantization is
not so easy to calculate. First note by the symmetry of the
environment that the quantization points $\{x_1,\dots, x_k\}$ can all be
assumed to be positive, and without loss of generality suppose that $x_1\leq
x_2 \leq\dots\leq x_k$. The contribution of any triplet 
$x_{i-1}, x_i, x_{i+1}$ to the quantization error $E_\rho(\xv)$ is
\begin{align*}
\int\limits_{x_{i-1}}^{\sqrt{\frac12 \(x_{i-1}^2 + x_i^2\)}} x^2 - x_{i-1}^2\,
dx\,\, &+
\int\limits_{\sqrt{\frac12 \(x_{i-1}^2 + x_i^2\)}}^{x_i} x_i^2 - x^2\, dx\,\,
+ \int\limits_{x_i}^{\sqrt{\frac12 \(x_i^2 + x_{i+1}^2\)}} x^2 - x_i^2\, dx\\
&+ \int\limits^{x_{i+1}}_{\sqrt{ \frac12\(x_i^2 + x_{i+1}^2\)}}
x_{i+1}^2 - x^2\, dx.
\end{align*}
Minimizing this expression with respect to $x_i$ gives,
$$
x_i^2 = \frac14\(x_{i-1}^2 + x_{i+1}^2\) + \frac1{4\sqrt{2}}\sqrt{x_{i-1}^4 + 
6 x_{i-1}^2 x_{i+1}^2 + x_{i+1}^4}.
$$
A similar procedure can be used to show that $x_1$ and $x_k$ must satisfy,
\begin{align*}
x_1 &= \frac{x_2}{\sqrt{7}},\\
x_k &= \frac{4+\sqrt{2+7 x_{k-1}^2}}{7}.
\end{align*}
Optimal quantization points can be found numerically by
initialising $x_i = i/k$ for $1\leq i\leq k$ and then iteratively updating
the points in order according to these relations. This procedure 
is guaranteed to converge to the optimal solution or a limit cycle, although
none of the latter were observed in any of the tests, probably because they
do not exist in this case or have starting conditions of measure zero.
An error of 1 part in a million was achieved in 36 iterations
for the six points in figure \ref{fhat}.
\end{exmp}
\begin{figure}
\begin{center}
\leavevmode
\epsfbox{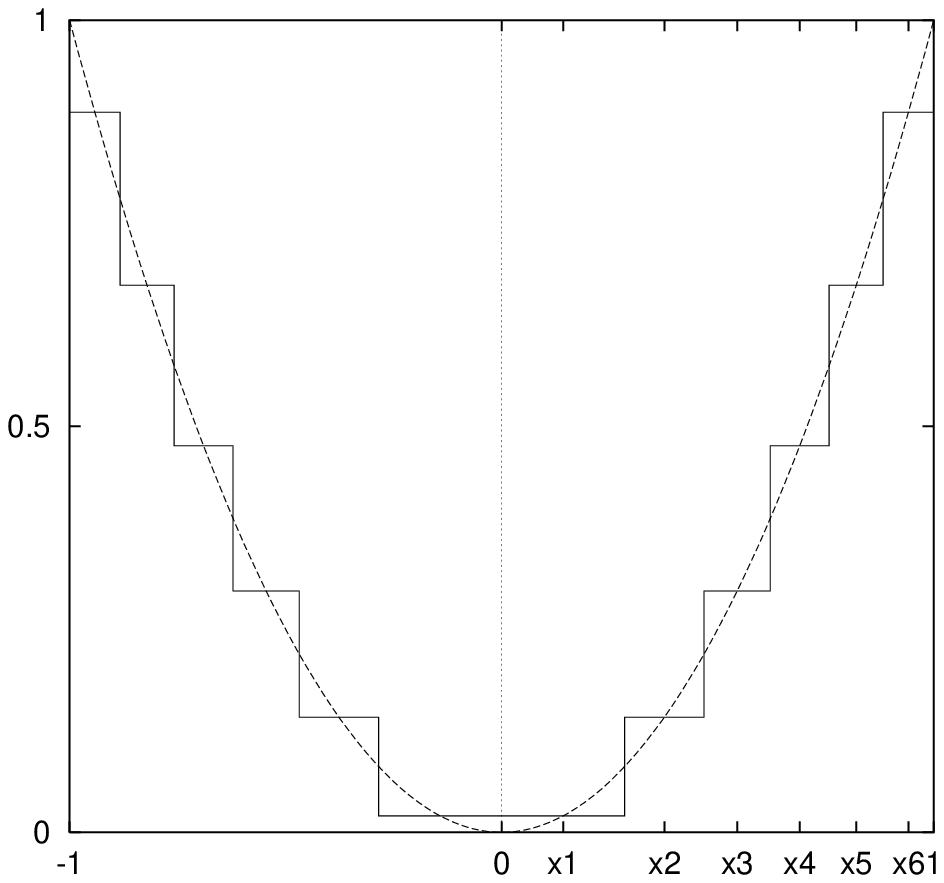}
\jcaption{Optimal quantization points and the corresponding approximation to
$f(x)=x^2$ for the quadratic environment of example \ref{ex3}. The six
quantization points are (to three significant figures) 
$x_1=0.142$, $x_2=0.377$, $x_3=0.545$, $x_4=0.690$, $x_5=0.821$, $x_6=0.942$.}
{fhat}
\end{center}
\end{figure}

\section{Learning $\rho$}

For most environments encountered in practise (e.g speech recognition or
image recognition), $\rho$ will be unknown. However it may be estimated or
{\em learnt} in the same manner that a representation is learnt---by
sampling from the environment. One method
is to sample $M$ times from $\F$ according to $Q$ to generate
$\{f_1,\dots,f_M\}$ and then $N$ times from
$X$ according to $P$ to generate $\{x_1,\dots,x_N\}$. For any pair $x_a,
x_b$ an estimate of $\rho(x_a,x_b)$ is given by 
$$
\rhat(x_a, x_b) = \frac1M\sum_{i=1}^M \sigma(f_i(x_a), f_i(x_b)).
$$
This generates
$\frac{N\times(N-1)}2$ training {\em triples},
$$\{(x_1,x_2,\rhat(x_1,x_2)), \dots, (x_{N-1}, x_N, \rhat(x_{N-1}, x_N))\},$$
which can be used as data to train, for example, a neural network. That is,
the neural network would have two sets of inputs, one set for $x_a$ and one
set for $x_b$, and a
real-valued output representing the network's estimate of
$\rhat(x_a,x_b)$. The network would be trained to approximate $\rhat$. 
Although the theory is not developed here, a similar analysis to
that given for representation learning could be applied in this situation to
generate estimates for the size of $M$ required to ensure that $\rhat$ is
close to $\rho$ on the training set, and also bounds on the size of $N$
ensuring good generalisation from the resulting neural network.

Another method for learning $\rho$ is to use the result of a previous
$(n,m)$ learning process. That is, a hypothesis $\comp{\gv}{\fbar}$
estimates $\rho$ by
$$
\rhat(x,y) = \frac1n \sum_{i=1}^n
\sigma\(\comp{g_i}{f}(x),\comp{g_i}{f}(y)\).
$$
Note that this method would work using the output of any $(n,m)$ learning
process, not just a representation learner. 

\section{Discussion}

It has been shown that the existence of an {\em environment} for a
quantization process leads to a {\em canonical} distortion measure which
turns out to be optimal if approximation of the functions in the environment
is the aim. A couple of brief ideas were mentioned for learning the
distortion measure by sampling from the environment.

If the environment consists of classification problems then it is fairly
easy to see (and is demonstrated in the next chapter) that learning the
distortion measure is equivalent to learning a representation for the
environment that effectively solves all the classification problems. Thus,
not surprisingly, finding a good distortion measure for
environments such as speech recognition and image recognition is formally
equivalent to solving the learning problem for those environments. This may
not seem like  much of a step forwards, as one unsolved problem has simply
been converted into another unsolved problem (although the claim in this
thesis is that representation learning should go part way towards solving
the latter problem), however from a purely
theoretical perspective it has its advantages because it offers a broader
framework in which to tackle both problems, and also allows the derivation
of the optimality property. In the next chapter it is shown how a
marriage of the ideas of the current chapter with the representation
learning ideas of earlier chapters leads to a much improved method for
representation learning, and offers some hope for solving the problem of
architecture selection in neural networks.  

\chapter{Learning a Representation Directly}
\label{flearnchap}
If the environment of the learner is of the form $X \xrightarrow{f} V \xrightarrow{\G} Y$ and
the learner knows the form of $V$, $\G$ and the environmental measure 
$Q$, it is possible to directly
learn a representation for the environment without learning the functions
$g\in\G$ as well. This is to be contrasted with the procedure used in
Chapter \ref{expchap} in which the output functions $g$ had to be learnt
along with the representation. 
In this chapter it is shown how to do this for several 
different function spaces $\G$, and an experiment is presented
demonstrating the effectiveness of the technique for a simple environment.

\section{An error measure for the representation.}
Let the environment of the learner be of the form $X \xrightarrow{f} V \xrightarrow{\G} Y$,
let $\sigma\colon Y\times Y\to [0,M]$ be a loss function on $Y\times Y$ and $P$
be a probability measure on $X$. 
Let $Q$ be the environmental probability measure on $\G$.
As the representation $f$ is fixed, the distortion measure $\rho$ \eqref{rho}
from the previous chapter can be written
$$
\rho(x,y) = \int_\G \sigma\bigl(g(f(x)), g(f(y))\bigr)\,dQ(g),
$$
for all $x,y\in X$. Define the distortion measure $\rho_\G$ on $V$ by
$$
\glsname{rhoG} = \int_\G \sigma(g(v), g(w))\, dQ(g),
$$
for all $v,w\in V$. Note that $\rho_\G(f(x), f(y)) = \rho(x,y)$ for all
$x,y\in X$. 

Suppose the learner knows what the spaces $\G$ and $V$ are, and the measure
$Q$, and is itself
trying to learn a representation mapping $X$ into $V$ that is appropriate
for the environment. Denote the learner's estimate for the representation by
$\fhat$.  
To generate data for the learner, suppose 
$X$ is sampled $N$ times according to $P$ 
to generate $\(x_1, \dots, x_N\)$ and $\G$ is sampled $M$ times according to
$Q$ to generate $\(\comp{g_1}{f},\dots,\comp{g_M}{f}\)$. The learner is given
the $M\times N$ pairs $\(x_i, \comp{g_j}{f}(x_i)\)$ as a training set. Note
that this is not an $(M,N)$ sample. From
this training set the learner can estimate the distance between any pair
$x_i$ and $x_j$ under $\rho$  by
$$
\rhat(x_i,x_j) = \frac1M\sum_{k=1}^M 
\sigma(\comp{g_k}{f}(x_i),\comp{g_k}{f}(x_j)).
$$
For sufficiently large $M$, $\rhat(x_i,x_j)$ will with high probability be a good
estimate of $\rho(x_i,x_j)$\jfootnote{Although I have not done this, the techniques
of chapter \ref{repchap} and appendix \ref{fundapp} 
could no doubt be extended to
provide bounds on $M$ ensuring good agreement bewteen $\rhat$ and $\rho$.}. 
Given that the learner knows the function space $\G$ and the environmental
measure $Q$, it can, {\em in principle}, calculate $\rho_\G(v,w)$ for any
pair $v,w\in V$. Thus it can calculate $\rho_\G(\fhat(x_i), \fhat(x_j))$ for
any pair $x_i$ and $x_j$ in its training set and compare this quantity with
$\rhat(x_i,x_j)$. By the relation, $\rho(x,y) = \rho_\G(f(x),f(y))$, if
$\fhat$ is close to the true representation $f$, and if $\rhat$ is a
good estimate of $\rho$, then $\rhat(x_i,x_j)$ should be close to
$\rho_\G(\fhat(x_i), \fhat(x_j))$. This suggests that the following error
measure is appropriate for the learner's representation $\fhat$:
\begin{equation}
\label{errmeas}
E(\fhat) = \sum_{i,j = 1}^N\left[\rho_\G(\fhat(x_i),\fhat(x_j)) -
\rhat(x_i,x_j)\right]^2.
\end{equation}
An error of close to zero indicates that $\fhat$ maps all the training
examples $x_i$ into $V$ in such a way that their mutual distances are the
same as the mutual distances between the $f(x_i)$, where $f$ is the true
representation. 
For differentiable $\rho_\G$, $E(\fhat)$ can be minimized using gradient
descent.

\section{Two Example Calculations of $\rho_\G$.}
\begin{exmp}
\label{ex1}
Suppose that $V=\R^n$ and $\G$ consists of all linear maps from $V$ into
$\R$. $\G$ is the vector space dual of $V$ and so is itself isomorphic to
$\R^n$. With this in mind, take the measure $Q$ on $\G$ to be Lebesgue
measure on $\R^n$, but restrict $Q$'s support to the cube
$[-\alpha,\alpha]^n$. Let $\sigma(x,y) = (x-y)^2$ for all $x,y\in\R$.
$\rho_\G$ can then be reduced as follows:
\begin{align*}
\rho_\G(\vv,\wv) &= \int_\G\sigma(g(\vv),g(\wv))\, dQ(g)\\
&= \frac1{(2\alpha)^n}
\int\limits_{\av\in [-\alpha,\alpha]^n} \(\av\cdot\vv - \av\cdot\wv\)^2\,
d\av \\
&= \frac1{(2\alpha)^n}
\int_{-\alpha}^\alpha\dots\int_{-\alpha}^\alpha \(\sum_{i=1}^n a_i v_i -
\sum_{i=1}^n a_i w_i\)^2\, da_1\dots da_n \\
& = \frac{\alpha^2}{3}\|\vv-\wv\|^2.
\end{align*}
Thus $\G$ and $Q$ induce the squared Euclidean distance on $V$ in this case.
\end{exmp}
\begin{exmp}
\label{ex2}
Take the same example as above but this time threshold the output of each
$g\in\G$ with the Heaviside step function, and take $Q$ to have support 
only on the unit ball in $\R^n$ (rather than the cube, $[-\alpha,\alpha]^n$).
This gives,
$$
\rho_\G(\vv, \wv) = \frac{\theta}{\pi},
$$
where $\theta$ is the angle between $\vec{v}$ and $\vec{w}$.
\end{exmp}

\section{Representation Learning for Classification Problems}
An environment consisting only of classification problems generates a very
simple metric on the input space $X$. Any classifier environment is
equivalent to a set of Boolean functions $\H\colon X\to \{0,1\}$ with the
property that if $h(x) = 1$ for some $x\in X$ and $h\in \H$, then $h'(x) = 0$
for all $h'\in\H, h'\neq h$. For example, in character recognition---a
classifier environment---each image $x\in X$ is an example of only one
character (it cannot be an image of both an ``A'' and a ``Z'', for example),
or possibly an image of no character, so that at most one of the
classifiers $h\in\H$ has $h(x) = 1$. A classifier environment $\H$ induces a
partition of the input space $X$ by
$$
X=X_{\text{junk}}\cup\bigcup_{h\in\H} X_h,
$$
where $X_h = \{x\in X\colon h(x) = 1\}$ and $X_{\text{junk}} = \{x\in X\colon h(x) = 0
\,\text{for all}\, h\in \H\}$.
If the environmental measure $Q$ is
assumed to be uniform over $\H$\jfootnote{In practice the environmental
measure $Q$ will not be uniform, for example not every character has equal
likelihood of being observed, however deviations from uniformity are usually 
small and hence a representation that is learnt under the assumption of
uniformity would be quite likely to perform well in a non-uniform
environment.} then the metric $\rho$ on $X$ takes only three values:
$\rho(x,x') = 0$ if $x,x'\in X_{\text{junk}}$ or if there exists 
$h\in\H$ such that $x,x'\in X_h$, $\rho(x,x') = 1/|\H|$ if one and only one
of $x,x'$ is in $X_{junk}$ and $\rho(x,x') = 2/|\H|$ otherwise. If $\H$ has
infinite cardinality then $\rho$ will be indentically zero, however in any
practical situation $\H$ will be effectively finite.

Suppose for the moment that the distribution $P$ on $X$ is such that
$X_{\text{junk}}$ is a set of measure zero, in which case $\rho$ is
effectively only two-valued.
If the output functions $\G$ are assumed to be linear
or thresholded linear then the formulae derived in the previous two examples
for $\rho_\G$ can be substituted into the expression for the error measure
\eqref{errmeas}, along with either $\rhat(x_i,x_j) = 0$ or $\rhat(x_i,x_j) =
2/|\H|$, depending on whether $x_i$ and $x_j$ have the same classification or
not\jfootnote{In this case, because it is known that $\H$ is a set of
classifiers and $Q$ is uniform, $\rhat$ is known to be equal to $\rho$.}.
A representation $\fhat$ minimizing the resulting expression can then be
found by gradient descent. An $\fhat$ with zero error would
map each subset $X_h$ to its own vector $v_h\in V$ such that
$\rho_\G(v_h,v_h') = c$ if $h\neq h'$.
Thus if $\G$ is as in example \ref{ex1} or \ref{ex2},
a perfect representation will map all inputs
$x\in X$ to one of $|\H|$ vectors
in $V$ that are all either mutually
separated by the same distance, or are the same angle apart. This places a
lower bound of $|\H|-1$ 
on the dimensionality of $V$ because it must be possible to embed a
tetrahedron with $|\H|$ vertices within $V$.
If the dimensionality of $V$
is just large enough then $X$ (or at least the support of $P$)
will be mapped onto the corners of a tetrahedron (if $\G$ is linear) or
vectors pointing towards the corners of a tetrahedron (if $\G$ is
thresholded linear). Note that if $X_{\text{junk}}$ does have posisitve 
probability under $P$ then it can be mapped to the centroid of the
tetrahedron and although this does not quite give $\rho(x,x') = 1/|\H|$ for
$x\in X_{\text{junk}}$ and $x'\in X_h$ for some $h\in\H$, it does give the
correct {\em qualitative} behaviour in that $\rho(x,x')$ is at least {\em
constant} in this case.

It may be that the exact space $\G$ used by the environment is not known to
the learner. If so, the learner can still proceed in a somewhat less
rigorous manner by selecting its own representation space $\G'$, guessing a
form for the metric $\rho_{\G'}$ and then plugging this into formula 
\eqref{errmeas}. This approach can be used to avoid the quite rigid
requirements that are imposed on the behaviour of a representation by the
$\rho_\G$'s in examples \ref{ex1} and \ref{ex2}, as follows.
Rather than forcing the representation $\fhat$ to map the inputs exactly towards (or
onto) the corners of a tetrahedron, it is sufficient
if the representation simply maps all $x$ with the same
classification into the same {\em region} in $V$, in such a way that the
distance between $\fhat(x)$ and the {\em centroid} of its region is smaller than
the distance between $\fhat(x)$ and the centroid of any other region. The
classification of any novel $x\in X$ can then be determined by computing
$\fhat(x)$ and finding the region with the closest centroid to $\fhat(x)$.
Learning a new classification category (i.e. a new task) using the same
representation would simply be a matter of collecting enough examples of the
new task to accurately estimate the location of its centroid.
One way of achieving this is to take
$\rhat(x,x') = 0$ 
if $x$ and $x'$ have the same classification (i.e. $x,x'\in X_h$ for some
$h\in\H$) , $\rhat(x,x') = 1$ otherwise, and take  
$$
\rho_\G(\fhat(x),\fhat(x')) = 1 - \exp\(-\frac{\|\fhat(x) -
\fhat(x')\|^2}{T}\).
$$
The error \eqref{errmeas} on a sample ${x_1,\dots,x_N}$ then becomes,
\begin{equation}
\label{classyerr}
E(\fhat) = 
\sum_{i,j = 1}^N\left[1 - \exp\(-\frac{\|\fhat(x_i) - \fhat(x_j)\|^2}{T}\) -
                      \rhat(x_i,x_j)\right]^2.
\end{equation}
$E(\fhat)$ will be minimal for a representation that draws inputs $x$ with the
same classification as close together as possible, and separates those with
different classifications as far apart as possible. 
The exponential causes $\fhat(x)$ and $\fhat(x')$ that are more than
a small Euclidean
distance apart to be viewed, under $\rho_\G$, as having nearly unit
separation. The parameter $T$
controls the tightness of the clusters in $V$, and their relative
separation. A small value for $T$ encourages representations to form very
tight clusters that are not necessarily separated very much, while a large
$T$ allows clusters to be looser, but forces them to be further apart.
If necessary, rather than making an arbitrary choice for the value of $T$,
it could be optimized along with the parameters of the representation. 

\section{Experiment}

The error measure \eqref{classyerr} was used to learn a representation for the
translation-invariant environment of section \ref{exp1}. The objects of
figure \ref{objects} were the classification tasks. The training data
$x_1,\dots,x_N$ was generated by selecting one of the four objects uniformly
at random and placing it uniformly at random within the ten-pixel input
space (wrapping at the edges was again permitted). The representations $f\in
\F$ had no hidden layers, one output node and no squashing function,
so that $f$ was in fact a linear map. Four
values of the parameter $T$ were tried: $1,0.1, 0.01$, and
$0.001$ with $0.01$ giving the best results in that the fewest local minima
were encountered with this value. $T=0.01$ was the
value used in the results reported below. $E(\fhat)$ was minimized using
conjugate gradient descent with stopping criteria of $E(\fhat)/N < 10^{-7}$
or an $L^\infty$ error\jfootnote{That is, the maximium absolute difference
between $1 - \exp\(-\frac{\|\fhat(x_i) - \fhat(x_j)\|^2}{T}\)$ and 
$\rhat(x_i,x_j)$ for any distinct pair $x_i,x_j$.} of less than $10^{-3}$. 
If neither of these criteria were achieved and there was a less than
$0.01\%$ improvement in the error over five iterations the network was
deemed to have struck a local minima and was restarted with a new randomly
generated set of weights. The weights were generated uniformly in the range
$[0,0.1]$. 

A succesful representation $\fhat$ was tested by first 
computing the four centroids $c_i = \frac1{10}\sum_{j=1}^{10}\fhat(x_{ij})$, 
$1\leq i \leq 4$,
where $x_{ij}$ is the input vector consisting of the $i$th object placed 
in the $j$th input position, and then counting up the number of inputs
$x_{ij}$ for which $\fhat(x_{ij})$ was not minimal distance from its
corresponding centroid $c_i$. The average of the four within group variances,
$$
\text{var}_i = 
\frac1{10}\sqrt{\sum_{j=1}^{10} \sum_{k=1}^{10}\left[\fhat(x_{ij}) -
\fhat(x_{ik})\right]^2}
$$ 
was also computed. $200$ independent simulations were
performed with training sets consisting of from $2$ to $40$
examples\jfootnote{The simulations took about two hours to run on a 66Mhz,
486DX notebook computer.}. The
average of the error and variance results across all simulations is plotted
in figure \ref{fres}.
A second set of simulations was performed with a thirty-pixel input space and
with $10$ objects consisting of from $1$ to $10$ active adjacent pixels. The
results are plotted in figure \ref{fres1}. Due to the larger problem size
the curves are the average of only $24$ separate simulations in this case.

\begin{figure}
\begin{center}
\leavevmode
\epsfbox{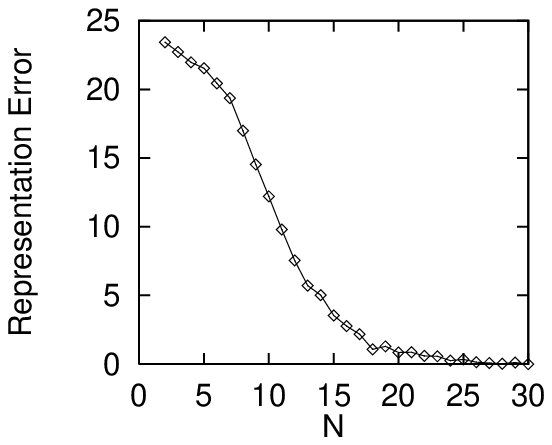}
\epsfbox{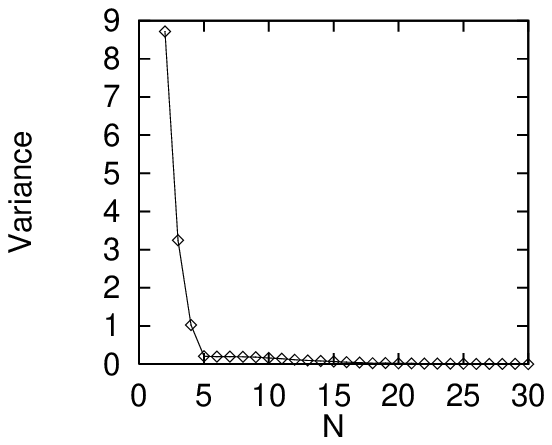}
\jcaption{Representation error and variance: experiment 1.}
{fres}
\end{center}
\end{figure}

\begin{figure}
\begin{center}
\leavevmode
\epsfbox{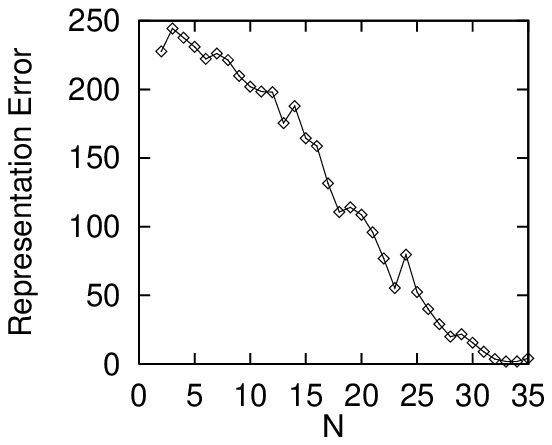}
\epsfbox{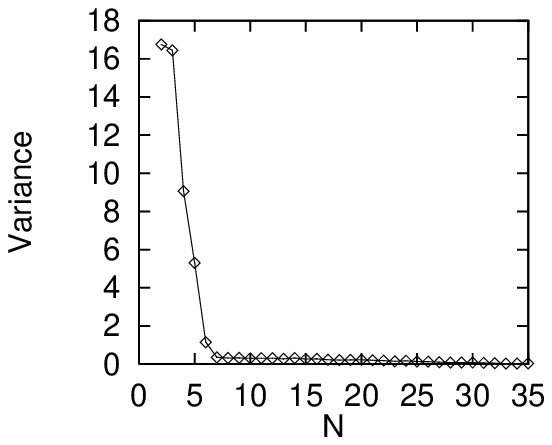}
\jcaption{Representation error and variance: experiment 2.}
{fres1}
\end{center}
\end{figure}
  
Note the very quick convergence to zero variance in both experiments.
This is significant because it indicates that
learning with a succsessful representation
$\fhat$ would require only one example of an object 
to determine its centroid accurately, and hence only one example of the
object to learn to recognise it perfectly. It is somewhat surprising that
the representation error does not fall to zero as quickly as the variance.
This is probably due to the fact that the small value of $T$ caused clusters
to be quite close together and hence required very small variance to ensure no
misclassification. This needs to be investigated further.

This method also greatly reduced
the number of restarts required---the average number in the first experiment
was $0.3$ and in the second $1.4$
(compare this with the graphs in figure \ref{tcount}).

\section{Discussion and Future Directions}
At least for the toy problem examined here, the experimental results 
show that \eqref{classyerr} is a good error measure to use for
representation learning in classification problems. Although this particular
form for the error measure may not be the best one to use in all problems,
the general idea of training the representation to {\em match a metric} on its
output space should be one that is widely applicable in practice.
Intuitively, matching a metric, particularly in the ``soft'' manner allowed
by \eqref{classyerr}, should be ``easier'' than explicitly
demanding a particular output from a given input, as happens with ordinary
learning and with the representation learning techniques of chapter
\ref{expchap}. 

Metric error measures also hold promise as a means of defining an error
measure on {\em all} layers of a representation directly, 
not just the final one as in \eqref{classyerr}. Although I do not yet have
any explicit method for this, the general idea is to note that if an
arbitrary metric $\rho$ is selected for the input space $X$, say the
Euclidean metric, and the metric $\rho_\G$ 
is known for the output space $V$, then the successive layers
of a representation $f$ can be thought of as mapping $X$ through a series 
of intermediary spaces $V_1,V_2,\dots$ in such a way that at each stage it
becomes ``easier'' for the remaining layers to produce behaviour at their
output nodes that matches $\rho_\G$. If the idea of an early layer mapping
that makes things ``easier'' for later layers could be quantified, it
could be used as an error measure to 
recursively train the first hidden layer, then the second
and so on. To finish on an even more speculative note, if there
was some way of measuring the distance between $\rho$ and
$\rho_\G$\jfootnote{A metric on metrics!} then it might be possible to
make an estimate of how many successive ``neural'' layers of processing
would be required to convert $\rho$ into $\rho_\G$ and possibly even the
number of nodes needed in each layer.

This is by far the most speculative and unfinished part of this thesis.
However, I believe
the general framework presented here is a most promising one for further
investigation.

\appendix
\chapter{Measurability and Permissibility}
\label{permissapp}
The definitions and results ensuring measurability of all the necessary
functions and sets in chapters \ref{ordchap} and \ref{repchap} 
and appendix \ref{fundapp} are presented in this appendix.

\begin{defn}
\label{measure}
Given any set of functions $\H\colon Z\to [0,M]$, where $Z$ is any
set, let \glsname{sigma_H} be the $\sigma$-algebra on $Z$ generated by
all inverse images under functions in $\H$ of all open balls (under
the usual topology on $\R$) in $[0,M]$.  Let \glsname{P_H} denote the
set of all probability measures on $\sigma_\H$.
\end{defn}
This definition ensures measurability of any function $h\in\H$ with respect
to any measure $P\in\P_\H$. For some of the results stated elsewhere (e.g.
theorem \ref{fundthm}), functions formed by taking suprema over $\H$ also
have to be measurable. This is assured if $\H$ is {\em permissible}. The
following two definitions are taken (with minor modifications) from Pollard
(1984)\label{PE2}.
\begin{defn}
\label{index}
A family, $\H$, of $[0,M]$-valued functions on a set $Z$
is {\em indexed} by the set $T$ if
there exists a function $f\colon Z\times T\to [0,M]$ such that 
$$
\H = \left\{f(\,\cdot\,,t)\colon t\in T\right\}.
$$ 
\end{defn}
\begin{defn}
\label{permissible}
The set $\H$ is {\em permissible} if it can be indexed by a set $T$ 
such that
\begin{enumerate}
\item $T$ is an {\em analytic} subset of a Polish space $\Tbar$, and
\item the function $f\colon Z\times T\to [0,M]$ indexing $\H$ by $T$ is measurable
with respect to the product $\sigma$-algebra $\sigma_\H\otimes \sigma(T)$, 
where
$\sigma(T)$ is the Borel $\sigma$-algebra induced by the topology on $T$.
\end{enumerate}
\end{defn}
An analytic subset $T$ of a Polish space $\Tbar$ is simply the
continuous image of a Borel subset $X$ of another Polish space $\Xbar$.
The analytic subsets of a Polish space include the Borel sets. They are
important because projections of analytic sets are analytic, and can be
measured in a complete measure space whereas
projections of Borel sets are not necessarily Borel, and hence cannot be
measured with a Borel measure. For more details see Dudley (1989)\label{DL1}, 
section 13.2.
\begin{defn}
\label{htimes}
Let $\H_1,\dots,\H_n$ be $n$ sets of functions mapping $Z$ into
$[0,M]$.  For all $h_1\in\H_1,\dots,h_n\in\H_n$, define
\glsname{h_1h_n} by
$$
h_1\oplus\dots\oplus h_n(\zv) = \frac1n\sum_{i=1}^n h_i(z_i),
$$ for all $\zv=(z_1,\dots,z_n) \in \Zn$.  Denote the set of all such
functions by \glsname{H_1H_n}.  Occasionally $(h_1,\dots,h_n)$ and
$\hv$ will also be used to denote the function $h_1\oplus\dots\oplus
h_n$.
\end{defn}
\begin{lem}
\label{htimespermiss}
$\H_1\oplus\dots\oplus\H_n\colon \Zn\to [0,M]$ is permissible if\,
$\H_1,\dots,\H_n$ are all permissible.
\end{lem}
\begin{pf}
Let each $\H_i$ be indexed by
$f_i\colon Z\times T_i\to [0,M]$ where $T_i$ is an analytic subset of the 
Polish space
$\Tbar_i$. Then $\H_1\oplus\dots\oplus\H_n$ is indexed by
$f\colon Z^n\times T_1\times\dots\times T_n\to [0,M]$ where
$$
f(\zv,t_1,\dots,t_n) = \frac1n \sum_{i=1}^n f_i(z_i,t_i),
$$
for all $\zv=(z_1,\dots,z_n)\in Z^n$ and $t_i\in T_i, 1\leq i\leq n$.  
$\Tbar_1\times\dots\times\Tbar_n$ with product topology 
is itself a Polish space (Dudley (1989)\label{DL2}, section
13.2) and it is easily verified that $T_1\times\dots\times T_n$ is an
analytic subset of $\Tbar_1\times\dots\times\Tbar_n$. Also $f$ is 
clearly measurable with respect to the product
$\sigma$-algebra $\sigma_{\H_1\oplus\dots\oplus\H_n}
\otimes \sigma_{T_1\times\dots\times T_n}$.
The result follows.
\end{pf}

For the representation learning results of chapter \ref{repchap} and 
particularly the
more general results of section \ref{genframe} to hold, the concept of
permissibility must be introduced for {\em hypothesis space families}, that
is, sets $H=\{\H\}$ where each $\H\in H$ is itself a set of functions
mapping $Z$ into $[0,M]$. 

\begin{defn}
$H$ is {\em f-permissible\jfootnote{``f'' for {\em family}.}}
if there exist sets
$S$ and $T$ that are analytic subsets of Polish spaces $\Sbar$ and $\Tbar$
respectively, and 
a function $f\colon Z\times T\times S\to [0,M]$, measurable with respect to 
$\sigma_{H_\sigma}\otimes\sigma(T)\otimes\sigma(S)$, such that 
$$
H = \bigl\{\{f(\,\cdot\,,t,s)\colon  t\in T\}\colon  s\in S\bigr\}.
$$
\end{defn}

\begin{defn}
For any hypothesis space family $H$, define $\glsname{H_sigma} =
\{h\colon h\in\H\colon \H\in H\}$ and define $\glsname{HHn}
=\{\H\oplus\dots\oplus\H\colon \H\in H\}$. For all $h\in H_\sigma$
define $\hbar\colon \P_{H_\sigma}\to[0,M]$ by $\hbar(P) =
\E{h}{P}$. Let $\overline{H}_\sigma = \{\hbar\colon h\in H_\sigma\}$
and for all $\H\in H$ let $\Hbar = \{\hbar\colon h\in\H\}$.  For all
$\H\in H$ define $\H^*\colon \P_{H_\sigma}\to [0,M]$ by $\H^*(P) =
\inf_{h\in\H} \hbar(P)$. Let $H^* = \{\H^*\colon \H\in H\}$.  For any
probability measure $Q\in \P_{\overline{H}_\sigma}$, define the
probability measure $Q_Z$ on $Z$ by $Q_Z(S) = \int_{\P_{H_\sigma}}
P(S)\,dQ(P)$, for all $S\in \sigma_{H_\sigma}$.
\end{defn}

\begin{lem}
\label{bigpermlem}
Let $H$ be a family of hypothesis spaces mapping $Z$ into
$[0,M]$. Take the $\sigma$-algebra on $Z$ to be $\sigma_{H_\sigma}$, the set
of probability measures on $Z$ to be $\P_{H_\sigma}$ and the $\sigma$-algebra
on $\P_{H_\sigma}$ to be $\sigma_{\overline{H}_\sigma}$.
With these choices, if $H$ is f-permissible then
\begin{enumerate}
\item\label{1} $H_\sigma$ is permissible.
\item\label{2} $H^n$ is f-permissible.
\item\label{3} $[H^n]_\sigma$ is permissible.
\item\label{4} $\overline{H}_\sigma$ is permissible.
\item\label{5} $\Hbar$ is permissible for all $\H\in H$.
\item\label{6} $\H$ is permissible for all $\H\in H$.
\item\label{7} $\H^*$ is measurable for all $\H\in H$.
\item\label{8} $H^*$ is permissible.
\item\label{9} For all $Q\in \P_{\overline{H}_\sigma}$, $Q_Z\in \P_{H_\sigma}$.
\end{enumerate}
\end{lem}
\begin{pf}
Let $H$ be indexed by $f\colon Z\times T\times S\to [0,M]$ in the
appropriate way. 
By taking the product topology on $T\times S$, the same construction used in
the proof of lemma \ref{htimespermiss} can be used to show (\ref{1}) and 
(\ref{2}). (\ref{3}), (\ref{4}) and (\ref{5}) 
are then direct consequences of (\ref{1})
and (\ref{2}), while (\ref{6}) follows immediately from f-permissibility of
$H$. (\ref{7}) follows from (\ref{5}) using an identical argument to that
used by Pollard (1984) in the `Measurable Suprema' section of his
appendix C. 

(\ref{8}) is more problematic. To
begin with it must be further assumed that $(\P_{H_\sigma},
\sigma_{\overline{H}_\sigma}, Q)$ is a complete probability space. If this
is not the case, simply add all sets of $Q$-measure zero to
$\sigma_{\overline{H}_\sigma}$ and generate a new $\sigma$-algebra from the
result.
Let $\(X,\sigma_X,\mu\)$ be 
a measure space and $T$ be an analytic
subset of a Polish space. Let $\A(X)$ denote the analytic subsets of any set
$X$ (note that the analytic sets actually depend on $\sigma_X$, but it seems
to be convention to write $\A(X)$).
The following three facts about analytic sets are stated in
Pollard (1984)\label{PE3}, appendix C.
\newcounter{bean}
\setcounter{bean}{1}
\begin{list}{(\alph{bean})}{\usecounter{bean}}
\item If $(X,\sigma_X,\mu)$ is a complete probability space 
then $\A(X) \subseteq \sigma_X$\footnote{Pollard states this property
incorrectly as $\A(X) = \sigma_X$ for a complete probability space.}.
\item $\A(X\times T)$ contains the product $\sigma$-algebra
$\sigma_X\otimes\sigma(T)$.
\item For any set $Y$ in $\A(X \times T)$, the projection $\pi_X Y$ of $Y$
onto $X$ is in $\A(X)$.
\end{list}
Now, define 
$g\colon\P_{H_\sigma}\times T\times S\to [0,M]$ by $g(P,t,s) = \int_Z
f(z,t,s)\, dP(z)$ where $f$ indexes $H$ as above. 
By Fubini's theorem and the construction of
$\sigma_{\overline{H}_\sigma}$, $g$ is a
$\sigma_{\overline{H}_\sigma}\otimes\sigma(T)\otimes\sigma(S)$-measurable
function. Let $G\colon \P_{H_\sigma}\times S\to [0,M]$ be defined by 
$G(P,s) = \inf_{t\in T} g(P,t,s)$. $G$ indexes $H^*$ in the appropriate way
for $H^*$ to be permissible, as long as it can be shown that $G$ is
$\sigma_{\overline{H}_\sigma}\otimes \sigma(S)$-measurable. This is where
analyticity 
becomes important. 
Let $g_\alpha = \{(P,t,s)\colon g(P,t,s) > \alpha\}$. By property (b) above,
$\A\(\P_{H_\sigma}\times T\times S\)$ contains $g_\alpha$. The set $G_\alpha
= \{(P,s)\colon G(P,s) > \alpha\}$ is the projection of $g_\alpha$ onto
$\P_{H_\sigma}\times S$, which by property (c) is also analytic. As
$(\P_{H_\sigma}, \sigma_{\overline{H}_\sigma}, Q)$ is assumed 
complete, $G_\alpha$ is measurable, by property (a). 
Thus $G$ is a measurable function and the permissibility of $H^*$ follows.

For (\ref{9}) note that for all $h\in H_\sigma$,
$$
\int_{\P_{H_\sigma}} \hbar(P)\,dQ(P) = \int_Z h(z)\,dQ_Z(z),
$$
and so each $h\in H_\sigma$ is $Q_Z$-measurable.
\end{pf}

\chapter{The Fundamental Theroem}
\label{fundapp}
The fundamental theorem (\ref{fundthm}) bounds the probability of
large deviation between empirical estimates of the error of
functions $h_1\oplus\dots\oplus h_n$ (recall definition \ref{htimes}) and
their true error. The theorem is crucial for deriving bounds on the sample size
required for good generalisation in representation learning, hence the
description {\em fundamental}. It is used heavily in sections \ref{form} and
\ref{genframe}, although only implicitly in the latter section. 
The fundamental theorem is a generalisation
of a similar theorem for ordinary learning derived by Haussler (1992), and  
the same techniques used by Haussler in deriving his theorem are
used here. 

Before the theorem can be
stated some preliminary definitions must be introduced, which in many cases
are repetitions of definitions given in chapters \ref{ordchap} and
\ref{repchap}. They
are reproduced here for clarity and smoothness of exposition.

\section{Miscellaneous definitions}
For any set $Z$, denote the set of $m\times n$ matrices over $Z$ by
$Z^{(m,n)}$
Matrices will always be denoted by lower-case, bold-face 
characters,
$$
\z = 
\begin{matrix}
z_{11} & \hdots & z_{1n} \\
\vdots & \ddots & \vdots \\
z_{m1} & \hdots & z_{mn}.
\end{matrix}
$$
The {\em rows} of $\z$ will always be denoted by 
$\zv_1,\dots,\zv_m$

Let $\H$ be any set of functions mapping $\Zn$ into $[0,M]$. For all
$h \in \H$ and $\z\in\Zmn$, define
$$
\E{h}{\z} \de \frac1m\sum_{i=1}^m h(\zv_i).
$$
If $P$ is any probability measure on $\Zn$ then for all $h\in\H$ define 
$$
\E{h}{P} \de \int_{\Zn} h(\zv)\,\,dP(\zv).
$$
$\E{h}{\z}$ is the {\em empirical error} of $h$ on the sample $\z$ and
$\E{h}{P}$ is the {\em true error} of $h$ with respect to $P$.

The main theorem of this section is about bounding the probability of
large deviation between $\E{h}{\z}$ and $\E{h}{P}$. Such deviations are
measured with the $d_\nu$ metric.
\begin{defn}
\label{dnu}
For all $x,y\in\R^+$ and all $\nu>0$, let 
$$
\d{x}{y} \de \frac{|x-y|}{\nu+x+y}.
$$
\end{defn}
\begin{lem}
\label{dnulem}
The $d_\nu$ metric has the following three properties:
\begin{enumerate}
\item $0\leq\d{r}{s}\leq 1$.
\item For all $r\leq s\leq t$,\,\, $\d{r}{s}\leq\d{r}{t}$ and $\d{s}{t}\leq\d{r}{t}$.
\item For $0\leq r,s\leq M,\,\, \frac{|r-s|}{\nu+2M} \leq\d{r}{s}\leq
\frac{|r-s|}{\nu}$.
\end{enumerate}
\end{lem}
Following Haussler, the second property is referred to as the 
{\em compatibility of $d_\nu$ with the ordering on the reals}.

\section{Psuedo-metric spaces and $\ep$-covers.}
\begin{defn}
\label{pseudodef}
A {\em pseudo-metric space} \glsname{Xrho} is a metric space without
the constraint that $\rho(x,y) = 0 \Rightarrow x=y$.  An {\em
  $\ep$-cover} of a pseudo-metric space $\(X,\rho\)$ is a set
$T\subseteq X$ such that for all $x\in X$, there exists $t\in T$ such
that $\rho(x,t) \leq \ep$. A set $T\subseteq X$ is called {\em
  $\ep$-separated} if for all distinct $s,t\in T$, $\rho(s,t) > \ep$.
Denote the smallest $\ep$-cover of a pseudo-metric space $\(X,\rho\)$
by \glsname{NepXrho}.  Denote the largest $\ep$-separated subset by
\glsname{MepXrho}.
$\M\(\ep,X,\rho\)$ is also referred to as a {\em packing number}. $X$
is {\em totally bounded} if it has a finite $\ep$-cover for all
$\ep>0$.
\end{defn}

The following four lemmas are the only ones about pseudo-metrics and
$\ep$-covers that are needed to prove the results in
the rest of this work. They are all quite trivial.
The first is due to Kolmogorov and
Tihomirov (1961)\label{KT1}.
\begin{lem}
\label{KTlem}
If $T$ is a totally bounded subset of a pseudo-metric space $(X,\rho)$,
then for all $\ep>0$,
$$
\M\(2\ep,T,\rho\) \leq \N\(\ep,T,\rho\) \leq \M\(\ep,T,\rho\).
$$
\end{lem}

\begin{lem}
\label{Jlem}
$$
\M\(2\ep,T,\rho\) = \infty \Rightarrow \N\(\ep,T,\rho\) = \infty.
$$
\end{lem}

\begin{lem}
\label{isom}
Let $(X,\rho)$ and $(Y,\sigma)$ be totally bounded pseudo-metric spaces. Let 
$\psi\colon X\to Y$ be surjective. If $\psi$ is a {\em contraction}, i.e for all 
$x,x'\in X$
$$
\sigma\(\psi(x),\psi(x')\) \leq \rho(x,x'),
$$
then for all $\ep>0$
$$
\N(\ep,Y,\sigma) \leq \N(\ep,X,\rho),
$$
with equality always if $\psi$ is an isometry.
\end{lem}

\begin{lem}
\label{trivlem}
For all $x > 0$,
$$
\N\(\ep,\F,x\rho\) = \N\(\frac{\ep}x,\F,\rho\).
$$
\end{lem}

\subsection{Capacity.}
\begin{defn}
\label{pmetric}
Given a set of functions $\H$ from any set $Z$ into $[0,M]$ and a
probability measure $P$ on
$Z$, define the pseudo-metric $d_P$ on $\H$ by,
$$
d_P(f,g) \de \int_Z |f(z)-g(z)|\,dP(z),
$$
for all $f,g\in\H$. 
\end{defn}
Note that $d_P$ is only a pseudo-metric because $f$ and $g$ could differ 
on a set of measure zero and still be equal under $d_P$. 

\begin{defn}
\label{capdef}
Given a set of functions $\H$ from any set $Z$ into $[0,M]$,
for all $\ep>0$ let
$$
\C(\ep,\H) = \sup_{P\in\P_\H}\N\left(\ep,\H,d_P\right)
$$
when the supremum exists and $\C(\ep,\H) = \infty$ otherwise (recall
definition \ref{measure} for the definition of the set of probability
measures $\P_\H$ on $Z$). Call
$\C(\ep,\H)$ the {\em $\ep$-capacity} of $\H$.
\end{defn}
Note that this definition of capacity differs slightly from that of
Haussler. He assumes that $Z$ is a complete, separable, metric space 
and takes the $\sigma$-algebra $\Sigma$ on $Z$ to be the Borel 
$\sigma$-algebra. It is is easily verified that in the case that $\H$ 
are Borel measurable, $\sigma_\H \subseteq \Sigma$ and so the capacity
as defined here is no larger than as defined by Haussler. None of this has
any consequence for the results in this thesis.

\section{The Fundamental Theorem}
\label{fundsec}
\begin{thm}
\label{fundthm}
Let $\H\subseteq\H_1\oplus\dots\oplus\H_n$ be a permissible set of functions
mapping $\Zn$ into $[0,M]$. Let $\z\in\Zmn$ be generated by
$m\geq\frac{2M}{\alpha^2\nu}$ independent trials
from $\Zn$ according to some product probability measure
$\Pv\in\P_\H$, $\Pv=P_1\times\dots\times P_n$. For all $\nu>0$,
$0<\alpha<1$, 
$$
\Pr\left\{\z\in\Zmn\colon \exists\hv\in\H\colon \d{\E{\hv}{\z}}{\E{\hv}{\Pv}} > \alpha \right\}
\leq 4 \C\(\alpha\nu/8,\H\) e^{-\frac{\alpha^2\nu nm}{8M}}.
$$
\end{thm}
The restriction $\Pv\in\P_\H$ is only required to ensure measurability of
all functions in $\H$ with respect to $\Pv$ (recall definition \ref{measure}).
The restriction of this result to $n=1$ is used several times elsewhere 
so for convenience it is stated here as a corollary. The corollary is also a
direct consequence of Haussler's theorem 3.

\begin{cor}
\label{n=1}
Let $\H$ be a permissible set of functions from
$Z$ into $[0,M]$. Let $\zv\in\Zm$ be generated by $m$ independent trials
from $Z$ according to some probability measure
$P\in\P_\H$. For all $\nu>0$, $0<\alpha<1$,
$$
\Pr\left\{\zv\in\Zm\colon \exists h\in\H\colon \d{\E{h}{\zv}}{\E{h}{P}} > \alpha \right\}
\leq 4 \C\(\alpha\nu/8,\H\) e^{-\frac{\alpha^2\nu m}{8M}}.
$$
\end{cor}

It is the exponent of the exponential in these two results that is most
important. In the general theorem, the probablity decays exponentially in
the {\em product} $mn$, whilst in the $n=1$ case it simply decays
exponentially in $m$. Ultimately, it is the appearance of the product that
makes representation learning feasible. Note also that the restriction
$m\geq 2M/(\alpha^2\nu)$ in the fundamental theorem does not actually have any
effect on the results presented elsewhere in the work because the bound is
always automatically exceeded (look for example at
the bound on $m$ in theorem \ref{mainthm}).

The proof of theorem \ref{fundthm} is quite involved. The path followed here
differs only where necessary from that taken by Haussler in proving his
theorem 3, although  
some delicate footwork is required in translating from the $n=1$ case 
to the higher dimensional case. 

\begin{defn}
\label{Hz}
Let $\H\subseteq\H_1\oplus\dots\oplus\H_n$ be a set of real-valued functions
on $\Zn$. For all $\hv=h_1\oplus\dots\oplus h_n\in\H$ 
and all $\z\in\Zmn$, define $\hv(\z)\in\R^{(m,n)}$ by
$$
\hv(\z) \de
\begin{matrix}
h_1(z_{11}) & \hdots & h_n(z_{1n}) \\
\vdots & \ddots & \vdots \\
h_1(z_{m1}) & \hdots & h_n(z_{mn}),
\end{matrix}
$$
and let $\H_{|\z}=\left\{\hv(\z)\colon \hv\in\H\right\}$ be the set of all such points.
Note that the notation is being overloaded here because the functions
$\hv\in\H$ are now defined on two separate spaces ($\Zn$ and $Z^{(m,n)}$) and 
moreover have different ranges ($\R$ and $\R^{(m,n)}$)in each case. However
this is not a problem  as it will always be
clear from the context which version of $\hv$ is meant.
Denote the function $\z\mapsto \H_{|\z}$ for all $\z\in\Zmn$ by
$\H_{|(m,n)}$.
\end{defn}
\begin{defn}
For all $\z\in \Ztmn$, let $\z(1)$ be the top half of $\z$ and $\z(2)$ be
the bottom half, viz:
$$
\z(1) = 
\begin{matrix}
z_{11} & \hdots & z_{1n} \\
\vdots & \ddots & \vdots \\
z_{m1} & \hdots & z_{mn} 
\end{matrix}
\qquad \z(2) =
\begin{matrix}
z_{m+11} & \hdots & z_{m+1n} \\
\vdots & \ddots & \vdots \\
z_{2m1} & \hdots & z_{2mn}
\end{matrix}
$$
\end{defn}
\begin{defn}
\label{dmusig}
For all integers $m,n\geq 1$ let $\Gtmn$ denote the set of all permutations
$\sigma$ of the sequence of pairs of integers 
$\left\{11,\dots,1n,\dots,2m\,1,\dots,2m\,n\right\}$
such that for all $i$, $1\leq i\leq m$, either
$\sigma(ij) = m\!+\!i\,j$ and $\sigma(m\!+\!ij) = ij$ or $\sigma(ij) = ij$ and 
$\sigma(m\!+\!i\,j) = m\!+\!i\,j$. Thus the permutations in $\Gtmn$ swap selected
indices in the first half of the sequence  $\left\{11,\dots,2m\,n\right\}$
with corresponding
indices in the second half. For any $\x\in\R^{(2m,n)}$
and $\sigma\in\Gtmn$, let 
\begin{align*}
\mu_1(\x,\sigma) &= \frac1{mn}\sum_{i=1}^m \sum_{j=1}^n \x_{\sigma(ij)}, \\
\mu_2(\x,\sigma) &= \frac1{mn}\sum_{i=m+1}^{2m} \sum_{j=1}^n \x_{\sigma(ij)}.
\end{align*}
Given any metric $d$ on $\R^+$, extend it to a pseudo-metric
\mathbold{d} on $\Rptm$ by
$$
\mathbold{d}(\x,\y) = \max_{\sigma\in\Gtmn} d(\mu_1(\x,\sigma),\mu_1(\y,\sigma)) +
                                        d(\mu_2(\x,\sigma),\mu_2(\y,\sigma)),
$$					
for all $\x,\y\in\Rptm$.
\end{defn}
\begin{defn}
\label{dl}
Define the pseudo-metric $\dl$ on $(\R^+)^{(m,n)}$ by,
$$
\dl(\x,\y) =
\frac1{mn}\sum_{i=1}^m\left|\sum_{j=1}^n\(x_{ij}-y_{ij}\)\right|,
$$
for all $\x,\y\in (\R^+)^{(m,n)}$. Note that this is {\em not} the extension described
in the previous definition of the $L^1$ metric on $\R^+$ to $(\R^+)^{(m,n)}$
(which would, in any case, be denoted by $\mathbold{L^1}$.).
\end{defn}
The following lemma follows directly from Haussler's lemma 10.
\begin{lem}
\label{bigd}
For any metric $d$ on $\R^+$, for all $\x,\y\in\Rptm$ and any 
$\sigma\in\Gtmn$, 
$$
d\(\mu_1(\x,\sigma),\mu_2(\x,\sigma)\) \leq
d\(\mu_1(\y,\sigma),\mu_2(\y,\sigma)\) + \mathbold{d}(\x,\y).
$$
\end{lem}
\begin{lem}
\label{bigdlem}
For any $\x,\y\in\Rptm$ and $\nu>0$,
$$
\mathbold{d_\nu}(\x,\y) \leq \frac2\nu\, \dl(\x,\y).
$$
\end{lem}
\begin{pf}
For any $\sigma\in\Gtmn$ and any $\x,\y\in\Rptm$,
\begin{align*}
\begin{split}
&\d{\mu_1(\x,\sigma)}{\mu_1(\y,\sigma)} + 
\d{\mu_2(\x,\sigma)}{\mu_2(\y,\sigma)} \\
&= \frac{\left| \sum_{i=1}^m \sum_{j=1}^n
\left(\x_{\sigma(ij)} - \y_{\sigma(ij)} \right) \right|}
{\nu m n + \sum_{i=1}^m \sum_{j=1}^n\left(\x_{\sigma(ij)} +
\y_{\sigma(ij)}\right)}
+
 \frac{\left| \sum_{i=m+1}^{2m} \sum_{j=1}^n
\left(\x_{\sigma(ij)} - \y_{\sigma(ij)} \right) \right|}
{\nu m n + \sum_{i=m+1}^{2m} \sum_{j=1}^n\left(\x_{\sigma(ij)} +
\y_{\sigma(ij)}\right)} \\
&\leq  \frac{\sum_{i=1}^{2m}\left|\sum_{j=1}^n
\left(\x_{\sigma(ij)} - \y_{\sigma(ij)} \right) \right|}
{\nu m n} \\
&= \frac2\nu\,\dl(\x,\y).
\end{split}
\end{align*}
\end{pf}

The next lemma follows from Haussler's lemma 11.
\begin{lem}
\label{permlem}
If a permutation $\sigma\in\Gtmn$ is chosen uniformly at random, then for
all $\nu>0$, $0<\alpha<1$, and $\x\in [0,M]^{(2m,n)}$,
$$
\Pr\left\{\sigma\in\Gtmn\colon \d{\mu_1(\x,\sigma)}{\mu_2(\x,\sigma)} > \alpha\right\}
\leq 2 e^{-\frac{2\alpha^2\nu m n}{M}}.
$$
\end{lem}

The following lemma relates the probability of large deviation between
the empirical and true error of a function $h$ to the probability of large 
deviation between two independent empirical estimates of $h$'s error. It
follows from Haussler's lemma 12.
\begin{lem}
\label{empdev}
Let $\H$ be a permissible set of functions from $\Zn$ into $[0,M]$ and
$P\in\P_\H$ be
a probability measure on $\Zn$. For all 
$\nu>0, 0<\alpha<1$ and $m\geq\frac{2M}{\alpha^2 \nu}$,
\begin{multline*}
\Pr\left\{\z\in \Zmn\colon \exists h\in\H\colon  \d{\E{h}{\z}}{\E{h}{P}} 
> \alpha\right\} \\
\leq 2 \Pr\left\{\z\in Z^{(2m,n)}\colon  \exists h\in\H\colon 
\d{\E{h}{\z(1)}}{\E{h}{\z(2)}} > \frac{\alpha}{2}\right\}.
\end{multline*}
\end{lem}

Recalling definition \ref{Hz}, note that for all $\ep>0$,
$\N\(\ep, \H_{|\z}, \mathbold{d_\nu}\)$ is a random variable on $\Ztmn$,
where $\mathbold{d_\nu}$ is the $\Rptm$ extension of $d_\nu$ given in definition
\ref{dmusig}. 
The following lemma bounds the probability in theorem \ref{fundthm} in
terms of the expected value of this random variable.
\begin{lem}
Let $\H\subseteq\H_1\oplus\dots\oplus\H_n$ be a 
permissible set of functions from
$\Zn$ into $[0,M]$.  Suppose that $\z\in\Zmn$  
is generated by $m\geq \frac{2M}{\alpha^2\nu}$ independent trials from
$\Zn$ according to some product probability 
measure $\Pv\in\P_\H$, $\Pv=P_1\times\dots\times P_n$.
Then for all $\nu>0$ and $0<\alpha<1$,
\begin{multline*}
\Pr\left\{\z\in\Zmn\colon \exists \hv\in\H\colon \d{\E{\hv}{\Pv}}{\E{\hv}{\z}} >
\alpha\right\} \\
\leq 4 \E{\N\(\alpha/4,\H_{|(2m,n)},\mathbold{d_\nu}\)}{\Pv^{2m}} 
e^{-\frac{\alpha^2\nu nm}{8 M}}.
\end{multline*}
\end{lem}
\begin{pf}
By lemma \ref{empdev},
\begin{multline*}
\Pr\left\{\z\in\Zmn\colon \exists \hv\in\H\colon \d{\E{\hv}{\Pv}}{\E{\hv}{\z}} >
\alpha\right\} \\
\leq 2 \Pr\left\{\z\in Z^{(2m,n)}\colon  \exists \hv\in\H\colon 
\d{\E{\hv}{\z(1)}}{\E{\hv}{\z(2)}} > \frac{\alpha}{2}\right\},
\end{multline*}
so only the latter quantity needs bounding.

First, for any $\z\in\Ztmn$ and any $\sigma\in\Gtmn$, let 
$$
\z_\sigma =
\begin{matrix}
z_{\sigma(11)} & \hdots & z_{\sigma(1n)} \\
\vdots & \ddots & \vdots \\
z_{\sigma(m1)} & \hdots & z_{\sigma(mn)}.
\end{matrix}
$$
For any fixed function $\hv\in\H$ and fixed $\z\in\Ztmn$, if 
$\sigma\in\Gtmn$ is chosen uniformly at random, lemma \ref{permlem} implies
$$
\Pr\left\{\sigma\in\Gtmn\colon 
\d{\E{\hv}{\z_\sigma(1)}}{\E{\hv}{\z_\sigma(2)}} > \frac{\alpha}{2}\right\} \leq
2e^{-\frac{\alpha^2\nu m n}{2 M}}.
$$
Now suppose $H$ is an $\alpha/4$-cover for $\(\H_{|\z}, \mathbold{d_\nu}\)$, 
and suppose $\sigma\in\Gtmn$ is such that 
\begin{equation}
\label{one}
\d{\E{\hv}{\z_\sigma(1)}}{\E{\hv}{\z_\sigma(2)}} > \frac{\alpha}{2}.
\end{equation}
As $H$ is an $\alpha/4$-cover for $\(\H_{|\z}, \mathbold{d_\nu}\)$, for all 
$\hv\in\H$, there exists $\hv'(\z)\in H$
such that 
\begin{equation}
\label{two}
\mathbold{d_\nu}\(\hv(\z),\hv'(\z)\) \leq \alpha/4.
\end{equation}
Equations \ref{one} and \ref{two}, along with lemma \ref{bigd}, imply,
$$
\d{\E{\hv'}{\z_\sigma(1)}}{\E{\hv'}{\z_\sigma(2)}} > \frac{\alpha}{4}.
$$
Thus 
\begin{align*}
&\Pr\left\{\sigma\in\Gtmn\colon \exists\hv\in\H\colon 
\d{\E{\hv}{\z_\sigma(1)}}{\E{\hv}{\z_\sigma(2)}} > \frac{\alpha}{2}\right\} \\
&\qquad\leq 
\Pr\left\{\sigma\in\Gtmn\colon \exists\hv'(\z)\in H\colon 
\d{\E{\hv'}{\z_\sigma(1)}}{\E{\hv'}{\z_\sigma(2)}} > \frac{\alpha}{4}\right\} \\
&\qquad\leq
2\N\(\alpha/4,\H_{|\z},\mathbold{d_\nu}\) e^{-\frac{\alpha^2\nu m n}{8 M}},
\end{align*}
the last inequality following from lemma \ref{permlem}.
If $\z\in\Ztmn$ is also selected at random according to the measure $\Pv$ in
the statement of the lemma, then
\begin{multline*}
\Pr\left\{\sigma\in\Gamma_{2mn},\z\in\Ztmn\colon \exists\hv\in \H\colon 
\d{\E{\hv}{\z_\sigma(1)}}{\E{\hv}{\z_\sigma(2)}} > \frac\alpha2\right\} \\
\leq 2\E{\N\(\alpha/4,\H_{|\z},\mathbold{d_\nu}\)}{\Pv^{2m}}
e^{-\frac{\alpha^2\nu m n}{8M}}.
\end{multline*}
As all the components of $\z\in\Ztmn$ are selected independently (although not
identically) and $\sigma$ only ever swaps components of $\z$ from ``under''
the same probability measure $P_i$\jfootnote{This is the only point
where the fact that $\Pv$ is a {\em
product} measure on $\Zn$ has been used. However it should be clear that it
is crucial in ensuring the argument carries through.}, the effect of
choosing $\sigma\in\Gtmn$ uniformly can be integrated out of the above bound,
which can then be written simply as
$$
\Pr\left\{\z\in\Ztmn\colon \exists\hv\in \H\colon 
\d{\E{\hv}{\z(1)}}{\E{\hv}{\z(2)}} > \frac\alpha2\right\}.
$$
The result follows.
\end{pf}
One more lemma is needed to prove the fundamental theorem.
\begin{lem}
For all $\alpha>0$ and all $\z\in Z^{(m,n)}$,
$$
\N\(\alpha,\H_{|\z},\mathbold{d_\nu}\) \leq \N\(\alpha\nu/2,\H, d_{P_\z}\),
$$
where in the right hand side $\z$ should be viewed as an $m$-fold sample 
from $\Zn$ and $P_\z$ as the corresponding empirical measure\jfootnote{That is, for any set $S\subseteq\Zn$, $P_\z(S) =
|S\cap\z|/|\z|$, where $\z$ is interpreted as a subset of $\Zn$ by
collecting all its rows together, $\z\equiv\{\zv_1,\dots,\zv_m\}$.} on
$\Zn$.
\end{lem}
\begin{pf}
By lemmas \ref{bigdlem}, \ref{isom} and \ref{trivlem},
$$
\N\(\alpha,\H_{|\z}, \mathbold{d_\nu}\) \leq 
\N\(\alpha\nu/2,\H_{|\z}, \dl\),
$$
and observe that the map $\psi\colon \(\H,d_{P_\z}\)\to\(\H_{|\z},\dl\)$, where
$\psi(\hv) = \hv(\z)$ for all $\hv\in\H$, is an isometry. Thus by
lemma \ref{isom} again the result follows.
\end{pf}

The proof of the fundamental theorem (\ref{fundthm}) follows immediately from
the last two lemmas.

\chapter{Bounding $\mathbold{\C(\ep,\H)}$}
\label{capapp}
For the fundamental theorem to be useful bounds need to be found
for the capacity $\C\(\ep,\H\)$ when $\H$ consists of classes of functions
common in machine learning applications. Haussler (1992)
showed how to do this for classes constructed out of compositions and free
products of certain simple function classes.
His results can be used to
determine the capacity of artificial neural networks (see his paper, section 
7). One slightly unsatisfactory aspect of his method is that somewhat arbitrary
metrics must be introduced on the simple function classes for the analysis
to carry through. In this section it is shown that such a restriction is
unnecessary: there is a {\em natural} metric---generated by the function class
as a whole---that can be defined for
any small function class embedded within a larger composition or product. 
This leads to simpler and more general formulae than those given in
Haussler, section 6. As an application, the capacity of the function space
$l_\comp{\G^n}{\Fbar}$ (see section \ref{gnfsec}) is calculated in terms of the
capacity of $\G$ and $\F$.
To begin with some
preliminary definitions are needed in order that a fundamental
and elegant theorem of Pollard's (1984)\label{PE4}---central to Haussler's
approach---may be understood.

\section{Pseudo-Dimension}
The capacity of a collection of real-valued functions can be bounded in 
terms of a single parameter known as the {\em pseudo-dimension}.
The pseudo-dimension is a real-valued extension of a similar concept 
called the {\em $VC$-dimension} introduced by Vapnik and Chervonenkis
(1971)\label{VC1}
to cover uniform convergence of empirical estimates over
Boolean-valued function classes. The pseudo-dimension was named as such 
by Haussler (1992), but essentially defined by Pollard (1984)\label{PE5}.

\begin{defn}
\label{orthant}
The notion of quadrant in $\R^2$ and octant in $\R^3$ generalises 
naturally to the concept of {\em orthant} in $\R^d$. 
That is, the orthants in $\R^d$ are the regions in which
the sign of the coordinates of each point are all the same (take the sign of
zero to be positive). Clearly
there are $2^d$ orthants in $\R^d$. 
\end{defn}
\begin{defn}
\label{F_Y}
Given a class of functions $\H\colon Z\to\R$ and any finite subset $Y$
of $Z$, define $\H_{|Y}\subset\R^{|Y|}$ by (for a similar notion
recall definition \ref{Hz})
$$
\H_{|Y} \de \left\{(h(y_1),\dots,h(y_{|Y|}))\colon h\in\H\right\}.
$$
\end{defn}
\begin{defn}
\label{shatter}
A set of real-valued functions $\H\colon Z\to\R$ is said to {\em
shatter} a finite set $Y\subseteq Z$ if 
there exists a {\em translation} of $\H_{|Y}$ 
by $x\in\R^{|Y|}$ such that $\H_{|Y} + x$ has a point in each 
orthant of $\R^{|Y|}$.
\end{defn}
Note that ``splatter'' would be a more appropriate term here as the points
of $Y$ are being {\em spread out} over all orthants by the action of $\H$,
however ``shatter'' appears to have stuck. 
\begin{defn}
The {\em pseudo-dimension} of a set of real
valued functions $\H\colon Z\to \R$ is the size of the largest subset
$Y\subseteq Z$ shattered by $\H$. If there is no largest such set then
set the pseudo-dimension of $\H$ equal to infinity.  Following
Haussler the pseudo-dimension of $\H$ will be denoted by
\glsname{pseudo-dimension}.
\end{defn} 

The term pseudo-dimension is justified by the following theorem due 
to Dudley (1978)\label{DE1}.
\begin{thm}
\label{vspace}
If $\H$ is a $d$-dimensional vector space of functions
from any set $Z$ into $\R$, $\dimp(\H) = d$.
\end{thm}
Another useful result due to Wenocur and Dudley (1981)\label{WD1} :
\begin{thm}
\label{monotone}
Let $\F$ be any family of functions from $Z$ into some interval
$I\subseteq\R$. Let $h$ be a fixed monotone function from
$I$ into $\R$ and let $\H=\{\comp{h}{f}\colon f\in\F\}$. Then $\dimp(\H) \leq
\dimp(\F)$ with equality if $h$ is strictly increasing or strictly
decreasing. 
\end{thm}

\sloppy
The following elegant theorem due to Pollard (1984)\label{PE6} illustrates
the significance of the pseudo-dimension in machine learning applications.
(Recall definitions \ref{pseudodef} and \ref{pmetric} for the definitions
of the packing numbers $\M\(\ep,\H,d_P\)$ and the pseudo-metric $d_P$).
\begin{thm}
\label{Poll}
Suppose the family of functions $\H\colon Z\to[0,M]$ has 
finite pseudo-dimension, $d$. Then for all $0<\ep\leq M$ and any probability 
measure $P$ on $Z$,
$$
\M\(\ep, \H, d_P\) < 2 \(\frac{2 e M}{\ep}ln\frac{2 e M}{\ep}\)^d.
$$
\end{thm}
By lemma \ref{KTlem},
$\N\(\ep,\H,d_P\) \leq \M\(\ep,\H,d_P\)$ and so 
$$
\C\(\ep,\H\) < 2 \(\frac{2 e M}{\ep}ln\frac{2 e M}{\ep}\)^d.
$$

\fussy
Hence if the pseudo-dimension of a class of functions is known, its capacity can
immediately be bounded. Thus the capacities of
the simple classes mentioned in theorems \ref{vspace} and \ref{monotone} can
be bounded already.
Most function  classes used in practice--artificial neural
networks, radial basis function networks, etc.--are constructed from
compositions and products of such simple classes. Techniques for bounding
the capacity of products and compositions are introduced below.
However before embarking on that project one more section is required 
introducing Haussler's decision-theoretic formulation of machine learning
which enables results bounding capacities to apply in a multitude of different
learning scenarios.

\section{Loss Functions}
\label{loss}
Without going into too much detail as to why (see section \ref{genprob} for
some motivation and Haussler section 1.1 for still more),
rather than a set of maps $\H\colon Z\to [0,M]$ the  
following structure is used:
$$
\begin{array}[b]{ccc}
X\stackrel{\H}{\longrightarrow} &A & \\
                          &\times &\stackrel{l}{\longrightarrow} [0,M]
\\*[6pt]
			  &Y & 
\end{array}
$$
The space $A$ is called the {\em action} space and $X$ and $Y$ are called the
{\em input} and {\em output} spaces respectively. The sample space $Z$ is
now defined to be $X\times Y$. $l$ is called the {\em loss function}.
This formulation is derived from statistical decision theory (see for
example Keifer (1987)\label{K1}).
The idea of this construction is that when provided with an input $x$, the
learner produces some action $a\in A$ and, depending on the ``state of the
environment'' $y\in Y$ corresponding to  $x$, receives a certain {\em
loss}, $l(a,y)$. The learner tries to find a {\em decision rule} or {\em
hypothesis} $h\in \H$ that minimizes its expected loss. In ordinary
classification problems using neural networks $Y$ is typically the 
set $\{0,1\}$, $A$ is the interval $[0,1]$ and $l$ is the mean-squared error
$l(y,a) = (y-a)^2$. Minimal (zero) loss then corresponds to a decision rule
that is a perfect classifier for the problem.
The main purpose of this construction is that it allows results concerning
the capacity of $\H$ to apply, with little modification, when $\H$ is used
in more general ways than simply as a set of classifiers. 

Note that for any function $h\in\H$ a function $l_h\colon
Z\to[0,M]$
may be defined by $l_h(z) = l(h(x),y)$ for all $z=(x,y)\in Z$. 
Denote the set of
all such functions by $l_\H$
The next few sections show how to 
bound the capacity of
$l_\H$ when $\H$ consists of compositions and products of simpler function
spaces. 

\section{Capacity of Compositions}
\label{compsec}
Suppose the function space $\H\colon X\to A$ is of the form 
$\H = \comp{\G}{\F}$ where the functions in $\F$ map $X$ into some set $V$,
and the functions in $\G$ map from $V$ into $A$. Denote this structure by
$$
X \xrightarrow{\F} V \xrightarrow{\G} A.
$$
For any function $g\in\G$ the function $l_g\colon V\times Y\to [0,M]$ can be
defined in analogy with the definition for $l_h$, i.e. $l_g(v,y) = l(g(v),
y)$ for all $(v,y)\in V\times Y$. Definition \ref{capdef} may then be used
directly to give a definition for the capacity of $l_\G$:
$$
\C(\ep,l_\G) = \sup_{P\in\P_{l_\G}}\N\left(\ep,l_\G,d_P\right).
$$
Note that $\P_{l_\G}$ is a set of probability measures on $V\times Y$, rather 
than $Z=X\times Y$. 

To define the capacity of $\F$ an appropriate pseudo-metric for $\F$ needs
to be found. This is done by pulling back the loss function so
that it is effectively a map from $Y\times V$ into $[0,M]$, rather than a
map from $Y\times A\to [0,M]$. 
\begin{defn}
\label{pdef}
Given the structure $X \xrightarrow{\F} V \xrightarrow{\G} A$ and any
probability measure $P\in \P_{l_\comp{\G}{\F}}$, define the
pseudo metric \glsname{d_Pl_G} on $\F$ by
$$
d_{[P,l_\G]}(f,f') \de \sup_{g\in\G} \int_Z 
\left|l_\comp{g}{f}(z) - l_\comp{g}{f'}(z)\right|\,dP(z),
$$
for all $f,f'\in \F$. 
\end{defn}
Its clear that $d_{[P,l_\G]}$ is symmetric and $d_{[P,l_\G]}(f,f)=0$ for all $f\in\F$.
The triangle inequality is also pretty trivial but it is proved here 
for completeness. For any $f_1,f_2,f_3 \in \F$, 
\begin{align*}
d_{[P,l_\G]}(f_1,f_2) &+ d_{[P,l_\G]}(f_2,f_3) \\
&= \sup_{g\in\G}\int_Z\left|l_\comp{g}{f_1}(z) -
			 l_\comp{g}{f_2}(z)\right|\,dP(z)
+ \sup_{g\in\G}\int_Z\left|l_\comp{g}{f_2}(z) -
				l_\comp{g}{f_3}(z)\right|\,dP(z) \\
&\geq \sup_{g\in\G}\int_Z\left|l_\comp{g}{f_1}(z) -
				l_\comp{g}{f_2}(z)\right| 
+ \left|l_\comp{g}{f_2}(z) -
		l_\comp{g}{f_3}(z)\right|\,dP(z) \\
&\geq \sup_{g\in\G}\int_Z\left|l_\comp{g}{f_1}(z) -
				l_\comp{g}{f_3}(z)\right|\,dP(z) \\
&= d_{[P,l_\G]}(f_1,f_3).
\end{align*}
Thus $d_{[P,l_\G]}$ is indeed a pseudo-metric.
\begin{defn}
\label{capf}
For the structure $X \xrightarrow{\F}V \xrightarrow{\G} A$, define the capacity of $\F$ by
$$
\glsname{C_l_GepF} = \sup_{P\in\P_{l_\comp{\G}{\F}}} \N\(\ep,\F, d_{[P,l_\G]}\).
$$
\end{defn}
\begin{defn}
For any probability measure $P\in\P_{l_\comp{\G}{\F}}$ on $X\times Y$ 
and any function $f\colon X\to V$, define the measure $P_f$ on $V\times Y$ by
$$
P_f(S) = P(f^{-1}(S))
$$
for any $S\in \sigma_{l_\G}$\jfootnote{This definition implicitly assumes
$f$ is a measurable function from $\(X,\sigma_{l_\comp{\G}{\F}}\)$
into $\(V,\sigma_{l_\G}\)$. This is guaranteed by the definition of
$\sigma_{l_\G}$ and $\sigma_{l_\comp{\G}{\F}}$.}($f^{-1}(S)$ is defined to be
the set $\{(x,y)\in X\times Y\colon (f(x),y)\in S\}$).
\end{defn}
\begin{lem}
\label{litcomplem}
Let $\H\colon X \to A$ be of the form $\H = \comp{\G}{\F}$ where $X \xrightarrow{\F}V
\xrightarrow{\G} A$. For all $\ep_1,\ep_2 > 0$,
$$
\C(\ep_1+\ep_2,l_\H) \leq \C_{l_\G}(\ep_1,\F)\C(\ep_2,l_\G).
$$
\end{lem}
\begin{pf}
The idea for this proof comes from Haussler's lemma 8.
Fix $P\in\P_{l_\comp{\G}{\F}}$ and let $F$ be a minimum size $\ep_1$-cover
for $(\F,d_{[P,l_\G]})$. By definition $|F| \leq \C_{l_\G}(\ep_1,\F)$. 
For each $f\in F$ let $G_f$ be a minimum size
$\ep_2$-cover for $(l_\G,d_{P_f})$. By definition again, $|G_f| \leq
\C(\ep_2,l_\G)$ for each $f\in F$. Let $H=\{\comp{g}{f}\colon  f\in F \,{\text{and}}\,
g\in G_f\}$. Let $l_H$ be defined in the usual way. Note that $|l_H| \leq
|H| \leq \C_{l_\G}(\ep_1,\F)\C(\ep_2,l_\G)$ and so the lemma will be proved if
$l_H$ can be shown to be an $\ep_1+\ep_2$-cover of $l_\H$. Hence, given any
$l_{\comp{g}{f}}\in l_\H$ choose $f'\in F$ such that $d_{[P,l_\G]}(f,f') \leq
\ep_1$ and $g'\in \G_{f'}$ such that $d_{P_{f'}}(g,g')\leq \ep_2$. 
Now,
\begin{align*}    
d_P(l_\comp{g}{f}, l_\comp{g'}{f'}) &= 
\int_Z \left|l_\comp{g}{f} - l_\comp{g'}{f'}\right|\,dP(z) \\
&\leq \int_Z \left|l_\comp{g}{f} - l_\comp{g}{f'}\right| + 
\left|l_\comp{g}{f'} - l_\comp{g'}{f'}\right|\,dP(z) \\
&\leq d_{[P,l_\G]}(f,f') + d_{P_{f'}}(l_g,l_{g'}) \\
&\leq \ep_1 + \ep_2.
\end{align*}
Thus $l_H$ is an $\ep_1+\ep_2$-cover for $l_\H$ and so the result follows.
\end{pf}
For multiple compositions this lemma may easily be extended by induction to
give the following lemma which is the corresponding result within this
framework to Haussler's lemma 8. 
\begin{lem}
\label{complem}
Let $\H = \prd{\H}{q}{\circ}{1}$ and $\G_i=\prd{\H}{q}{\circ}{i+1}$, for
$1\leq i\leq q-1$. 
For all $\ep_1,\ldots,\ep_q > 0$
let $\ep = \sum_{i=1}^q \ep_i$. Then
$$
\C\(\ep,l_\H\) \leq \C(\ep_q,l_{\H_q})\,
\prod\limits_{i=1}^{q-1} \C_{l_{\G_i}}\(\ep_i,\H_i\).
$$
\end{lem}

\section{Capacity of Products}
In this section the problem of bounding the capacity of a 
{\em product} of function spaces is tackled. As in the
previous section a lemma is proved bounding the capacity of a product in
terms of the capacities of its individual components.

\sloppy
Let $X_1,\dots,X_n$, $Y_1,\dots,Y_n$ and $A_1,\dots,A_n$ be $n$ be separate
input, output and action spaces respectively. For each $i$, $1\leq i\leq n$, 
let $l_i\colon Y_i\times A_i\to [0,M_i]$ be a loss function and 
let $Z_i = X_i\times Y_i$.
Define $X=\pr{X}{\times}{n}, Y=\pr{Y}{\times}{n}$, $A=\pr{A}{\times}{n}$ and
$Z=\pr{Z}{\times}{n}$.
Define a loss function $l$ on $Y\times A$ by
$$
l(\yv,\av) \de \frac1n \sum_{i=1}^n l_i(y_i,a_i),
$$
for all $\yv=(\pr{y}{,}{n}) \in Y$ and $\av=(\pr{a}{,}{n})\in A$.
There is a natural identification between 
$\XY = \pr{X}{\times}{n}\,\times\,\pr{Y}{\times}{n}$ and 
$Z = X_1\times Y_1\,\times\dots\times\,X_n\times Y_n$ by
$(x_1,\dots,x_n,y_1,\dots,y_n)\leftrightarrow (x_1,y_1,\dots,x_n,y_n)$, 
hence $Z$ can equivalently be written as $Z=\XY$.
Note also that the range of $l$ is $[0,M]$ where $M=\frac1n\sum_{i=1}^n M_i$.
\fussy

The function space $\H\colon X\to A$ is assumed to be a subset of a
product of function spaces, i.e.
$\H\subseteq\H_1\times\dots\times\H_n$ where for each $i$, $1\leq i\leq n$,
$\H_i\colon X_i\to A_i$ and the action of 
$\hv=(\pr{h}{,}{n})\in\H$ is defined by
$$
\hv(x_1,\dots,x_n) = (h_1(x_1),\dots,h_n(x_n)).
$$
To tackle the problem in its full generality assume $\H$ has the structure
$\H=\comp{\G}{\F}$, where $X \xrightarrow{\F}V \xrightarrow{\G} A$ and assume that $\G$ and
$\F$ are also subsets of product function spaces: 
$\G\subseteq\G_1\times\dots\times\G_n$ and $\F\subseteq\F_1\times\dots\times\F_n$.
Given $\comp{\gv}{\fv}\in\comp{\G}{\F}$ and a loss function
$l=\frac1n\sum_{i=1}^n l_i$ as above, define $l_\comp{\gv}{\fv}$ in
the usual way\jfootnote{That is, $l_\comp{\gv}{\fv}(\zv) =
l(\comp{\gv}{\fv}(\xv),\yv)$ where $\zv = (\xv,\yv)$. Note that
$l_\comp{\gv}{\fv}(\zv) = \frac1n\sum_{i=1}^n l_i(\comp{g_i}{f_i}(x_i),y_i)$.}
and let $l_\comp{\G}{\F}$ denote the set of all
such functions.

With these definitions, for all {\em product probability
measures\jfootnote{$\P_{l_\H}$ contains more than just product
probability measures but they will be ignored.}}
$\Pv=\pr{P}{\times}{n} \in \P_{l_\H}$, 
the pseudo-metric $d_{[\Pv,l_\G]}$ can be defined on $\F$ as in definition 
\ref{pdef}, 
$$
d_{[\Pv,l_\G]}(\fv,\fv') = \sup_{\gv\in\G}\int_Z 
\left|l_\comp{\gv}{\fv}(\zv) - l_\comp{\gv}{\fv'}(\zv)\right|\, d\Pv(\zv).
$$
The pseudo-metric $d_{[P_i,l_{\G_i}]}$ on $\F_i$ also follows definition
\ref{pdef},
$$
d_{[P,l_{\G_i}]}(f,f') \de \sup_{g\in\G_i}\int_{Z_i}
\left|l_\comp{g}{f}(z) -
l_\comp{g}{f'}(z)\right|\, dP_i(z),
$$
for all $f,f'\in\F_i$\jfootnote{Strictly speaking $l_{\G_i}$ and $l_\comp{g}{f}$
should be written as ${l_i}_{\G_i}$ and ${l_i}_\comp{g}{f}$ respectively,
however notation like that is just too
ugly. Hopefully there should not be any confusion.}.

\flushbottom
\begin{lem}
\label{litlem}
Given the function space $\H=\comp{\G}{\F}$ with 
$X \xrightarrow{\F} V \xrightarrow{\G} A$,
$\F\subseteq\F_1\times\dots\times\F_n$ and
$\G\subseteq\G_1\times\dots\times\G_n$,
then for any pair of functions $\fv=\(\pr{f}{,}{n}\), \fv'=\(f_1',\dots,f_n'\)$
in $\F$ and any product probability measure 
$\Pv=\pr{P}{\times}{n}\in \P_{l_\H}$, 
\begin{equation}
d_{[\Pv,l_\G]}(\fv,\fv') \leq \frac1n\sum_{i=1}^n 
d_{[P_i, l_{\G_i}]}(f_i,f'_i).
\end{equation}
If instead $\Pv$ is product probability measure in $\P_{l_\G}$, then
for all $\gv=\(\pr{g}{,}{n}\)$, 
$\gv'=\(g'_1,\dots,g'_n\)\in \G$,
$$
d_\Pv(\gv,\gv') \leq \frac1n\sum_{i=1}^n d_{P_i}(g_i,g'_i).
$$
\end{lem}
\begin{pf}
Let $\fv, \fv'$ and $\Pv$ be as in the first inequality.
\begin{align*}
d_{[\Pv,l_\G]}&(\fv,\fv') \\ 
&= \sup_{\gv\in\G}\int_Z
\left|l_\comp{\gv}{\fv}(\zv) - l_\comp{\gv}{\fv'}(\zv) \right|\, d\Pv(\zv) \\
&= \sup_{\(\pr{g}{,}{n}\)\in\G}
\int\limits_{z_1\in Z_1}\!\!\dots\!\! \int\limits_{z_n\in Z_n} 
\left|\frac1n\sum_{i=1}^n \(l_\comp{g_i}{f_i}(z_i)
- l_\comp{g_i}{f'_i}(z_i)\)\right|\, dP_1(z_1)\dots dP_n(z_n) \\
&\leq \sup_{\(\pr{g}{,}{n}\)\in\G}
\int\limits_{z_1\in Z_1}\!\!\dots\!\!\int\limits_{z_n \in Z_n} 
\frac1n\sum_{i=1}^n\left|l_\comp{g_i}{f_i}(z_i)
- l_\comp{g_i}{f'_i}(z_i)\right|\,dP_1(z_1)\dots dP_n(z_n) \\
&=\frac1n\sum_{i=1}^n \sup_{g_i\in\G_i}\int_{Z_i}
\left|l_\comp{g_i}{f_i}(z) - l_\comp{g_i}{f'_i}(z)\right|\,dP_i(z)\\
&= \frac1n\sum_{i=1}^n d_{[P_i,l_{\G_i}]}(f_i,f'_i).
\end{align*}
The second inequality is proved in a similar way.
\end{pf}

\raggedbottom
Recalling definition \ref{capf} for the capacity $\C_{l_\G}(\ep,\F)$,
the last lemma gives the following lemma.
\begin{lem}
\label{prodlem}
For the function space $\H=\comp{\G}{\F}$ with 
$X \xrightarrow{\F} V \xrightarrow{\G} A$,
$\F\subseteq\F_1\times\dots\times\F_n$ and
$\G\subseteq\G_1\times\dots\times\G_n$,
\begin{align*}
\C_{l_\G}(\ep,\F) &\leq \prod_{i=1}^n \C_{l_{\G_i}}\(\ep,\F_i\) \\
\C(\ep,l_\G) &\leq \prod_{i=1}^n \C(\ep,l_{\G_i}).
\end{align*}
\end{lem}

\begin{pf}
For the first inequality fix $\Pv=P_i\times\dots\times P_n\in\P_{l_\H}$ and 
let $F_i$ be an $\ep$-cover for $(\F_i, d_{[P_i,l_{\G_i}]})$, $1\leq i\leq n$.
Let $F = \pr{F}{\times}{n}$. As $|F| = \prod\limits_{i=1}^n |F_i|$
the inequality will be proved if it can be shown that $F$ 
is an $\ep$-cover for $(\F, d_{[\Pv,l_\G]})$.
Let $\fv=\(\pr{f}{,}{n}\)$ be any function in $\F$. For each 
$1\leq i\leq n$ choose $f'_i\in F_i$ so that 
$d_{[P,l_{\G_i}]}(f_i, f'_i) \leq \ep$.
Let $\fv'=\(f_1',\dots,f_n'\)$. 
By lemma \ref{litlem}
$$
d_{[\Pv,l_\G]}(\fv,\fv') \leq \frac1n\sum_{i=1}^n d_{[P_i,l_{\G_i}]}(f_i,f'_i) 
\leq \ep.
$$
Thus $F$ is an $\ep$-cover for $\F$. The second inequality is proved
similarly.
\end{pf}

\section{Diagonal Function Spaces}

\sloppy
\begin{defn}
\label{dbardef}
Given a function space $\F\colon X\to V$, define $\Fbar\subseteq\F^n$
by $\Fbar=\{\(f,\dots,f\)\colon f\in\F\}$. Denote $(f,\dots,f)$ by
$\fbar$. $\fbar$ acts on $X^n$ as per usual: $\fbar(x_1,\dots,x_n) =
(f(x_1),\dots,f(x_n))$.  For any product probability measure $\Pv =
P_1\times\dots\times P_n$ on $Z^n$, define the {\em average} measure
$\Pbar$ on $Z$ by $\Pbar=\frac1n\sum_{i=1}^n P_i$.  For the structure
$X \xrightarrow{\F} V \xrightarrow{\G} A$ and for any probability
measure $P\in\P_{l_\comp{\G}{\F}}$, define the pseudo-metric
\glsname{dstar_Pl_G} on $\F$ by
$$
d^*_{[P,l_\G]}(f,f') = \int_Z 
\sup_{g\in\G} \left|l_\comp{g}{f}(z) - l_\comp{g}{f'}(z)\right|\, dP(z)
$$
for all $f,f'\in\F$.
Define
$$
\glsname{Cstar_l_GepF} = \sup_{P\in\P_{l_\comp{\G}{\F}}} \N(\ep,\F,d^*_{[P,l_\G]}).
$$
\end{defn}
\begin{lem}
\label{litbarlem}
Given the structure $X^n \xrightarrow{\Fbar} V^n \xrightarrow{\G^n} A^n$
and any product probability measure 
$\Pv=\pr{P}{\times}{n}\in\P_{l_\comp{\G^n}{\Fbar}}$,
$$
d_{[\Pv,l_{\G^n}]}(\fbar,\fbar') \le d^*_{[\Pbar,{l_\G}]}(f,f')
$$
for all $f,f'\in\F$.
\end{lem}
\begin{pf}
\begin{align*}
d_{[\Pv,l_{\G^n}]}(\fbar,\fbar') &\le \frac1n\sum_{i=1}^n d_{[P_i,l_\G]}(f,f') 
\qquad\text{(by lemma \ref{litlem})} \\
		&= \frac1n\sum_{i=1}^n\sup_{g\in\G}\int_Z
		\left|l_\comp{g}{f}(z)-l_\comp{g}{f'}(z)\right|\,dP_i(z) \\
		&\le \frac1n\sum_{i=1}^n\int_Z\sup_{g\in\G}
		\left|l_\comp{g}{f}(z)-l_\comp{g}{f'}(z)\right|\,dP_i(z) \\
		&= d^*_{[\Pbar,l_\G]}(f,f').
\end{align*}
\end{pf}
The following lemma follows immediately from this lemma, 
lemma \ref{trivlem} and lemma \ref{isom}.
\begin{lem}
\label{barlem}
For the structure $X^n \xrightarrow{\Fbar} V^n \xrightarrow{\G^n} A^n$ and any product 
probability distribution $\Pv\in\P_{l_\comp{\G^n}{\Fbar}}$ on $Z^n$,
$$
\C_{l_{\G^n}}(\ep,\Fbar) \leq \C^*_{l_\G}(\ep,\F).
$$
\end{lem}

\fussy
Now the most important capacity of all can be calculated.
\begin{thm}
\label{compgnf}
For the structure 
$$
X^n \xrightarrow{\Fbar} V^n \xrightarrow{\G^n} A^n,
$$
a loss function $l\colon Y\times A\to[0,M]$,
and all $\ep,\ep_1,\ep_2>0$ such that $\ep_1+\ep_2=\ep$,
$$
\C(\ep,l_\comp{\G^n}{\Fbar}) \leq \C(\ep_1,l_\G)^n\,\C^*_{l_\G}(\ep_2,\F).
$$
\end{thm}
\begin{pf}
The result is immediate from lemmas \ref{barlem}, \ref{complem} and
\ref{prodlem}.
\end{pf}

\chapter{Neural Network bounds.}
\label{nnetapp}
It is shown how the results of the previous chapter may be
used to determine, following Haussler again, 
the capacity of certain classes of continuous functions
that can be defined over {\em graphs}. As an application the capacity of
neural networks is determined as a function of the number of weights in the
network, using some more techniques from Haussler (1992). The capacity
bounds parallel similar results given in Haussler, in fact his main
result (theorem 11) can be derived as a special case.

\begin{defn}
A {\em feedforward continuous function network} is a continuous 
class of functions defined on a directed acyclic graph.
Nodes in the graph with no incoming edges are called {\em input} nodes
and those with no outgoing edges are called {\em output} nodes. The rest are
called {\em hidden} nodes. The hidden nodes and output nodes are known
collectively as {\em computational} nodes. To each non-input node $n$ is attached a class of
continuous functions $\H$ mapping $\R^{\text{indegree($n$)}}$ into
$\R$ where $\text{indegree($n$)}$ is the number of incoming edges to $n$.
The {\em depth} of a node is defined to be the length of the longest path
from any input node to that node. By composing all the function classes of the
graph in the obvious way, the entire graph can be viewed as 
a class of functions mapping $\R^{n_{\text{in}}}$ into $\R^{n_{\text{out}}}$ 
where $n_{\text{in}}$ and
$n_{\text{out}}$ are the number of input and output nodes of the graph respectively.
\end{defn}

By possibly adding dummy nodes that only compute the identity function, a
continuous function network can be viewed as a series of {\em layers} in which the input
nodes lie in the first layer, the output nodes in the last layer and
each hidden layer contains nodes that only compute functions of the
outputs of the nodes in the immediately preceeding layer. Hence 
the class of functions $\H$ computed by the network
may be viewed as a composition
$\H=\H_{\text{out}}\circ\H_k\circ\dots\circ\H_1$ where $k$ is the depth of
the deepest hidden node in the network, $\H_i$ is the space of functions
computed by layer $k$  and $\H_{\text{out}}$ is the space
of functions formed by the
output nodes. Denoting the number of computational
nodes in hidden layer $i$ by $n_i$ and the number of dummy nodes by $m_i$,
the entire structure of the network can be written
$$
\R^{n_{\text{in}}} \xrightarrow{\H_1} \R^{n_1+m_1} \xrightarrow{\H_2} \R^{n_2+m_2}
\longrightarrow\dots\longrightarrow\R^{n_{k-1} + m_{k-1}} \xrightarrow{\H_k} \R^{n_k+m_k} 
\xrightarrow{\H_{\text{out}}}\R^{n_{\text{out}}}.
$$

To apply the formulae of the previous section (particulary lemma
\ref{complem}), assume that there is some loss function $l\colon
\R^{n_{\text{out}}} \times Y \to \R$ measuring the network's performance.
Lemma \ref{complem} immediately yields
\begin{equation}
\label{form1}
\C\(\ep,l_\H\) \leq \C(\ep_{\text{out}},l_{\H_{\text{out}}})\,
\prod\limits_{i=1}^{k} \C_{l_{\G_i}}\(\ep_i,\H_i\),
\end{equation}
where $\G_i = \H_{\text{out}}\circ\H_k\dots\circ\H_{i+1}$ for $1\leq i \leq k$,
$\ep = \ep_{\text{out}}+\sum_{i=1}^k \ep_i$ and $\ep_i>0$ for all $1\leq i
\leq k$ and $\ep_{\text{out}} > 0$. 

To determine the capacity of each of the hidden 
layers $\H_i$ and the output layer $\H_{\text{out}}$ the following ideas from
Haussler may be used. Consider the general
problem of determining the capacities of $\F$ and $\G$ in the structure 
$$
\R^p \xrightarrow{\F} \R^q \xrightarrow{\G} \R^r.
$$
To begin with, for any function space
$\H\colon \R^p\to\R^q$ and any probability measure $P$ on $\R^p$, define the
pseudo-metric \glsname{d_L1} on $\H$ by,
$$
d_{L^1(P)}(h,h') \de \int_{\R^p} L^1(h(x),h'(x)) dP(x),
$$
for all $h,h'\in\H$. Let 
$$
\glsname{CepHd_L1} \de \sup_P \N\(\ep,\H,d_{L^1(P)}\),
$$
where the supremum is over all probablity measures $P$ on
$\R^p$\jfootnote{To be definite, assume all functions in $\H$ are measurable
with respect to the Borel $\sigma$-algebras on $\R^p$ and $\R^q$ and take the
supremum in the definition of $C\(\ep,\H, d_{L^1}\)$ to be  
over all probability measures on the Borel $\sigma$-algebra on $\R^p$.}.
Next, given a loss function $l\colon Y\times A\to [0,M]$, define the pseudo-metric
$\rho_l$ on $A$ by 
$$
\rho_l(a,b) \de \sup_{y\in Y} | l(y,a)-l(y,b) |
$$
for all $a,b\in A$.
If the action space $A$ is a subset of Euclidean space of some dimension,
then for many common machine learning scenarios it can be shown that there
exists a constant $c_l > 0$ such that $\rho_l \leq c_l L^1$ where $L^1$ is
the usual $L^1$ metric on $A$ (see Haussler, section 7). Assuming such a
bound holds, it is then trivial to show for $\G$ in the above structure
that 
\begin{equation}
\label{capg}
\C(\ep,l_\G) \leq \C\(\frac\ep{c_l},\G, d_{L^1}\).
\end{equation}
Furthermore, if  $b_\G > 0$ is a uniform Lipschitz bound on $\G$ 
(that is, $L^1(g(u), g(v)) \leq b_\G L^1(u,v)$ for all $g\in\G$ and
$u,v\in\R^q$), then the following bounds can be shown to hold for $\F$:
\begin{align}
\label{cappf}
\C_{l_\G}(\ep, \F) &\leq \C\(\frac{\ep}{c_l b_\G},\F,d_{L^1}\) \\
\C^*_{l_\G}(\ep, \F) &\leq \C\(\frac{\ep}{c_l b_\G},\F,d_{L^1}\).
\end{align}

The above formulae may now be directly applied to \eqref{form1} to yield
\begin{equation}
\label{form2}
\C\(\ep,l_\H\) \leq 
\prod\limits_{i=1}^{k+1} \C\(\frac{\ep_i}{c_l b_{\G_i}},\H_i, d_{L^1}\),
\end{equation}
where $b_{\G_i}$ is a uniform Lipschitz bound for each 
$\G_i = \H_{k+1}\circ\dots\circ\H_{i+1}, 1\leq i \leq k+1$. The formula has
been simplified by writing $\H_{k+1}$ for $\H_{\text{out}}$, setting
$b_{\G_{k+1}} = 1$ and $\ep_{k+1} = \ep_{\text{out}}$. 

Let $\F_1,\dots,\F_p$ be $p$ classes of functions mapping $\R^q$ into $\R$.
Define the {\em free product} of $\F_1$ through $\F_p$ to be the class of
functions 
$$
\F = \{(f_1,\dots,f_p)\colon  f_i \in\F_i, 1\leq i \leq p\},
$$
where $(f_1,\dots,f_p)\colon \R^q\to\R^p$ is the function defined by
$$
(f_1,\dots,f_p)(x) = (f_1(x),\dots,f_p(x)).
$$
Note that the free product differs from the product used elsewhere in the
thesis which has always been the Cartesian product.
Observe that each layer $\H_i$ in a continuous function network 
is in fact a free product of $n_i$ 
computational nodes and $m_i$ identity functions. Haussler's lemma 7 states
that the capacity of a  free product is less than the product of the
capacities of its individual components, which along with the fact that the
capacity of the identity is one\jfootnote{As is the capacity of any class
consisting of just a single function.} gives
\begin{equation}
\label{cl1}
\C(\ep,l_\H) \leq \prod\limits_{i=1}^{k+1}
\prod\limits_{j=1}^{n_i} \C\(\frac{\ep_i}{c_l b_{\G_i}},\H_{ij}, d_{L^1}\)
\end{equation}
where $\H_{ij}$ is the function class computed by the $j$'th node in layer
$i$. 

Equation \eqref{cl1} applies to any continuous function network whatsoever and
there is no further reduction to be made without knowing more about the
function classes $\H_{ij}$. In the case of neural networks, $\H_{ij}$ is
often formed by composing a fixed monotone ``squashing'' function with any 
linear combination of its inputs. Without weakening the following results it
may be assumed that $\H_{ij}$ is the composition of a monotone function with
{\em any} finite dimensional vector space of functions (not just those that
can be formed by linear combinations of the inputs). Denoting the dimension
of this space by $W_{ij}$ and assuming that the range of all functions in
$\H_{ij}$ is $[0,M]$, theorems \ref{vspace}, \ref{monotone} and
\ref{Poll} may be applied to the capacity of $\H_{ij}$ to give
$$
\C\(\frac{\ep_i}{c_l b_{\G_i}},\H_{ij}, d_{L^1}\) \leq 
2 \left[\frac{c b_{\G_i}}{\ep_i}\ln\frac{c b_{\G_i}}{\ep_i}\right]^{W_{ij}}
$$
where $c = 2 e M c_l$. Substituting this into \eqref{cl1}, setting $W_i =
\sum_{j=1}^{n_i} W_{ij}$ (the total number of ``parameters'' in layer $i$) and
$W=\sum_{i=1}^{k+1} W_i$ (the total number of parameters altogether),
yields
\begin{equation}
\label{cl2}
\C(\ep,l_\H) \leq 2^W\prod\limits_{i=1}^{k+1} \left[\frac{c b_{\G_i}}{\ep_i}
\ln \frac{c b_{\G_i}}{\ep_i}\right]^{W_i}.
\end{equation}
To obtain the best possible bound, $\ep_1,\dots,\ep_{k+1}$ should be chosen
so that the right hand side of \eqref{cl2} is minimal, subject to the condition
$\sum_{i=1}^{k+1}\ep_i = \ep$. Using Lagrange multipliers, each $\ep_i$ can be
shown to satisfy 
$$
\frac{W_i}{\ep_i}\left(1 + \frac1{\ln \frac{c b_{\G_i}}{\ep_i}}\right) =
\lambda.
$$
Although this cannot be solved exactly for $\ep_i$\jfootnote{At least not to
the knowledge of the author.}, a reasonable approximation to this system of
equations for small $\ep_i$ is
$$
\frac{W_i}{\ep_i} = \lambda.
$$
Using $\sum_{i=1}^{k+1} \ep_i = \ep$ gives,
$$
\ep_i = \frac{W_i\ep}{W}.
$$
Hence \eqref{cl2} becomes
\begin{equation}
\label{blurt}
\C(\ep,l_\H) \leq 2^W\prod\limits_{i=1}^{k+1} 
\left[\frac{c b_{\G_i}W}{W_i\ep} \ln \frac{c b_{\G_i}W}{W_i \ep}\right]^{W_i}.
\end{equation}
Denoting the average Lipschitz bound of the nodes in layer $i$ by $b_i$,
observe that each Lipschitz bound $b_{\G_i}$ must be less than or equal to
$\prod_{j=i+1}^{k+1} b_j$. Thus if $b_i>1$ for each $i$ then 
all $b_{\G_i}$ are bounded by 
$\prod_{j=2}^{k+1} b_j$. 
If all the layers in the network have the same number of parameters then
$W/W_i = k+1 = d$---the {\em depth} of the network. Using these assumptions
and the fact that $2 \ln x < x$, formula \eqref{blurt} becomes
$$
\C(\ep,l_\H) \leq \left[\frac{2 e M d \prod_{j=2}^d b_j}{\ep}\right]^{2 W}
$$
which is identical to the formula in Haussler's theorem $11$ (if one
replaces $c_2-c_1$ in his formula by $M$).

One problem about the bounds in this section  is their dependence on
Lipschitz bounds. By deriving similar bounds for hard-limiting neural
networks (which don't have Lipschitz bounds), 
Baum and Haussler (1989)\label{BH1} showed that such dependence is not always
necessary.
It remains an open problem to determine precisely the conditions under which
the Lipschitz bounds can be ignored.

\bibliographystyle{abbrv}
\bibliography{ms}

\begin{thebibliography}{10}

\bibitem{AB}
M.~Anthony and N.~Biggs.
\newblock {\em {C}omputational {L}earning {T}heory}.
\newblock Cambridge University Press, Cambridge, 1992.

\bibitem{BH}
E.~Baum and D.~Haussler.
\newblock What size net gives valid generalization.
\newblock {\em Neural Comput.}, 1:151--160, 1989.

\bibitem{BK2}
K.~L. Buescher and P.~R. Kumar.
\newblock {L}earning {S}tochastic {F}unctions by {S}mooth {S}imultaneous
  {E}stimation.
\newblock In {\em Proceedings of the Fifth Annual Conference on Computational
  Learning Theory}, New York, 1992. ACM Press.

\bibitem{CT}
T.~M. Cover and J.~A. Thomas.
\newblock {\em Elements of Information Theory}.
\newblock John Wiley \& Sons, Inc., New York, 1991.

\bibitem{Dudley78}
R.~M. Dudley.
\newblock Central limit theorems for empirical measures.
\newblock {\em Ann. Probab.}, 6:899--929, 1978.

\bibitem{Dud}
R.~M. Dudley.
\newblock {\em Real Analysis and Probability}.
\newblock Wadsworth \& Brooks/Cole, California, 1989.

\bibitem{Getal}
S.~Geman, E.~Bienenstock, and R.~Doursat.
\newblock Neural networks and the bias/variance dilemma.
\newblock {\em Neural Comput.}, 4:1--58, 1992.

\bibitem{Haussler}
D.~Haussler.
\newblock Decision theoretic generalizations of the pac model for neural net
  and other learning applications.
\newblock {\em Inform. Comput.}, 100:78--150, 1992.

\bibitem{H}
D.~Haussler, M.~Kearns, H.~S. Seung, and N.~Tishby.
\newblock Rigorous learning curve bounds from statistical mechanics.
\newblock Deposited in the Neuroprose archive., 1994.

\bibitem{Keifer}
J.~C. Keifer.
\newblock {\em Introduction to Statistical Inference}.
\newblock Springer-Verlag, New York, 1987.

\bibitem{Pollard}
D.~Pollard.
\newblock {\em Convergence of Stochastic Processes}.
\newblock Springer-Verlag, New York, 1984.

\bibitem{PollardEP}
D.~Pollard.
\newblock {\em Empirical Processes: Theory and Applications}.
\newblock NSF-CBMS Regional Conference Series in Probability and Statistics,
  Vol 2. Inst. of Math. Stat. and Am. Stat. Assoc., 1990.

\bibitem{Retal}
D.~Rumelhart, G.~Hinton, and R.~Williams.
\newblock Learning representations by back-propagating errors.
\newblock {\em Nature}, 323:533--536, 1986.

\bibitem{Valiant}
L.~G. Valiant.
\newblock A theory of the learnable.
\newblock {\em Comm. ACM}, 27:1134--1142, 1984.

\bibitem{Vapi}
V.~Vapnik.
\newblock {I}nductive {P}rinciples of the {S}earch for {E}mpirical
  {D}ependences.
\newblock In {\em Proceedings of the Second Annual Conference on Computational
  Learning Theory}, pages 2--21, New York, 1989. ACM Press.

\bibitem{VC2}
V.~N. Vapnik.
\newblock {\em Estimation of Dependences based on Empirical Data}.
\newblock Springer-Verlag, New York, 1982.

\bibitem{VC1}
V.~N. Vapnik and A.~Y. Chervonenkis.
\newblock On the uniform convergence of relative frequencies of events to their
  probabilities.
\newblock {\em Theory Probab. Appl.}, 16:264--280, 1971.

\bibitem{Edelman}
Y.~Weiss and S.~Edelman.
\newblock Representation with receptive fields: gearing up for recognition.
\newblock Technical Report CS-TR 93-09, Weizmann Institute, 1993.

\bibitem{WD}
R.~Wenocur and R.~Dudley.
\newblock Some special vapnik-chervonenkis classes.
\newblock {\em Discrete Math.}, 33:313--318, 1981.

\end{thebibliography}
\end{document}